\documentclass[11pt]{article}
\usepackage[margin=1in]{geometry}
\usepackage{booktabs}
\usepackage{array}
\usepackage{multirow}
\usepackage{multicol}

\usepackage[
    backend=biber,
    style=numeric-comp,
    maxcitenames=2,
    sortcites,
    sorting=none,
    giveninits=true,
    natbib,
    hyperref=true,
    maxbibnames=99,
    uniquename=init,
    doi=false,isbn=false,url=false,eprint=false
]{biblatex}

\usepackage[colorlinks=true, linkcolor=BlueViolet, citecolor=BlueViolet, urlcolor=BlueViolet]{hyperref}

\title{Conformal inference is (almost) free for neural networks\\trained with early stopping}

\usepackage{subfigure}
\usepackage{placeins}
\usepackage{graphicx, color, url}
\usepackage{dsfont}
\usepackage{amssymb}
\usepackage{amsfonts}
\usepackage{url}
\usepackage{amsbsy}
\usepackage{setspace}
\usepackage{bm}
\usepackage{textcomp}
\usepackage{booktabs}
\usepackage{rotating}

\usepackage{algorithm, algorithmic}

\usepackage{amsmath,amsthm}

\usepackage{graphicx}
\usepackage{amsbsy}
\usepackage{mathtools}
\usepackage{bbm}

\usepackage[dvipsnames]{xcolor}

\newtheorem{theorem}{Theorem}
\newtheorem{corollary}{Corollary}
\newtheorem{proposition}{Proposition}

\newtheorem{lemma}{Lemma}

\renewcommand{\P}[1]{\mathbb{P}\left[#1\right]}
\newcommand{\E}[1]{\mathbb{E}\left[#1\right]}
\newcommand{\I}[1]{\mathbbm{1}\left[#1\right]}

\makeatletter
\newcommand*\bigcdot{\mathpalette\bigcdot@{.5}}
\newcommand*\bigcdot@[2]{\mathbin{\vcenter{\hbox{\scalebox{#2}{$\m@th#1\bullet$}}}}}
\makeatother

\DeclareMathOperator*{\argmin}{arg\,min}

\usepackage[colorinlistoftodos]{todonotes}

\author{Ziyi Liang\thanks{Department of Mathematics, University of Southern California, Los Angeles, CA, USA.}, Yanfei Zhou\thanks{Department of Data Sciences and Operations, University of Southern California, Los Angeles, CA, USA.},
Matteo Sesia\footnotemark[2]}

\begin{document}

\maketitle

\begin{abstract}
Early stopping based on hold-out data is a popular regularization technique designed to mitigate overfitting and increase the predictive accuracy of neural networks. Models trained with early stopping often provide relatively accurate predictions, but they generally still lack precise statistical guarantees unless they are further calibrated using independent hold-out data. This paper addresses the above limitation with {\em conformalized early stopping}: a novel method that combines early stopping with conformal calibration while efficiently recycling the same hold-out data. This leads to models that are both accurate and able to provide exact predictive inferences without multiple data splits nor overly conservative adjustments. Practical implementations are developed for different learning tasks---outlier detection, multi-class classification, regression---and their competitive performance is demonstrated on real data.
\end{abstract}

\section{Introduction}

Deep neural networks can detect complex data patterns and leverage them to make accurate predictions in many applications, including computer vision, natural language processing, and speech recognition, to name a few examples.
These models can sometimes even outperform skilled humans~\citep{silver2016mastering}, but they still make mistakes.
Unfortunately, the severity of these mistakes is compounded by the fact that the predictions computed by neural networks are often overconfident~\citep{guo2017calibration}, partly due to overfitting~\citep{thulasidasan2019mixup,ovadia2019can}.
Several training strategies have been developed to mitigate overfitting, including dropout \citep{srivastava2014dropout}, batch normalization~\citep{ioffe2015batch}, weight normalization~\citep{salimans2016weight}, data augmentation~\citep{shorten2019survey}, and early stopping \citep{PRECHELT1997}; the latter is the focus of this paper.

Early stopping consists of continuously evaluating after each batch of stochastic gradient updates (or {\em epoch}) the predictive performance of the current model on {\em hold-out} independent data.
After a large number of gradient updates, only the intermediate model achieving the best performance on the hold-out data is utilized to make predictions.
This strategy is often effective at mitigating overfitting and can produce relatively accurate predictions compared to fully trained models, but it does not fully resolve overconfidence because it does not lead to models with finite-sample guarantees.

A general framework for quantifying the predictive uncertainty of any {\em black-box} machine learning model is that of conformal inference~\citep{vovk2005algorithmic}.
The idea is to apply a pre-trained model to a {\em calibration} set of hold-out observations drawn at random from the target population.
If the calibration data are exchangeable with the test point of interest, the model performance on the calibration set can be translated into statistically rigorous predictive inferences.
This framework is flexible and can accommodate different learning tasks, including out-of-distribution testing~\citep{smith2015conformal}, classification~\citep{vovk2003mondrian}, and regression~\citep{vovk2005algorithmic}.
For example, in the context of classification, conformal inference can give prediction sets that contain the correct label for a new data point with high probability.
In theory, the quality of the trained model has no consequence on the {\em average} validity of conformal inferences, but it does affect their reliability and usefulness on a case-by-case level.
In particular, conformal uncertainty estimates obtained after calibrating an overconfident model may be too conservative for some test cases and too optimistic for others~\citep{romano2020classification}.
The goal of this paper is to combine conformal calibration with standard early stopping training techniques as efficiently as possible, in order to produce more reliable predictive inferences with a finite amount of available data.

Achieving high accuracy with deep learning often requires large training sets~\citep{marcus2018deep}, and conformal inference makes the overall pipeline even more data-intensive.
As high-quality observations can be expensive to collect, in some situations practitioners may naturally wonder whether the advantage of having principled uncertainty estimates is worth a possible reduction in predictive accuracy due to fewer available training samples.
This concern is relevant because the size of the calibration set cannot be too small if one wants stable and reliable conformal inferences~\citep{vovk2012conditional,sesia2020comparison}.
In fact, very large calibration sets may be necessary to obtain stronger conformal inferences that are valid not only on average but also conditionally on some important individual features; see \citet{vovk2003mondrian,romano2019malice,barber2019limits}.

This paper resolves the above dilemma by showing that conformal inferences for deep learning models trained with early stopping can be obtained almost ``for free''---without spending more precious data.
More precisely, we present an innovative method that blends model training with early stopping and conformal calibration using the same hold-out samples, essentially obtaining rigorous predictive inferences at no additional data cost compared to standard early stopping.
It is worth emphasizing this result is not trivial.
In fact, naively applying existing conformal calibration methods using the same hold-out samples utilized for early stopping would not lead to theoretically valid inferences, at least not without resorting to very conservative corrections.

The paper is organized as follows. Section \ref{sec:ces} develops our {\em conformalized early stopping} (CES) method, starting from outlier detection and classification, then addressing regression. Section \ref{sec:numerical_results} demonstrates the advantages of CES through numerical experiments. Section~\ref{sec:discussion} concludes with a discussion and some ideas for further research.
Additional details and results, including a theoretical analysis of the naive benchmark mentioned above, can be found in the Appendices, along with all mathematical proofs.

\subsection*{Related Work}

Conformal inference~\citep{saunders1999transduction,vovk1999machine,vovk2005algorithmic} has become a very rich and active area of research~\citep{lei2013distribution,lei2014distribution,lei2018distribution,barber2019predictive}.
Many prior works studied the computation of efficient conformal inferences starting from pre-trained {\em black-box} models, including for example in the context of outlier detection~\citep{smith2015conformal,guan2019prediction,Liang_2022_integrative_p_val,bates2021testing}, classification~\citep{vovk2003mondrian,hechtlinger2018cautious,romano2020classification,angelopoulos2020uncertainty,bates2021distributionfree}, and regression \citep{vovk2005algorithmic,lei2014distribution,romano2019conformalized}.
Other works have studied the general robustness of conformal inferences to distribution shifts~\citep{tibshirani2019conformal,sesia2022conformal} and, more broadly, to failures of the data exchangeability assumption~\citep{barber2022conformal,gibbs2022conformal}.
Our research is orthogonal, as we look inside the black-box model and develop a novel early-stopping training technique that is naturally integrated with conformal calibration. Nonetheless, the proposed method could be combined with those described in the aforementioned papers.
Other recent research has explored different ways of bringing conformal inference into the learning algorithms \citep{colombo2020training,bellotti2021optimized,stutz2021learning,einbinder2022training}, and some of those works apply standard early stopping techniques, but they do not address our problem.

This paper is related to \citet{efficiency_first_cp}, which proposed a general theoretical adjustment for conformal inferences computed after model selection. That method could be utilized to account for early stopping without further data splits, as explained in Section~\ref{sec:naive_benchmark}. However, we will demonstrate that even an improved version of such analysis remains overly conservative in the context of model selection via early stopping, and the alternative method developed in this paper performs much better in practice.
Our solution is inspired by Mondrian conformal inference~\citep{vovk2003mondrian} as well as by the integrative conformal method of \citet{Liang_2022_integrative_p_val}. The latter deals with the problem of selecting the best model from an arbitrary machine learning toolbox  to obtain the most powerful conformal p-values for outlier testing. The idea of \citet{Liang_2022_integrative_p_val} extends naturally to the early stopping problem in the special cases of outlier detection and classification, but the regression setting requires substantial technical innovations.
The work of \citet{Liang_2022_integrative_p_val} is also related to \citet{marandon2022machine}, although the latter is more distant from this paper because it focuses on theoretically controlling the false discovery rate~\citep{benjamini1995controlling}  in multiple testing problems.
Finally, this paper draws inspiration from \citet{kim2020predictive}, which shows that models trained with bootstrap (or bagging) techniques can also lead to valid conformal inferences essentially for free.

\section{Methods} \label{sec:ces}

\subsection{Standard Conformal Inference and Early Stopping} \label{sec:setup}

Consider $n$ data points, $Z_i$ for $i \in \mathcal{D} = [n] = \{1,\ldots,n\}$, sampled exchangeably (e.g., i.i.d.)~from an unknown distribution $P_{Z}$ with support on some space $\mathcal{Z}$. Consider also an additional test sample, $Z_{n+1}$. 
In the context of outlier detection, one wishes to test whether $Z_{n+1}$ was sampled exchangeably from $P_Z$.
In classification or regression, one can write $Z_i = (X_i, Y_i)$, where $X_i$ is a feature vector while $Y_i$ is a discrete category or a continuous response, and the goal is to predict the unobserved value of $Y_{n+1}$ given $X_{n+1}$ and the data in $\mathcal{D}$.

\begin{figure*}[!htb]
  \centering
  \subfigure[Conformal inference after early stopping.]{\includegraphics[width=0.4\linewidth]{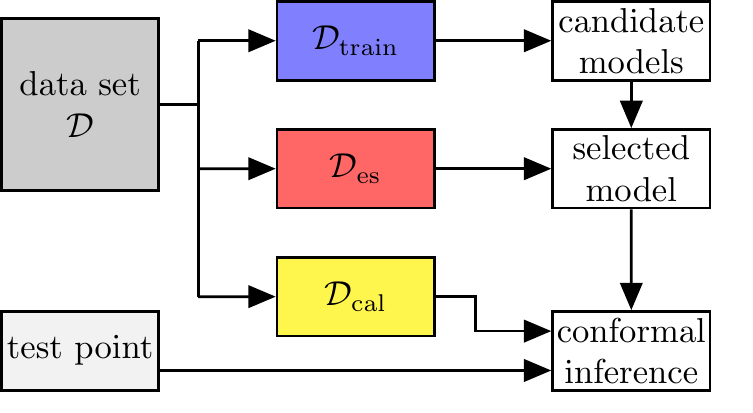}}~~~~~
  \subfigure[Conformalized early stopping (CES).]{\includegraphics[width=0.4\linewidth]{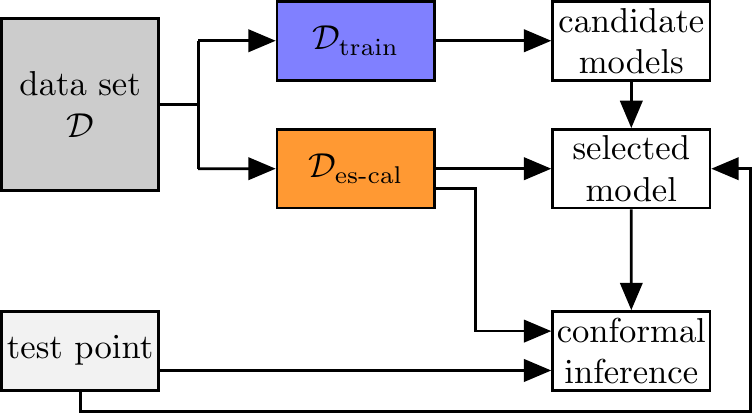}}
  \caption{Schematic visualization of rigorous conformal inferences for models trained with early stopping. (a) Conventional pipeline requiring a three-way sample split. (b) Conformalized early stopping, requiring only a two-way split.}
  \label{fig:ces_data_splitting}
\end{figure*}

The standard pipeline begins by randomly splitting the data in $\mathcal{D}$ into three disjoint subsets: $\mathcal{D}_{\text{train}}, \mathcal{D}_{\text{es}}, \mathcal{D}_{\text{cal}} \subset [n]$.
The samples in $\mathcal{D}_{\text{train}}$ are utilized to train a model $M$ via stochastic gradient descent, in such a way as to (approximately) minimize the desired loss $\mathcal{L}$, while the observations in $\mathcal{D}_{\text{es}}$ and $\mathcal{D}_{\text{cal}}$ are held out.
We denote by $M_t$ the model learnt after $t$ epochs of stochastic gradient descent, for any $t \in [t^{\text{max}}]$, where $t^{\text{max}}$ is a pre-determined maximum number of epochs. For simplicity, $\mathcal{L}$ is assumed to be an additive loss, in the sense that its value calculated on the training data after $t$ epochs is
$\mathcal{L}_{\text{train}}(M_t) = \sum_{i \in \mathcal{D}_{\text{train}}} \ell(M_t; Z_i)$,
for some appropriate function $\ell$. For example, a typical choice for regression would be the squared-error loss: $\ell(M_t; Z_i) = \left[Y_i - \hat{\mu}(X_i; M_t) \right]^2$, where $\hat{\mu}(X_i; M_t)$ indicates the value of the regression function at $X_i$, as estimated by $M_t$.
Similarly, the loss evaluated on $\mathcal{D}_{\text{es}}$ is denoted as $\mathcal{L}_{\text{es}}(M_t) = \sum_{i \in \mathcal{D}_{\text{es}}} \ell(M_t; Z_i)$.
After training for $t^{\text{max}}$ epochs, early stopping selects the model $\hat{M}_{\text{es}}$ that minimizes the loss on $\mathcal{D}_{\text{es}}$:
$\hat{M}_{\text{es}} = \argmin_{M_t \, : \, 0 \leq t \leq t^{\text{max}}} \mathcal{L}_{\text{es}}(M_t)$.
Conformal calibration of $\hat{M}_{\text{es}}$ is then conducted using the independent hold-out data set $\mathcal{D}_{\text{cal}}$, as sketched in Figure~\ref{fig:ces_data_splitting}~(a).
This pipeline requires a three-way data split because: (i) $\mathcal{D}_{\text{train}}$ and $\mathcal{D}_{\text{es}}$ must be disjoint to ensure the early stopping criterion is effective at mitigating overfitting; and (ii) $\mathcal{D}_{\text{cal}}$ must be disjoint from $\mathcal{D}_{\text{train}} \cup \mathcal{D}_{\text{es}}$ to ensure the performance of the selected model $\hat{M}_{\text{es}}$ on the calibration data gives us an unbiased preview of its future performance at test time, enabling valid conformal inferences.

\subsection{The Limitations of a Naive Benchmark}
\label{sec:naive_benchmark}

\begin{figure*}[!htb]
  \centering
  \includegraphics[width=0.4\linewidth]{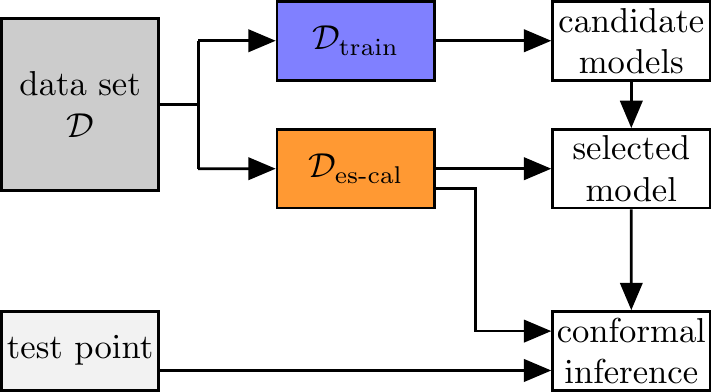}
  \caption{Schematic visualization of heuristic conformal inferences based on a naive benchmark that utilizes the same hold-out data twice. }
  \label{fig:naive_diagram}
\end{figure*}

An intuitive alternative to the standard approach described in the previous section is to naively perform standard conformal inference using the same hold-out samples utilized for early stopping, as sketched in Figure~\ref{fig:naive_diagram}.
This approach, which we call the {\em naive benchmark}, may seem appealing as it avoids a three-way data split, but it does not provide rigorous inferences. In fact, the necessary exchangeability with the test point is broken if the same hold-out data are used twice---first to evaluate the early stopping criterion and then to perform conformal calibration.
In principle, the issue could be corrected by applying a conservative adjustment to the nominal significance level of the conformal inferences, as studied by \citet{efficiency_first_cp} and reviewed in Appendix~\ref{app:naive-benchmarks}. 
However, this leads to overly conservative inferences in practice when applied with the required theoretical correction, as demonstrated by the numerical experiments summarized in Figure~\ref{fig:exp_regression_bio}, even if a tighter adjustment developed in Appendix~\ref{app:naive-benchmarks} is utilized instead of that of \citet{efficiency_first_cp}.
Intuitively, the problem is that there tend to be complicated dependencies among the candidate models provided to the early stopping algorithm, but the available analyses are not equipped to handle such intricacies and must therefore take a pessimistic viewpoint of the model selection process.
Thus, the naive benchmark remains unsatisfactory, although it can serve as an informative benchmark for the novel method developed in this paper.
Interestingly, we will see empirically that the naive benchmark applied without the required theoretical corrections often performs similarly to the rigorous method proposed in this paper, especially for large data sets.

\begin{figure}[!htb]
    \centering
    \includegraphics[width=0.65\linewidth]{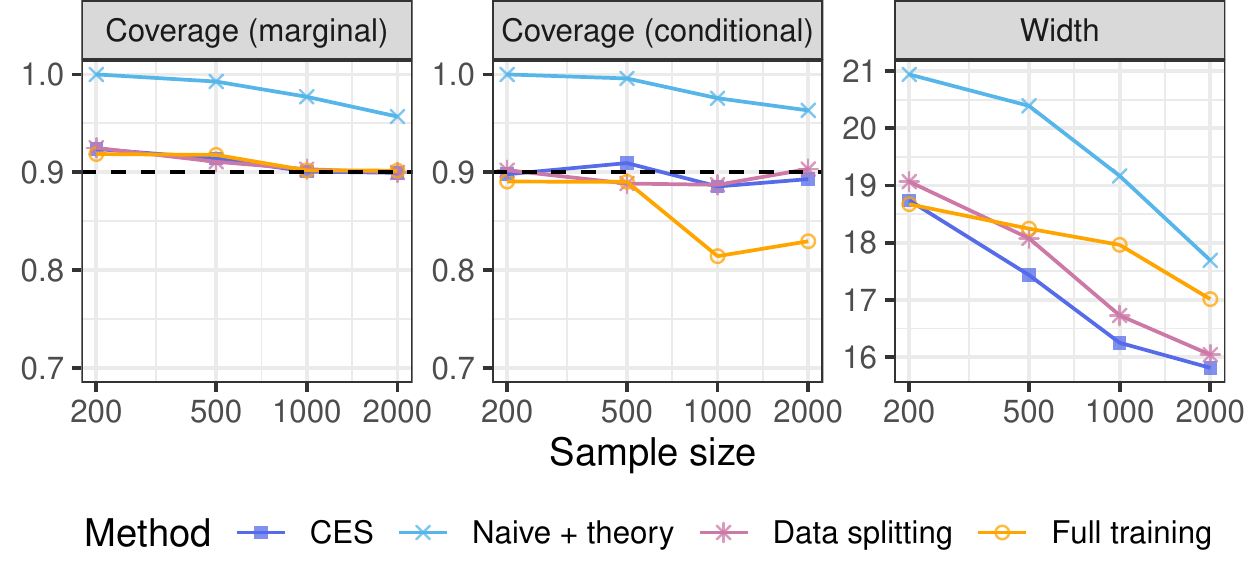}\vspace{-0.3cm}
    \caption{Average performance, as a function of the sample size, of conformal inferences based on neural networks trained and calibrated with different methods, on the {\em bio} regression data~\citep{data-bio}. Ideally, the coverage of the conformal prediction intervals should be close to 90\% and their width should be small. All methods shown here guarantee 90\% marginal coverage. See Table~\ref{tab:exp_regression_bio} for more detailed results and standard errors.}
    \label{fig:exp_regression_bio}
\end{figure}

\subsection{Preview of our Contribution}

This paper develops a novel conformalized early stopping (CES) method to jointly carry out both early stopping and conformal inference using a single hold-out data set, denoted in the following as  $\mathcal{D}_{\text{es-cal}}$. The advantage of this approach is that it avoids a three-way data split, so that more samples can be allocated to  $\mathcal{D}_{\text{train}}$, without breaking the required data exchangeability. As a result, CES often leads to relatively more reliable and informative conformal inferences compared to other existing approaches; e.g., see for example the empirical performance preview shown in Figure~\ref{fig:exp_regression_bio}.
The CES method is based on the following idea inspired by \citet{Liang_2022_integrative_p_val}.
Valid conformal inferences can be obtained by calibrating $\hat{M}_{\text{es}}$ using the same data set $\mathcal{D}_{\text{es-cal}}$ used for model selection, as long as the test sample  $Z_{n+1}$ is also involved in the early stopping rule exchangeably with all other samples in $\mathcal{D}_{\text{es-cal}}$.
This concept, sketched in Figure~\ref{fig:ces_data_splitting}~(b), is not obvious to translate into a practical method, however, for two reasons.
Firstly, the ground truth for the test point (i.e., its outlier status or its outcome label) is unknown. Secondly, the method may need to be repeatedly applied for a large number of distinct test points in a computationally efficient way, and one cannot re-train the model separately for each test point.
In the next section, we will explain how to overcome these challenges in the special case of early stopping for outlier detection; then, the solution will be extended to the classification and regression settings.

\subsection{CES for Outlier Detection} \label{sec:outlier}

Consider testing whether $Z_{n+1}$, is an {\em inlier}, in the sense that it was sampled from $P_Z$ exchangeably with the data in $\mathcal{D}$.
Following the notation of Section~\ref{sec:setup}, consider a partition of $\mathcal{D}$ into two subsets, $\mathcal{D}_{\text{train}}$ and $\mathcal{D}_{\text{es-cal}}$, chosen at random independently of everything else, such that $\mathcal{D}=\mathcal{D}_{\text{train}} \cup \mathcal{D}_{\text{es-cal}}$.
The first step of CES consists of training a deep one-class classifier $M$ using the data in $\mathcal{D}_{\text{train}}$ via stochastic gradient descent for $t^{\text{max}}$ epochs, storing all parameters characterizing the intermediate model after each $\tau$ epochs. We refer to $\tau \in [t^{\text{max}}]$ as the {\em storage period}, a parameter pre-defined by the user.
Intuitively, a smaller $\tau$ increases the memory cost of CES but may also lead to the selection of a more accurate model.
While the memory cost of this approach is higher compared to that of standard early-stopping training techniques, which only require storing one model at a time, it is not prohibitively expensive.
In fact, the candidate models do not need to be kept in precious RAM memory but can be stored on a relatively cheap hard drive.
As reasonable choices of $\tau$ may typically be in the order of $T = \lfloor t^{\text{max}} / \tau \rfloor \approx 100$, the cost of CES is not excessive in many real-world situations.
For example, it takes approximately 100 MB to store a pre-trained standard ResNet50 computer vision model, implying that CES would require approximately 10 GB of storage in such applications---today this costs less than \$0.25/month in the cloud.

After pre-training and storing $T$ candidate models, namely $M_{t_1}, \ldots, M_{t_T}$ for some sub-sequence $(t_1, \ldots, t_T)$ of $[t^{\text{max}}]$, the next step is to select the appropriate early-stopped model based on the hold-out data in $\mathcal{D}_{\text{es-cal}}$ as well as the test point $Z_{n+1}$.
Following the notation of Section~\ref{sec:setup}, define the value of the one-class classification loss $\mathcal{L}$ for model $M_t$, for any $t \in [T]$, evaluated on $\mathcal{D}_{\text{es-cal}}$ as: $\mathcal{L}_{\text{es-cal}}(M_t) = \sum_{i \in \mathcal{D}_{\text{es-cal}}} \ell(M_t; Z_i)$.
Further, for any $z \in \mathcal{Z}$, define also $\mathcal{L}_{\text{es-cal}}^{+1}(M_t,z)$ as:
\begin{align} \label{eq:loss-ces}
\mathcal{L}_{\text{es-cal}}^{+1}(M_t,z) = \mathcal{L}_{\text{es-cal}}(M_t) + \ell(M_t; z).
\end{align}
Therefore, $\mathcal{L}_{\text{es-cal}}^{+1}(M_t,Z_{n+1})$ can be interpreted as the cumulative value of the loss function calculated on an augmented hold-out data set including also $Z_{n+1}$.
Then, we select the model $\hat{M}_{\text{ces}}(Z_{n+1})$ minimizing $\mathcal{L}_{\text{es-cal}}^{+1}(M_t,Z_{n+1})$:
\begin{align} \label{eq:ces-model}
\hat{M}_{\text{ces}}(Z_{n+1}) = \argmin_{M_{t_j} \, : \, 1 \leq j \leq T} \mathcal{L}_{\text{es-cal}}^{+1}(M_{t_j},Z_{n+1}).
\end{align}
Note that the computational cost of evaluating~\eqref{eq:ces-model} is negligible compared to that of training the models.

Next, the selected model $\hat{M}_{\text{ces}}(Z_{n+1})$ is utilized to compute a conformal p-value~\citep{bates2021testing} to test whether $Z_{n+1}$ is an inlier.
In particular, $\hat{M}_{\text{ces}}(Z_{n+1})$ is utilized to compute {\em nonconformity scores} $\hat{S}_i(Z_{n+1})$ for all samples $i \in \mathcal{D}_{\text{es-cal}} \cup \{n+1\}$. These scores rank the observations in $\mathcal{D}_{\text{es-cal}} \cup \{n+1\}$ based on how the one-class classifier $\hat{M}_{\text{ces}}(Z_{n+1})$ perceives them to be similar to the training data; by convention, a smaller value of $\hat{S}_i(Z_{n+1})$ suggests $Z_i$ is more likely to be an outlier.
Suitable scores are typically included in the output of standard one-class classification models, such as those provided by the Python library PyTorch.
For simplicity, we assume all scores are almost-surely distinct; otherwise, ties can be broken at random by adding a small amount of independent noise. Then, the conformal p-value $\hat{u}_0(Z_{n+1})$ is given by the usual formula:
\begin{align}\label{eq:conformal_pval}
    \hat{u}_0(Z_{n+1}) = \frac{1 + |i \in \mathcal{D}_{\text{es-cal}}: \hat{S}_{i} \leq \hat{S}_{n+1}|}{1+|\mathcal{D}_{\text{es-cal}}|},
\end{align}
making the dependence of $\hat{S}_{i}$ on $Z_{n+1}$ implicit in the interest of space.
This method, outlined by Algorithm~\ref{alg:od_full_seq}, gives p-values that are exactly valid in finite samples, in the sense that they are stochastically dominated by the uniform distribution under the null hypothesis.
\begin{algorithm}[H]
    \caption{Conformalized early stopping for outlier detection}
    \label{alg:od_full_seq}
    \begin{algorithmic}[1]
        \STATE \textbf{Input}: Exchangeable data points $Z_1, \ldots, Z_n$; test point $Z_{n+1}$.
        \STATE \textcolor{white}{\textbf{Input}:} Maximum number of training epochs $t^{\text{max}}$; storage period hyper-parameter $\tau$.
        \STATE \textcolor{white}{\textbf{Input}:} One-class classifier trainable via (stochastic) gradient descent.
        \STATE Randomly split the exchangeable data points into $\mathcal{D}_{\text{train}}$ and $\mathcal{D}_{\text{es-cal}}$.
        \STATE Train the one-class classifier for $t^{\text{max}}$ epochs and save the intermediate models $M_{t_1} , \dots, M_{t_T}$.
        \STATE Pick the most promising model $\hat{M}_{\text{ces}}(Z_{n+1})$ based on~\eqref{eq:ces-model}, using the data in $\mathcal{D}_{\text{es-cal}} \cup \{n+1\}$.
        \STATE Compute nonconformity scores $\hat{S}_i(Z_{n+1})$ for all $i \in \mathcal{D}_{\text{es-cal}} \cup \{n+1\}$ using model $\hat{M}_{\text{ces}}(Z_{n+1})$.
        \STATE \textbf{Output}: Conformal p-value $\hat{u}_0(Z_{n+1})$ given by \eqref{eq:conformal_pval}.
    \end{algorithmic}
\end{algorithm}

\begin{theorem}\label{thm:od_full}
Assume $Z_{1}, \ldots, Z_{n}, Z_{n+1}$ are exchangeable random samples, and let $\hat{u}_0(Z_{n+1})$ be the output of Algorithm~\ref{alg:od_full_seq}, as given in~\eqref{eq:conformal_pval}. Then, $\P{\hat{u}_0(Z_{n+1}) \leq \alpha} \leq \alpha$ for any $\alpha \in (0,1)$.
\end{theorem}

\subsection{CES for Classification}  \label{sec:classification}

The above CES method will now be extended to deal with $K$-class classification problems, for any $K \geq 2$.
Consider $n$ exchangeable pairs of observations $(X_i,Y_i)$, for $i \in \mathcal{D} = [n]$, and a test point $(X_{n+1}, Y_{n+1})$ whose label $Y_{n+1} \in [K]$ has not yet been observed. The goal is to construct an informative prediction set for $Y_{n+1}$ given the observed features $X_{n+1}$ and the rest of the data, assuming $(X_{n+1},Y_{n+1})$ is exchangeable with the observations indexed by $\mathcal{D}$.
An ideal goal would be to construct the smallest possible prediction set with guaranteed {\em feature-conditional coverage} at level $1-\alpha$, for any fixed $\alpha \in (0,1)$. Formally, a prediction set $\hat{C}_{\alpha}(X_{n+1}) \subseteq [K]$ has feature-conditional coverage at level $1-\alpha$ if $\mathbb{P}[Y_{n+1} \in \hat{C}_{\alpha}(X_{n+1}) \mid X_{n+1} = x] \geq 1-\alpha$, for any $x \in \mathcal{X}$, where $\mathcal{X}$ is the feature space.
Unfortunately, perfect feature-conditional coverage is extremely difficult to achieve unless the feature space $\mathcal{X}$ is very small~\cite{barber2019limits}. Therefore, in practice, one must be satisfied with obtaining relatively weaker guarantees, such as {\em label-conditional coverage} and {\em marginal coverage}. Formally,  $\hat{C}_{\alpha}(X_{n+1})$ has $1-\alpha$ label-conditional coverage if $\mathbb{P}[Y_{n+1} \in \hat{C}_{\alpha}(X_{n+1}) \mid Y_{n+1} = y] \geq 1-\alpha$, for any $y \in [K]$, while marginal coverage corresponds to $\mathbb{P}[Y_{n+1} \in \hat{C}_{\alpha}(X_{n+1}) ] \geq 1-\alpha$. Label-conditional coverage is stronger than marginal coverage, but both criteria are useful because the latter is easier to achieve with smaller (and hence more informative) prediction sets.

We begin by focusing on label-conditional coverage, as this follows most easily from the results of Section~\ref{sec:outlier}. This solution will be extended in Appendix~\ref{app:class-marg} to target marginal coverage.
The first step of CES consists of randomly splitting $\mathcal{D}$ into two subsets, $\mathcal{D}_{\text{train}}$ and $\mathcal{D}_{\text{es-cal}}$, as in Section~\ref{sec:outlier}. The samples in $\mathcal{D}_{\text{es-cal}}$ are further divided into subsets $\mathcal{D}^y_{\text{es-cal}}$ with homogeneous labels; that is, $\mathcal{D}^y_{\text{es-cal}} = \{i \in \mathcal{D}_{\text{es-cal}} : Y_i = y \}$ for each $y \in [K]$.
The data in $\mathcal{D}_{\text{train}}$ are utilized to train a neural network classifier via stochastic gradient descent, storing the intermediate candidate models $M_t$ after each $\tau$ epochs.
This is essentially the same approach as in Section~\ref{sec:outlier}, with the only difference being that the neural network is now designed to perform $K$-class classification rather than one-class classification. Therefore, this neural network should have a soft-max layer with $K$ nodes near its output, whose values corresponding to an input data point with features $x$ are denoted as $\hat{\pi}_y(x)$, for all $y \in [K]$.
Intuitively, we will interpret $\hat{\pi}_y(x)$ as approximating (possibly inaccurately) the true conditional data-generating distribution; i.e., $\hat{\pi}_{y}(x) \approx \P{Y=y \mid X=x}$.

For any model $M_t$, any $x \in \mathcal{X}$, and any $y \in [K]$, define the augmented loss $\mathcal{L}_{\text{es-cal}}^{+1}(M_t,x,y)$  as:
\begin{align} \label{eq:loss-ces-class}
\mathcal{L}_{\text{es-cal}}^{+1}(M_t,x,y) = \mathcal{L}_{\text{es-cal}}(M_t) + \ell(M_t; x, y).
\end{align}
Concretely, a typical choice for $\ell$ is the cross-entropy loss: $\ell(M_t; x, y) = - \log \hat{\pi}^t_y(x)$, where $\hat{\pi}^t$ denotes the soft-max probability distribution estimated by model $M_t$.
Intuitively, $\mathcal{L}_{\text{es-cal}}^{+1}(M_t,x,y)$ is the cumulative value of the loss function calculated on an augmented hold-out data set including also the imaginary test sample $(x,y)$.
Then, for any $y \in [K]$, CES selects the model $\hat{M}_{\text{ces}}(X_{n+1},y)$ minimizing $\mathcal{L}_{\text{es-cal}}^{+1}(M_t,X_{n+1},y)$ among the $T$ stored models:
\begin{align} \label{eq:ces-model-class}
\hat{M}_{\text{ces}}(X_{n+1},y) = \argmin_{M_{t_j} \, : \, 1 \leq j \leq T} \mathcal{L}_{\text{es-cal}}^{+1}(M_{t_j},X_{n+1},y).
\end{align}
The selected model $\hat{M}_{\text{ces}}(X_{n+1},y)$ is then utilized to compute a conformal p-value for testing whether $Y_{n+1}=y$.
In particular, we compute nonconformity scores $\hat{S}_i^y(X_{n+1})$ for all $i \in \mathcal{D}^y_{\text{es-cal}} \cup \{n+1\}$, imagining that $Y_{n+1}=y$. Different types of nonconformity scores can be easily accommodated, but in this paper, we follow the {\em adaptive} strategy of~\citet{romano2020classification}. The computation of these nonconformity scores based on the selected model $\hat{M}_{\text{ces}}$ is reviewed in Appendix~\ref{app:class-scores}.
Here, we simply note the p-value is given by:
\begin{align}\label{eq:conformal_pval-class}
    \hat{u}_y(X_{n+1}) = \frac{1 + |i \in \mathcal{D}^y_{\text{es-cal}}: \hat{S}^y_{i} \leq \hat{S}^y_{n+1}|}{1+|\mathcal{D}^y_{\text{es-cal}}|},
\end{align}
again making the dependence of $\hat{S}^y_{i}$ on $X_{n+1}$ implicit.
Finally, the prediction set $\hat{C}_{\alpha}(X_{n+1})$ is constructed by including all possible labels for which the corresponding null hypothesis cannot be rejected at level $\alpha$:
\begin{align} \label{eq:pred-set-class}
  \hat{C}_{\alpha}(X_{n+1}) = \left\{ y \in [K] : \hat{u}_y(X_{n+1})  \geq \alpha \right\}.
\end{align}
This method, outlined by Algorithm~\ref{alg:class_full_seq}, guarantees label-conditional coverage at level $1-\alpha$.

\begin{algorithm}[H]
    \caption{Conformalized early stopping for multi-class classification}
    \label{alg:class_full_seq}
    \begin{algorithmic}[1]
        \STATE \textbf{Input}: Exchangeable data points $(X_{1},Y_{1}), \ldots, (X_{n},Y_{n})$ with labels $Y_i \in [K]$.
        \STATE \textcolor{white}{\textbf{Input}:} Test point with features $X_{n+1}$. Desired coverage level $1-\alpha$.
        \STATE \textcolor{white}{\textbf{Input}:} Maximum number of training epochs $t^{\text{max}}$; storage period hyper-parameter $\tau$.
        \STATE \textcolor{white}{\textbf{Input}:} $K$-class classifier trainable via (stochastic) gradient descent.
        \STATE Randomly split the exchangeable data points into $\mathcal{D}_{\text{train}}$ and $\mathcal{D}_{\text{es-cal}}$.
        \STATE Train the $K$-class classifier for $t^{\text{max}}$ epochs and save the intermediate models $M_{t_1} , \dots, M_{t_T}$.
        \FOR{$ y \in [K]$}
        \STATE Define $\mathcal{D}^y_{\text{es-cal}} = \{i \in \mathcal{D}_{\text{es-cal}} : Y_i = y \}$ and imagine $Y_{n+1}=y$.
        \STATE Pick the model $\hat{M}_{\text{ces}}(X_{n+1},y)$ according to~\eqref{eq:ces-model-class}, using the data in $\mathcal{D}_{\text{es-cal}} \cup \{n+1\}$.
        \STATE Compute scores $\hat{S}_i^y(X_{n+1})$, $\forall i \in \mathcal{D}^y_{\text{es-cal}} \cup \{n+1\}$, using $\hat{M}_{\text{ces}}(X_{n+1},y)$; see Appendix~\ref{app:class-scores}.
        \STATE Compute the conformal p-value $\hat{u}_y(X_{n+1})$ according to \eqref{eq:conformal_pval-class}.
        \ENDFOR
        \STATE \textbf{Output}: Prediction set $\hat{C}_{\alpha}(X_{n+1})$ given by \eqref{eq:pred-set-class}.
    \end{algorithmic}
\end{algorithm}

\begin{theorem}\label{thm:class_full}
Assume $(X_{1},Y_{1}), \ldots, (X_{n+1},Y_{n+1})$ are exchangeable, and let $\hat{C}_{\alpha}(X_{n+1})$ be the output of Algorithm~\ref{alg:class_full_seq}, as given in~\eqref{eq:pred-set-class}, for any given $\alpha \in (0,1)$.
Then, $\mathbb{P}[Y_{n+1} \in \hat{C}_{\alpha}(X_{n+1}) \mid Y_{n+1} = y] \geq 1-\alpha$ for any $y \in [K]$.
\end{theorem}

\subsection{CES for Regression} \label{sec:regression}

This section extends CES to regression problems with a continuous outcome.
As in the previous sections, consider a data set containing $n$ exchangeable observations $(X_i,Y_i)$, for $i \in \mathcal{D} = [n]$, and a test point $(X_{n+1}, Y_{n+1})$ with a latent label $Y_{n+1} \in \mathbb{R}$. The goal is to construct a reasonably narrow {\em prediction interval} $\hat{C}_{\alpha}(X_{n+1})$ for $Y_{n+1}$ that is guaranteed to have marginal coverage above some level $1-\alpha$, i.e., $\mathbb{P}[Y_{n+1} \in \hat{C}_{\alpha}(X_{n+1}) ] \geq 1-\alpha$, and can also practically achieve reasonably high feature-conditional coverage.
Developing a CES method for this problem is more difficult compared to the classification case studied in Section~\ref{sec:classification} due to the infinite number of possible values for $Y_{n+1}$. In fact, a naive extension of Algorithm~\ref{alg:class_full_seq} would be computationally unfeasible in the regression setting, for the same reason why full-conformal prediction \citep{vovk2005algorithmic} is generally impractical.
The novel solution described below is designed to leverage the particular structure of an early stopping criterion based on the squared-error loss evaluated on hold-out data. Focusing on the squared-error loss makes CES easier to implement and explain using classical {\em absolute residual} nonconformity scores~\citep{vovk2005algorithmic,lei2016RegressionPS}.
However, similar ideas could also be repurposed to accommodate other scores, such as those based on quantile regression~\citep{romano2019conformalized}, conditional distributions~\citep{chernozhukov2019distributional,izbicki2019flexible}, or conditional histograms~\citep{sesia2021conformal}.

\begin{figure}[!htb]
    \centering
    \includegraphics[width=0.45\linewidth]{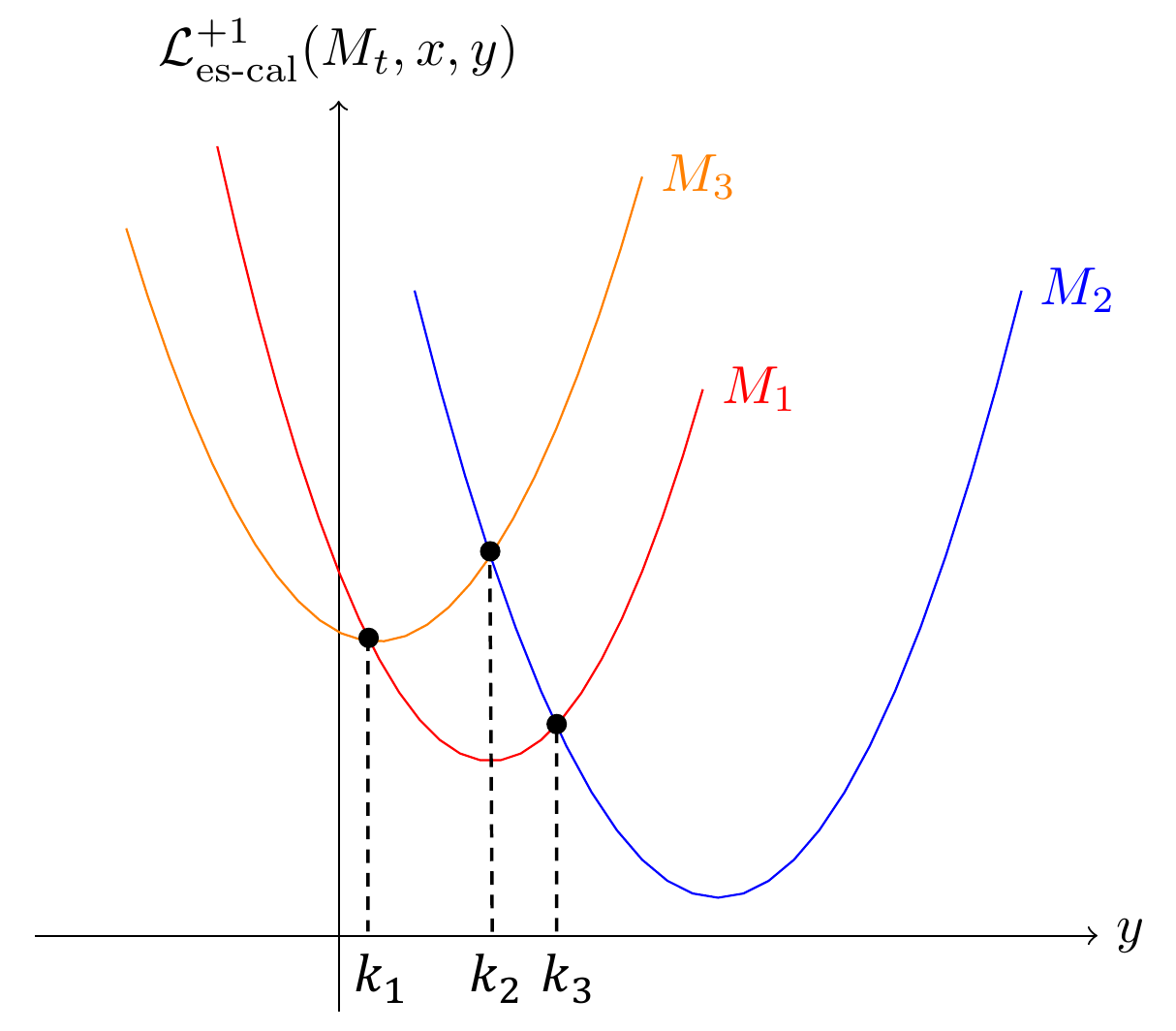}
    \caption{Squared-error loss on test-augmented hold-out data for three alternative regression models $M_1, M_2$ and $M_3$, as a function of the place-holder outcome $y$ for the test point. The CES method utilizes the best model for each possible value of $y$, which is identified by the lower envelope of these three parabolas. In this case, the lower envelope has two finite knots at $k_{1}$ and $k_{3}$.}
    \label{fig:quadratic_losses}%
\end{figure}

As usual, we randomly split $\mathcal{D}$ into $\mathcal{D}_{\text{train}}$ and $\mathcal{D}_{\text{es-cal}}$.
The data in $\mathcal{D}_{\text{train}}$ are utilized to train a neural network via stochastic gradient descent, storing the intermediate models $M_t$ after each $\tau$ epoch.
The approach is similar to those in Sections~\ref{sec:outlier}--\ref{sec:classification}, although now the output of a model $M_t$ applied to a sample with features $x$ is denoted by $\hat{\mu}_t(x)$ and is designed to approximate (possibly inaccurately) the conditional mean of the unknown data-generating distribution; i.e., $\hat{\mu}_t(x) \approx \E{Y \mid X=x}$. (Note that we will omit the superscript $t$ unless necessary to avoid ambiguity).
For any model $M_t$ and any $x \in \mathcal{X}$, $y \in \mathbb{R}$, define
\begin{align} \label{eq:loss-ces-reg}
  \mathcal{L}_{\text{es-cal}}^{+1}(M_t,x,y)
  & = \mathcal{L}_{\text{es-cal}}(M_t) + [y-\hat{\mu}_t(x)]^2.
\end{align}
Consider now the following optimization problem,
\begin{align} \label{eq:ces-model-reg}
  \hat{M}_{\text{ces}}(X_{n+1},y) = \argmin_{M_{t_j} \, : \, 1 \leq j \leq T} \mathcal{L}_{\text{es-cal}}^{+1}(M_{t_j},X_{n+1},y),
\end{align}
which can be solved simultaneously for all $y \in \mathbb{R}$ thanks to the amenable form of~\eqref{eq:loss-ces-reg}. In fact, each $\mathcal{L}_{\text{es-cal}}^{+1}(M_t,x,y)$ is a simple quadratic function of $y$; see the sketch in Figure~\ref{fig:quadratic_losses}.
This implies $\hat{M}_{\text{ces}}(X_{n+1},y)$ is a step function, whose parameters can be computed at cost $\mathcal{O}(T \log T)$ with an efficient divide-and-conquer algorithm designed to find the lower envelope of a family of parabolas \citep{devillers1995incremental,nielsen1998output}; see Appendix~\ref{app:lower-envelope}.


Therefore, $\hat{M}_{\text{ces}}(X_{n+1},y)$ has $L$ distinct steps, for some $L = \mathcal{O}(T \log T)$ that may depend on $X_{n+1}$, and it can be written as a function of $y$ as:
\begin{align} \label{eq:reg-step-func}
  \hat{M}_{\text{ces}}(X_{n+1},y) = \sum_{l=1}^{L} m_l(X_{n+1}) \I{y \in (k_{l-1}, k_{l}] },
\end{align}
where $m_l(X_{n+1}) \in [T]$ represents the best model selected within the interval $(k_{l-1}, k_{l}]$ such that $m_l(X_{n+1}) \neq m_{l-1}(X_{n+1})$ for all $l \in [L]$.
Above, $k_{1}\leq k_{2} \leq \dots \leq k_{L}$ denote the {\em knots} of $\hat{M}_{\text{ces}}(X_{n+1},y)$, which also depend on $X_{n+1}$ and are defined as the boundaries in the domain of $y$ between each consecutive pair of steps, with the understanding that $k_0 = -\infty$ and $k_{L+1} = +\infty$.
Then, for each step $l \in [L]$, let $\mathcal{B}_l$ indicate the interval $\mathcal{B}_l = (k_{l-1}, k_{l}]$ and, for all $i \in \mathcal{D}_{\text{es-cal}}$, evaluate the nonconformity score $\hat{S}_i(X_{n+1},\mathcal{B}_l)$ for observation $(X_i,Y_i)$ based on the regression model indicated by $m_l(X_{n+1})$; i.e.,
\begin{align} \label{eq:scores-reg}
  \hat{S}_i(X_{n+1},\mathcal{B}_l) = | Y_i - \hat{\mu}_{m_l(X_{n+1})}(X_{i})|.
\end{align}

Let $\hat{Q}_{1-\alpha}(X_{n+1},\mathcal{B}_l)$ denote the $\lceil (1-\alpha)(1+|\mathcal{D}_{\text{es-cal}}|) \rceil$-th smallest value among all nonconformity scores $\hat{S}_i(X_{n+1},\mathcal{B}_l)$, assuming for simplicity that there are no ties; otherwise, ties can be broken at random.
Then, define the interval $\hat{C}_{\alpha}(X_{n+1}, \mathcal{B}_l)$ as that obtained by applying the standard conformal prediction method with absolute residual scores based on the regression model $\hat{\mu}_{m_l(X_{n+1})}(X_{n+1})$:
\begin{align} \label{eq:reg-int-tmp}
  \hat{C}_{\alpha}(X_{n+1}, \mathcal{B}_l) = \hat{\mu}_{m_l(X_{n+1})}(X_{n+1}) \pm \hat{Q}_{1-\alpha}(X_{n+1},\mathcal{B}_l).
\end{align}
Finally, the prediction interval $\hat{C}_{\alpha}(X_{n+1})$ is given by:
\begin{align} \label{eq:reg-int}
  \hat{C}_{\alpha}(X_{n+1}) = \text{Convex}\left( \cup_{l=1}^{L} \{ \mathcal{B}_l \cap \hat{C}_{\alpha}(X_{n+1}, \mathcal{B}_l) \} \right),
\end{align}
where $\text{Convex}(\cdot)$ denotes the convex hull of a set.
This procedure is summarized in Algorithm~\ref{alg:reg} and it is guaranteed to produce prediction sets with valid marginal coverage.

\begin{algorithm}[H]
    \caption{Conformalized early stopping for regression}
    \label{alg:reg}
    \begin{algorithmic}[1]
        \STATE \textbf{Input}: Exchangeable data points $(X_{1},Y_{1}), \ldots, (X_{n},Y_{n})$ with outcomes $Y_i \in \mathbb{R}$.
        \STATE \textcolor{white}{\textbf{Input}:} Test point with features $X_{n+1}$. Desired coverage level $1-\alpha$.
        \STATE \textcolor{white}{\textbf{Input}:} Maximum number of training epochs $t^{\text{max}}$; storage period hyper-parameter $\tau$.
        \STATE \textcolor{white}{\textbf{Input}:} Regression model trainable via (stochastic) gradient descent.
        \STATE Randomly split the exchangeable data points into $\mathcal{D}_{\text{train}}$ and $\mathcal{D}_{\text{es-cal}}$.
        \STATE Train the regression model for $t^{\text{max}}$ epochs and save the intermediate models $M_{t_1} , \dots, M_{t_T}$.
        \STATE Evaluate $\hat{M}_{\text{ces}}(X_{n+1},y)$ as in \eqref{eq:reg-step-func}, using Algorithm~\ref{alg:envelope}.
        \STATE Partition the domain of $Y$ into $L$ intervals $\mathcal{B}_l$, for $l \in [L]$, based on knots of $\hat{M}_{\text{ces}}(X_{n+1},y)$.
        \FOR{$ l \in [L]$}
        \STATE Evaluate nonconformity scores $\hat{S}_i(X_{n+1},\mathcal{B}_l)$ for all $i \in \mathcal{D}_{\text{es-cal}}$ as in \eqref{eq:scores-reg}.
        \STATE Compute $\hat{Q}_{1-\alpha}(X_{n+1},\mathcal{B}_l)$: the $\lceil (1-\alpha)(1+|\mathcal{D}_{\text{es-cal}}|) \rceil$-th smallest value among $\hat{S}_i(X_{n+1},\mathcal{B}_l)$.
 \STATE Construct the interval $\hat{C}_{\alpha}(X_{n+1}, \mathcal{B}_l)$ according to~\eqref{eq:reg-int-tmp}.
        \ENDFOR
        \STATE \textbf{Output}: Prediction interval $\hat{C}_{\alpha}(X_{n+1})$ given as a function of $\{\hat{C}_{\alpha}(X_{n+1}, \mathcal{B}_l)\}_{l=1}^{L}$ by \eqref{eq:reg-int}.
    \end{algorithmic}
\end{algorithm}

\begin{theorem}\label{thm:reg}
Assume $(X_{1},Y_{1}), \ldots, (X_{n+1},Y_{n+1})$ are exchangeable, and let $\hat{C}_{\alpha}(X_{n+1})$ be the output of Algorithm~\ref{alg:reg}, as given by~\eqref{eq:reg-int}, for any given $\alpha \in (0,1)$.
Then, $\mathbb{P}[Y_{n+1} \in \hat{C}_{\alpha}(X_{n+1})] \geq 1-\alpha$.
\end{theorem}

The intuition behind the above method is as follows.
Each intermediate interval $\hat{C}_{\alpha}(X_{n+1}, \mathcal{B}_l)$, for $l \in [L]$, may be thought of as being computed by applying, under the null hypothesis that $Y_{n+1} \in \mathcal{B}_l$, the classification method from Section~\ref{sec:classification} for a discretized version of our problem based on the partition $\{\mathcal{B}_l\}_{l=1}^{L}$. Then, leveraging the classical duality between confidence intervals and p-values, it becomes clear that taking the intersection of $\mathcal{B}_l$ and $\hat{C}_{\alpha}(X_{n+1}, \mathcal{B}_l)$ essentially amounts to including the ``label'' $\mathcal{B}_l$ in the output prediction if the null hypothesis $Y_{n+1} \in \mathcal{B}_l$ cannot be rejected.
The purpose of the final convex hull operation is to generate a contiguous prediction interval, which is what we originally stated to seek.

One may intuitively be concerned that this method may output excessively wide prediction interval if the location of $\{\mathcal{B}_l \cap \hat{C}_\alpha(X_{n+1}, \mathcal{B}_l)\}$ is extremely large in absolute value.
However, our numerical experiments will demonstrate that, as long as the number of calibration data points is not too small, the selected models in general provide reasonably concentrated predictions around the true test response regardless of the placeholder value $y$. Therefore, the interval $\hat{C}_\alpha(X_{n+1}, \mathcal{B}_l)$ tends to be close to the true $y$ even if $\mathcal{B}_l$ is far away, in which case $\hat{C}_\alpha(X_{n+1}, \mathcal{B}_l) \cap \mathcal{B}_l = \emptyset$ does not expand the final prediction interval $\hat{C}_\alpha(X_{n+1})$.

Although it is unlikely, Algorithm~\ref{alg:reg} may sometimes produce an empty set, which is an uninformative and potentially confusing output. A simple solution consists of replacing any empty output with the naive conformal prediction interval computed by Algorithm~\ref{alg:naive-reg} in Appendix~\ref{app:naive-benchmarks}, which leverages an early-stopped model selected by looking at the original calibration data set without the test point.
This approach is outlined by Algorithm~\ref{alg:reg-noempty} in Appendix~\ref{app:reg-noempty}.
As the intervals given by Algorithm~\ref{alg:reg-noempty} always contain those output by Algorithm~\ref{alg:reg}, it follows that Algorithm~\ref{alg:reg-noempty} also enjoys guaranteed coverage; see Corollary~\ref{thm:reg-noempty}.

\subsection{CES for Quantile Regression} \label{sec:regression-cqr}

The CES method for regression described in Section~\ref{sec:regression} relies on classical nonconformity scores~\citep{vovk2005algorithmic,lei2016RegressionPS} that are not designed to deal efficiently with heteroscedastic data~\citep{romano2019conformalized,sesia2020comparison}.
However, the idea can be extended to accommodate other nonconformity scores, including those based on quantile regression~\citep{romano2019conformalized}, conditional distributions~\citep{izbicki2019flexible, chernozhukov2019distributional}, or conditional histograms~\citep{sesia2021conformal}.
The reason why we have so far focused on the classical absolute residual scores is that they are more intuitive to apply in conjunction with an early stopping criterion based on the squared-error loss. In this section, we extend CES to the conformalized quantile regression (CQR) method of \citet{romano2019conformalized}. A review of the CQR method is provided in Appendix~\ref{app:cqr_review}.

As in the previous section, consider a data set containing $n$ exchangeable observations $(X_i,Y_i)$, for $i \in \mathcal{D} = [n]$, and a test point $(X_{n+1}, Y_{n+1})$ with a latent label $Y_{n+1} \in \mathbb{R}$. 
We first randomly split $\mathcal{D}$ into two subsets, $\mathcal{D}_{\text{train}}$ and $\mathcal{D}_{\text{es-cal}}$. The data in $\mathcal{D}_{\text{train}}$ are utilized to train a neural network quantile regression model \cite{taylor2000quantile} by seeking to minimize the pinball loss instead of the squared error loss, for each target level $\beta=\beta_{\text{low}}$ and $\beta=\beta_{\text{high}}$ (e.g., $\beta_{\text{low}} = \alpha/2$ and $\beta_{\text{high}}=1-\alpha/2$). Note that the same neural network, with two separate output nodes, can be utilized to estimate conditional quantiles at two different levels; e.g., as in \citet{romano2019conformalized}.
For any $t \in [t^{\text{max}}]$, let $M_{\beta, t}$ denote the intermediate neural network model stored after $t$ epochs of stochastic gradient descent, following the same notation as in Section~\ref{sec:regression}.
For each target level $\beta$ and any $x \in \mathcal{X}$, let $\hat{q}_{\beta, t}(x)$ denote the approximate $\beta-$th conditional quantile of the unknown conditional distribution $\mathbb{P}(Y \mid X=x)$ estimated by $M_{\beta, t}$.

\begin{figure}[!htb]
    \centering
    \includegraphics[width=0.68\linewidth]{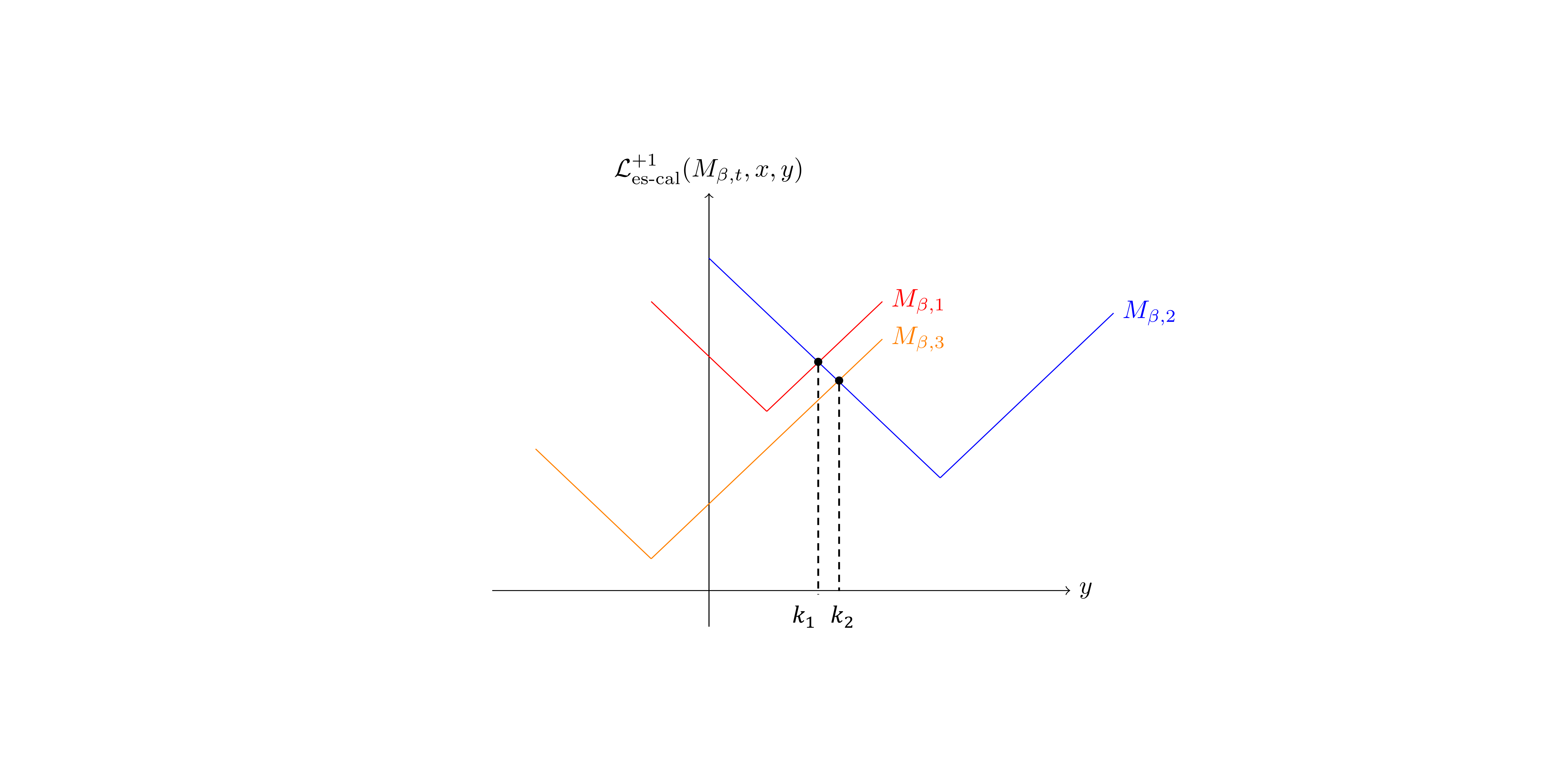}
    \vspace{-1.5em}
    \caption{Pinball loss functions on test-augmented hold-out data for three alternative regression models, $M_1, M_2$ and $M_3$, as a function of the place-holder outcome $y$ for the test point. The CES method utilizes the best model for each possible value of $y$, which is identified by the lower envelope of these three pinball loss functions. In this case, the lower envelope has a single finite knot at $k_{2}$.}
    \label{fig:pinball_losses}%
\end{figure}

Similarly to Section~\ref{sec:regression}, for any model $M_{\beta, t}$ and any $x \in \mathcal{X}$, $y \in \mathbb{R}$, define the augmented loss evaluated on the calibration data including also a dummy test point $(x,y)$ as:
\begin{align}\label{eq:loss-ces-reg-cqr}
  \begin{split}
    \mathcal{L}_{\text{es-cal}}^{+1}(M_{\beta, t},x,y)
    & = \mathcal{L}_{\text{es-cal}}(M_{\beta, t}) + \mathcal{L}(M_{\beta, t}, x, y) \\
    & = \sum_{i \in \mathcal{D}_{\text{es-cal}}} \rho_\beta(Y_i, \hat{q}_{\beta, t}(X_i)) + \rho_\beta(y, \hat{q}_{\beta, t}(x)),
  \end{split}
\end{align}
where $\rho_\beta$ denotes the pinball loss function defined in~\eqref{eq:reg-pinball_loss}.
For any model $M_{\beta, t}$, the augmented loss is equal to a constant plus a convex function of $y$, namely $\rho_\beta(y, \hat{q}_{\beta, t}(x))$.
Therefore, for any fixed $x$, the quantity in~\eqref{eq:loss-ces-reg-cqr} can be sketched as a function of $M_{\beta, t}$ and $y$ as shown in Figure~\ref{fig:pinball_losses}. This is analogous to Figure~\ref{fig:quadratic_losses} from Section~\ref{sec:regression}, with the difference that now the quadratic functions have been replaced by piece-wise linear ``pinball'' functions.

After pre-training and storing $T$ candidate models, namely $M_{\beta,t_1}, \ldots, M_{\beta,t_T}$ for some sub-sequence $(t_1, \ldots, t_T)$ of $[t^{\text{max}}]$, consider the following optimization problem,
\begin{align} \label{eq:ces-model-reg-cqr}
  \hat{M}_{\beta, \text{ces}}(X_{n+1},y) = \argmin_{M_{\beta, t_j} \, : \, 1 \leq j \leq T} \mathcal{L}_{\text{es-cal}}^{+1}(M_{\beta, t_j},X_{n+1},y).
\end{align}
This problem is equivalent to identifying the lower envelope of a family of shifted pinball loss functions, similarly to Section~\ref{sec:regression}; see Figure~\ref{fig:pinball_losses} for a schematic visualization.
Again, this  lower envelope can be found at computational cost $\mathcal{O}(T \log T)$, with the same divide-and-conquer algorithm described in Appendix~\ref{app:lower-envelope}.
In particular, $\hat{M}_{\beta, \text{ces}}(X_{n+1},y)$ is a step function with respect to $y$ with $L$ distinct steps, for some $L = \mathcal{O}(T \log T)$, and it can be written as:
\begin{align} \label{eq:reg-step-func-cqr}
 \hat{M}_{\beta, \text{ces}}(X_{n+1},y) = \sum_{l=1}^{L} m_{\beta, l}(X_{n+1}) \I{y \in (k^{\beta}_{l-1}, k^{\beta}_{l}] },
\end{align}
where $m_{\beta, l}(X_{n+1}) \in [T]$ represents the best model selected within the interval $(k^{\beta}_{l-1}, k^{\beta}_{l}]$ such that $m_{\beta,l}(X_{n+1}) \neq m_{\beta, l-1}(X_{n+1})$ for all $l \in [L]$.
Above, $k^{\beta}_{1}\leq k^{\beta}_{2} \leq \dots \leq k^{\beta}_{L}$ denote the {\em knots} of $\hat{M}_{\beta, \text{ces}}(X_{n+1},y)$, which also depend on $X_{n+1}$ and are defined as the boundaries in the domain of $y$ between each consecutive pair of steps, with the understanding that $k^{\beta}_0 = -\infty$ and $k^{\beta}_{L+1} = +\infty$; see Figure~\ref{fig:pinball_losses} for a schematic visualization.

After computing $\hat{M}_{\beta, \text{ces}}(X_{n+1},y)$ in~\eqref{eq:ces-model-reg-cqr} for both  $\beta_{\text{low}}$ and $\beta_{\text{high}}$, we concatenate the respective knots $k_1^{\text{low}}, \dots, k^{\text{low}}_{L_1}, k_1^{\text{high}}, \dots, k_{L_2}^{\text{high}}$ and sort them into $k_1 \leq k_2 \leq k_{L_1+L_2}$, so that within each interval $\mathcal{B}_l = (k_{l-1}, k_{l}]$ for step $l \in [L_1 + L_2]$, there exist exactly one best model for $\beta_{\text{low}}$ and exactly one best model for $\beta_{\text{high}}$. 
Then, for each interval $\mathcal{B}_l = (k_{l-1}, k_{l}]$ associated with step $l \in[L_1 + L_2]$, evaluate the nonconformity score $\hat{E}_i(X_{n+1}, \mathcal{B}_l)$ for all $i \in \mathcal{D}_{\text{es-cal}}$, based on the regression model indicated by $m_{\beta_{\text{low}},l}(X_{n+1})$ and $m_{\beta_{\text{high}},l}(X_{n+1})$; i.e.,
\begin{align} \label{eq:scores-reg-cqr}
  \hat{E}_i(X_{n+1}, \mathcal{B}_l) = \max\left\{\hat{q}_{m_{\beta_{\text{low}},l}(X_{n+1})}(X_i) - Y_i, Y_i - \hat{q}_{m_{\beta_{\text{high}},l}(X_{n+1})}(X_i)\right\}.
\end{align}

Let $\hat{Q}_{1-\alpha}(X_{n+1},\mathcal{B}_l)$ denote the $\lceil (1-\alpha)(1+|\mathcal{D}_{\text{es-cal}}|) \rceil$-th smallest value among all nonconformity scores $\hat{E}_i(X_{n+1},\mathcal{B}_l)$, assuming for simplicity that there are no ties; otherwise, ties can be broken at random. 
Then, define the interval $\hat{C}_{\alpha}(X_{n+1}, \mathcal{B}_l)$ as that obtained by applying the conformal prediction method of \citet{romano2019conformalized} with nonconformity scores~\eqref{eq:scores-reg-cqr} based on the estimated conditional quantiles $\hat{q}_{m_{\beta_{\text{low}},l}(X_{n+1})}(X_{n+1})$ and $\hat{q}_{m_{\beta_{\text{high}},l}(X_{n+1})}(X_{n+1})$; that is,
\begin{align} \label{eq:reg-int-tmp-cqr}
  \hat{C}_{\alpha}(X_{n+1}, \mathcal{B}_l) = [\hat{q}_{m_{\beta_{\text{low}},l}(X_{n+1})}(X_{n+1}) - \hat{Q}_{1-\alpha}(X_{n+1},\mathcal{B}_l), \hat{q}_{m_{\beta_{\text{high}},l}(X_{n+1})}(X_{n+1}) + \hat{Q}_{1-\alpha}(X_{n+1},\mathcal{B}_l)].
\end{align}
Finally, the output prediction interval $\hat{C}_{\alpha}(X_{n+1})$ is given by:
\begin{align} \label{eq:reg-int-cqr}
  \hat{C}_{\alpha}(X_{n+1}) = \text{Convex}\left( \cup_{l=1}^{L} \{ \mathcal{B}_l \cap \hat{C}_{\alpha}(X_{n+1}, \mathcal{B}_l) \} \right).
\end{align}
This procedure, summarized in Algorithm~\ref{alg:reg-cqr}, guarantees valid marginal coverage.

\begin{algorithm}[H]
    \caption{Conformalized early stopping for quantile regression}
    \label{alg:reg-cqr}
    \begin{algorithmic}[1]
        \STATE \textbf{Input}: Exchangeable data points $(X_{1},Y_{1}), \ldots, (X_{n},Y_{n})$ with outcomes $Y_i \in \mathbb{R}$.
        \STATE \textcolor{white}{\textbf{Input}:} Test point with features $X_{n+1}$. Desired coverage level $1-\alpha$.
        \STATE \textcolor{white}{\textbf{Input}:} Maximum number of training epochs $t^{\text{max}}$; storage period hyper-parameter $\tau$.
        \STATE \textcolor{white}{\textbf{Input}:} Trainable quantile regression model with target quantiles [$\beta_{\text{low}}$, $\beta_{\text{high}}$].
        \STATE Randomly split the exchangeable data points into $\mathcal{D}_{\text{train}}$ and $\mathcal{D}_{\text{es-cal}}$.
        \STATE Train for $t^{\text{max}}$ epochs and save the intermediate models $M_{\beta_{\text{low}}, t_1} , \dots, M_{\beta_{\text{low}}, t_T}$, $M_{\beta_{\text{high}}, t_1} , \dots, M_{\beta_{\text{high}}, t_T}$.
        \STATE Evaluate $\hat{M}_{\beta_{\text{low}},\text{ces}}(X_{n+1},y)$ and $\hat{M}_{\beta_{\text{high}},\text{ces}}(X_{n+1},y)$ as in \eqref{eq:reg-step-func-cqr}, using Algorithm~\ref{alg:envelope-cqr}.
        \STATE Partition the domain of $Y$ into $L_1+L_2$ intervals $\mathcal{B}_l$, for $l \in [L_1+L_2]$, based on the knots of $\hat{M}_{\beta_{\text{low}},\text{ces}}(X_{n+1},y)$ and $\hat{M}_{\beta_{\text{high}},\text{ces}}(X_{n+1},y)$.
        \FOR{$ l \in [L_1+ L_2]$}
        \STATE Evaluate nonconformity scores $\hat{E}_i(X_{n+1},\mathcal{B}_l)$ for all $i \in \mathcal{D}_{\text{es-cal}}$ as in \eqref{eq:scores-reg-cqr}.
        \STATE Compute $\hat{Q}_{1-\alpha}(X_{n+1},\mathcal{B}_l)$: the $\lceil (1-\alpha)(1+|\mathcal{D}_{\text{es-cal}}|) \rceil$-th smallest value among $\hat{E}_i(X_{n+1},\mathcal{B}_l)$.
 \STATE Construct the interval $\hat{C}_{\alpha}(X_{n+1}, \mathcal{B}_l)$ according to~\eqref{eq:reg-int-tmp-cqr}.
        \ENDFOR
        \STATE \textbf{Output}: Prediction interval $\hat{C}_{\alpha}(X_{n+1})$ given as a function of $\{\hat{C}_{\alpha}(X_{n+1}, \mathcal{B}_l)\}_{l=1}^{L}$ by \eqref{eq:reg-int-cqr}.
    \end{algorithmic}
\end{algorithm}

\begin{theorem} \label{thm:reg-cqr}
Assume $(X_{1},Y_{1}), \ldots, (X_{n+1},Y_{n+1})$ are exchangeable, and let $\hat{C}_{\alpha}(X_{n+1})$ be the output of Algorithm~\ref{alg:reg-cqr}, as given by~\eqref{eq:reg-int-cqr}, for any given $\alpha \in (0,1)$.
Then, $\mathbb{P}[Y_{n+1} \in \hat{C}_{\alpha}(X_{n+1})] \geq 1-\alpha$.
\end{theorem}

Similarly to Section~\ref{sec:regression}, it is possible (although unlikely) that Algorithm~\ref{alg:reg-cqr} may sometimes produce an empty prediction set.
Therefore, we present Algorithm~\ref{alg:reg-noempty-cqr} in Appendix~\ref{app:reg-noempty-quantile}, which extends Algorithm~\ref{alg:reg-cqr} in such a way as to explicitly avoid returning empty prediction intervals.
As the intervals given by Algorithm~\ref{alg:reg-noempty-cqr} always contain those output by Algorithm~\ref{alg:reg-cqr}, it follows from Theorem~\ref{thm:reg-cqr} that Algorithm~\ref{alg:reg-noempty-cqr} also enjoys guaranteed coverage; see Corollary~\ref{thm:reg-noempty-cqr}.

\subsection{Implementation Details and Computational Cost}

Beyond the cost of training the neural network (which is relatively expensive but does not need to be repeated for different test points) and the storage cost associated with saving the candidate models (which we have argued to be feasible in many applications), CES is quite computationally efficient.
Firstly, CES treats all test points individually and could process them in parallel, although many operations do not need to be repeated. In particular, one can recycle the evaluation of the calibration loss across different test points; e.g., see~\eqref{eq:loss-ces-class}. 
Thus, the model selection component can be easily implemented at cost $\mathcal{O}((n_{\text{es-cal}} + n_{\text{test}}) \cdot T + n_{\text{test}} \cdot n_{\text{es-cal}})$ for classification (of which outlier detection is a special case) and $\mathcal{O}((n_{\text{es-cal}} + n_{\text{test}}) \cdot T \cdot \log T + n_{\text{test}} \cdot n_{\text{es-cal}})$ for regression, where $n_{\text{es-cal}} = |\mathcal{D}_{\text{es-cal}}|$ and $T$ is the number of candidate models. Note that the $T \cdot \log T$ dependence in the regression setting comes from the divide-and-conquer algorithm explained in Appendix~\ref{app:lower-envelope}.

It is possible that the cost of CES may become a barrier in some applications, particularly if $T$ is very large, despite the slightly more than linear scaling. Hence, we recommend employing moderate values of $T$ (e.g., 100 or 1000).

\section{Numerical Experiments} \label{sec:numerical_results}


\subsection{Outlier Detection} \label{sec:num_od}

The use of CES for outlier detection is demonstrated using the {\em CIFAR10} data set \citep{cifar10}, a collection of 60,000 32-by-32 RGB images from 10 classes including common objects and animals.
A convolutional neural network with ReLU activation functions is trained on a subset of the data to minimize the cross-entropy loss. The maximum number of epochs is set to be equal to 50.
The trained classification model is then utilized to compute conformity scores for outlier detection with the convention that cats are inliers and the other classes are outliers.
In particular, a nonconformity score for each $Z_{n+1}$ is defined as 1 minus the output of the soft-max layer corresponding to the label ``cat''.
This can be interpreted as an estimated probability of $Z_{n+1}$ being an outlier.
After translating these scores into a conformal p-value $\hat{u}_0(Z_{n+1})$, the null hypothesis that $Z_{n+1}$ is a cat is rejected if $\hat{u}_0(Z_{n+1}) \leq \alpha = 0.1$.

The total number of samples utilized for training, early stopping, and conformal calibration is varied between 500 and 2000.
In each case, CES is applied using 75\% of the samples for training and 25\% for early stopping and calibration. Note that the calibration step only utilizes inliers, while the other data subsets also contain outliers.
The empirical performance of CES is measured in terms of the probability of falsely rejecting a true null hypothesis---the false positive rate (FPR)---and the probability of correctly rejecting a false null hypothesis---the true positive rate (TPR).
The CES method is compared to three benchmarks. The first benchmark is naive early stopping with the best ({\em hybrid}) theoretical correction for the nominal coverage level described in Appendix~\ref{app:naive-analysis}. The second benchmark is early stopping based on data splitting, which utilizes 50\% of the available samples for training, 25\% for early stopping, and 25\% for calibration.
The third benchmark is full training without early stopping, which simply selects the model obtained after the last epoch.
The test set consists of 100 independent test images, half of which are outliers.
All results are averaged over 100 trials based on independent data subsets.

\begin{figure}[!htb]
    \centering
    \includegraphics[width=0.65\linewidth]{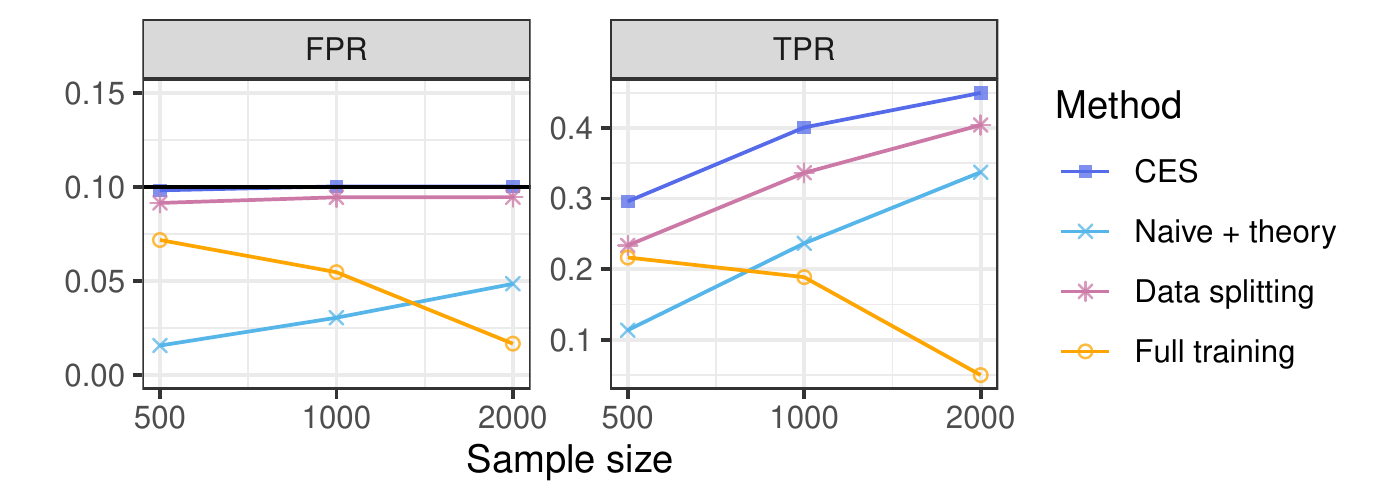}\vspace{-0.5cm}
    \caption{Average performance, as a function of the sample size, of conformal inferences for outlier detection based on neural networks trained and calibrated with different methods, on the {\em CIFAR10} data~\citep{cifar10}. Ideally, the TPR should be as large as possible while maintaining the FPR below 0.1. See Table~\ref{tab:exp_oc} for more detailed results and standard errors.}
    \label{fig:exp_oc}
\end{figure}

Figure~\ref{fig:exp_oc} summarizes the performance of the four methods as a function of the total sample size; see Table~\ref{tab:exp_oc} in Appendix~\ref{app:numerical-results} for the corresponding standard errors. All methods control the FPR below 10\%, as expected, but CES achieves the highest TPR.
The increased power of CES compared to data splitting is not surprising, as the latter relies on a less accurate model trained on less data.
By contrast, the naive benchmark trains a model more similar to that of CES, but its TPR is not as high because the theoretical correction for the naive conformal p-values is overly pessimistic.
Finally, full training is the least powerful competitor for large sample sizes because its underlying model becomes more and more overconfident as the training set grows.
Note that Table~\ref{tab:exp_oc} also includes the results obtained with the naive benchmark detailed in Appendix~\ref{app:naive-benchmarks}, applied without the theoretical correction necessary to guarantee marginal coverage. Remarkably, the results show that the naive benchmark performs similarly to the CES method, even though only the latter has the advantage of enjoying rigorous finite-sample guarantees.

\subsection{Multi-class Classification} \label{sec:num_mc}

The same {\em CIFAR10} data~\citep{cifar10} are utilized to demonstrate the performance of CES for a 10-class classification task.
These experiments are conducted similarly to those in Section~\ref{sec:num_od}, with the difference that now the soft-max output of the convolutional neural network is translated into conformal prediction sets, as explained in Appendix~\ref{app:class-scores}, instead of conformal p-values.
The CES method is compared to the same three benchmarks from Section~\ref{sec:num_od}. All prediction sets guarantee 90\% marginal coverage, and their performances are evaluated based on average cardinality.

\begin{figure}[!htb]
    \centering
    \includegraphics[width=0.65\linewidth]{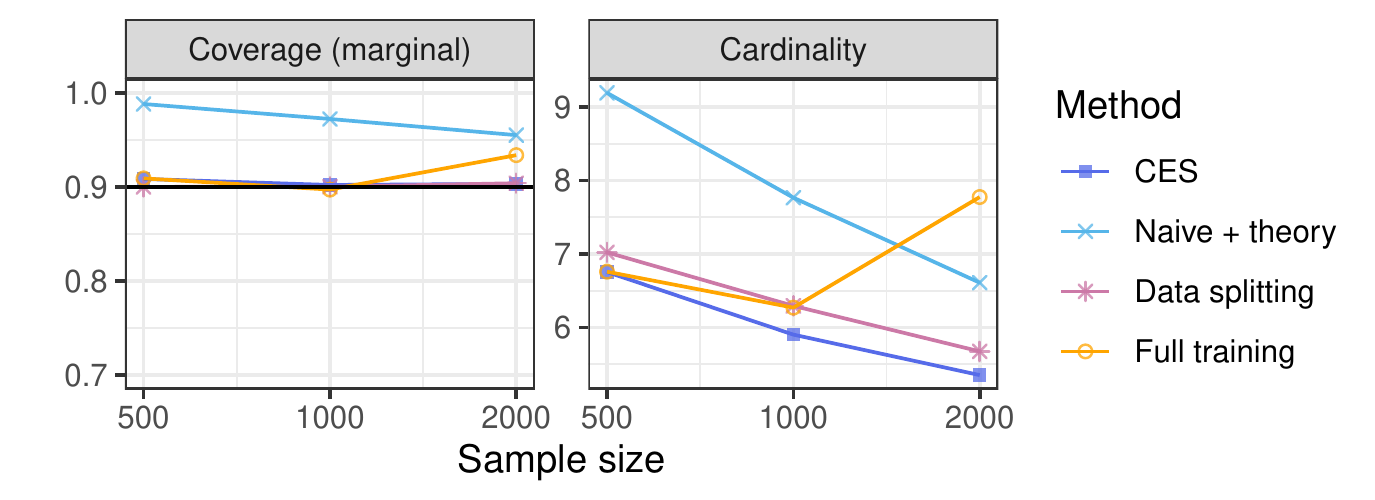}\vspace{-0.5cm}
    \caption{Average performance, as a function of the sample size, of conformal prediction sets for multi-class classification based on neural networks trained and calibrated with different methods, on the {\em CIFAR10} data~\citep{cifar10}. Ideally, the coverage should be close to 90\% and the cardinality should be small. See Table~\ref{tab:exp_mc} for more detailed results and standard errors.}
    \label{fig:exp_mc}
\end{figure}

Figure \ref{fig:exp_mc} summarizes the results averaged over 100 independent realizations of these experiments, while Table~\ref{tab:exp_mc} in Appendix~\ref{app:numerical-results} reports on the corresponding standard errors. 
While all approaches always achieve the nominal coverage level, the CES method is able to do so with the smallest, and hence most informative, prediction sets.
As before, the more disappointing performance of the data splitting benchmark can be explained by the more limited amount of data available for training, that of the naive benchmark by the excessive conservativeness of its theoretical correction, and that of the full training benchmark by overfitting.
Table~\ref{tab:exp_mc} also includes the results obtained with the naive benchmark without the theoretical correction, which again performs similarly to CES.

\subsection{Regression}

We now apply the CES method from Section~\ref{sec:regression} to the following 3 public-domain regression data sets from the UCI Machine Learning repository \citep{Pinar2012}: physicochemical properties of protein tertiary structure ({\em bio})~\citep{data-bio}, hourly and daily counts of rental bikes ({\em bike})~\citep{data-bike}, and concrete compressive strength ({\em concrete})~\citep{data-concrete}. These data sets were previously also considered by \citet{romano2019conformalized}, to which we refer for further details.
As in the previous sections, we compare CES to the usual three benchmarks: naive early stopping with the {\em hybrid} theoretical correction for the nominal coverage level, early stopping based on data splitting, and full model training without early stopping.
All methods utilize the same neural network with two hidden layers of width 128 and ReLU activation functions, trained for up to 1000 epochs.
The models are calibrated in such a way as to produce conformal prediction sets with guaranteed 90\% marginal coverage for a test set of 100 independent data points.
The total sample size available for training, early stopping and calibration is varied between 200 and 2000.
These data are allocated for specific training, early-stopping, and calibration operations as in Sections~\ref{sec:num_od}--\ref{sec:num_mc}.
The performance of each method is measured in terms of marginal coverage, worst-slab conditional coverage~\citep{cauchois2020knowing}---estimated as described in \citet{sesia2020comparison}---and average width of the prediction intervals. All results are averaged over 100 independent experiments, each based on a different random sample from the original raw data sets.

Figure~\ref{fig:exp_regression_bio} summarizes the performance of the four alternative methods on the {\em bio} data, as a function of the total sample size;  see Table~\ref{fig:exp_regression_bike} in Appendix~\ref{app:numerical-results} for the corresponding standard errors. These results show that all methods reach 90\% marginal coverage in practice, as anticipated by the mathematical guarantees, although the theoretical correction for the naive early stopping method appears to be overly conservative. The CES method clearly performs best, in the sense that it leads to the shortest prediction intervals while also achieving approximately valid conditional coverage. By contrast, the conformal prediction intervals obtained without early stopping have significantly lower conditional coverage, which is consistent with the prior intuition that fully trained neural networks can sometimes suffer from overfitting.
More detailed results from these experiments can be found in Table~\ref{tab:exp_regression_bio} in Appendix~\ref{app:numerical-results}.
Analogous results corresponding to the {\em bike} and {\em concrete} data sets can be found in Figures~\ref{fig:exp_regression_bike}--\ref{fig:exp_regression_concrete} and Tables~\ref{tab:exp_regression_bike}--\ref{tab:exp_regression_concrete} in Appendix~\ref{app:numerical-results}.
Tables~\ref{tab:exp_regression_bio}--\ref{tab:exp_regression_concrete} also include the results obtained with the naive benchmark applied without the necessary theoretical correction, which performs similarly to CES.

Finally, it must be noted that the widths of the prediction intervals output by the CES method in these experiments are very similar to those of the corresponding intervals produced by naively applying early stopping without data splitting and without the theoretical correction described in Appendix~\ref{app:naive-benchmarks}.
This naive approach was not taken as a benchmark because it does not guarantee valid coverage, unlike the other methods. Nonetheless, it is interesting to note that the rigorous theoretical properties of the CES method do not come at the expense of a significant loss of power compared to this very aggressive heuristic, and in this sense, one may say that the conformal inferences computed by CES are ``almost free''.

\subsection{Quantile Regression}

We apply the CES quantile regression method from Section~\ref{sec:regression-cqr} to the following publicly available and commonly investigated regression data sets from the UCI Machine Learning repository \cite{Pinar2012}: medical expenditure panel survey number 21 ({\em MEPS\_21}) \cite{meps_21}; blog feedback ({\em blog\_data}) \cite{blog_data}; Tennessee’s student teacher achievement ratio ({\em STAR}) \cite{star}; community and crimes ({\em community}) \cite{community}; physicochemical properties of protein tertiary structure ({\em bio})~\cite{data-bio}; house sales in King County ({\em homes}) \cite{homes}; and hourly and daily counts of rental bikes ({\em bike})~\cite{data-bike}. These data sets were previously also considered by \citet{romano2019conformalized}.

As in the previous sections, we compare CES to the usual three benchmarks, now implemented based on quantile regression: naive early stopping with the {\em hybrid} theoretical correction for the nominal coverage level, early stopping based on data splitting and full model training without early stopping. We follow the same model architecture and data preprocessing steps as in \citet{romano2019conformalized}. To be specific, the input features are standardized to have zero mean and unit variance, and the response values are rescaled by diving the absolute mean of the training responses. All methods utilize the same neural network with three hidden layers and ReLU activation functions between layers, trained for up to 2000 epochs. The parameters are trained minimizing the pinball loss function (see Appendix~\ref{app:cqr_review}) with Adam optimizer \cite{kingma2014adam}, minibatches of size 25, 0 weight decay and dropout, and fixed learning rate (0.001 for {\em STAR, homes, bike, and bio}, 0.0001 for {\em community}, and 0.00005 for {\em MEPS\_21 and blog\_data}). 

The models are calibrated in such a way as to produce conformal prediction sets with guaranteed 90\% marginal coverage for a test set of 1000 independent data points. The total sample size available for training, early stopping and calibration is varied between 200 and 2000 (200 and 1000 for small data sets such as {\em community} and {\em STAR}).
These data are allocated for specific training, early-stopping, and calibration operations as in Sections~\ref{sec:num_od}--\ref{sec:num_mc}.
Again, the performance of each method is measured in terms of marginal coverage, worst-slab conditional coverage~\cite{cauchois2020knowing}, and average width of the prediction intervals. All results are averaged over 25 independent experiments, each based on a different random sample from the original raw data sets.

\begin{figure}[!htb]
    \centering
    \includegraphics[width=0.8\linewidth]{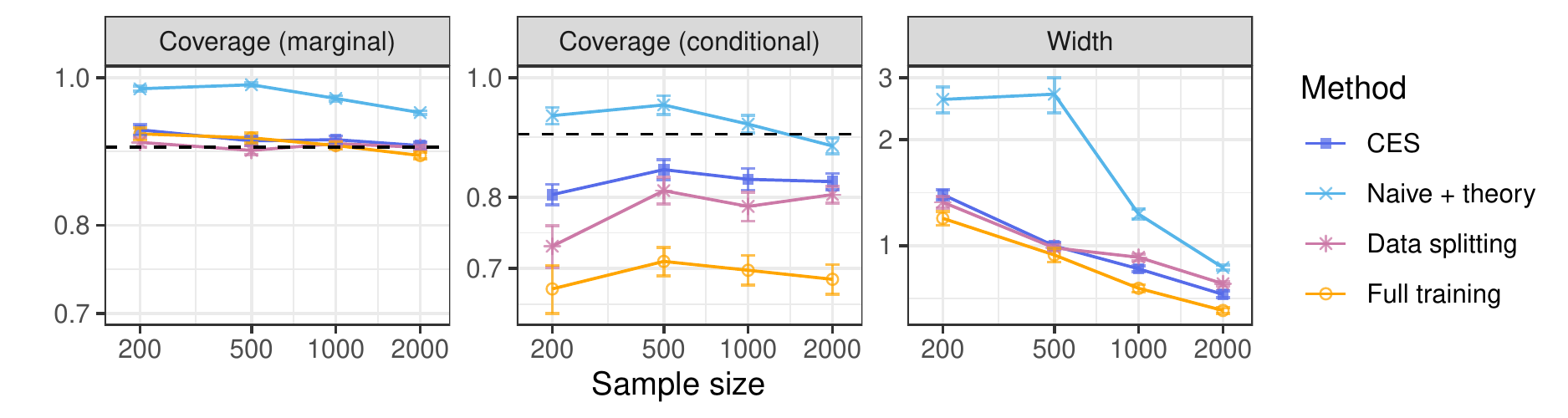}\vspace{-0.5cm}
    \caption{Average performance, as a function of the sample size, of conformal prediction sets for quantile regression based on neural networks trained and calibrated with different methods, on the {\em homes} data~\cite{homes}. The marginal coverage is theoretically guaranteed to be above 90\%. Ideally, the conditional coverage should high and the prediction intervals should be tight. See Table~\ref{tab:exp_qt_regression_homes_tab} for additional details and standard errors.}
    \label{fig:exp_qt_regression_homes}
\end{figure}

Figure~\ref{fig:exp_qt_regression_homes} summarizes the performance of the four alternative methods on the {\em homes} data, as a function of the total sample size; The error bar corresponding to standard errors are plotted around each data point. These results show that all methods reach 90\% marginal coverage in practice, as anticipated by the mathematical guarantees, although the theoretical correction for the naive early stopping method appears to be overly conservative. Full training, though producing the smallest prediction bands, has very low conditional coverage, which indicates that fully trained neural network models can suffer from overfitting and therefore is not appealing. Data splitting method beats full training as it gives higher approximated conditional coverage, and CES further beats data splitting in terms of conditional coverage, meanwhile producing prediction intervals of similar length as data splitting. These patterns hold true in general for additional data sets, as illustrated by Figures~\ref{fig:exp_qt_regression_community}--\ref{fig:exp_qt_regression_bike} and by Tables~\ref{tab:exp_qt_regression_homes_tab}--\ref{tab:exp_qt_regression_bike_tab}.
Tables~\ref{tab:exp_qt_regression_homes_tab}--\ref{tab:exp_qt_regression_bike_tab} also include the results obtained with the naive benchmark applied without the necessary theoretical correction, which performs similarly to CES.

\section{Discussion} \label{sec:discussion}

This paper has focused on early stopping and conformal calibration because these are two popular techniques, respectively designed to mitigate overfitting and reduce overconfidence, that were previously combined without much thought. However, the relevance of our methodology extends well beyond the problem considered in this paper.
In fact, related ideas have already been utilized in the context of outlier detection to tune hyper-parameters and select the most promising candidate from an arbitrary toolbox of machine learning models \citep{Liang_2022_integrative_p_val}.
The techniques developed in this paper also allow one to calibrate, without further data splits, the most promising model selected in a data-driven way from an arbitrary machine learning toolbox in the context of multi-class classification and regression.

As mentioned in Section~\ref{sec:naive_benchmark} and detailed in Appendix~\ref{app:naive-benchmarks}, the naive benchmark that uses the same hold-out data twice, both for standard early stopping and standard conformal calibration, is not theoretically valid without conservative corrections. Nonetheless, our numerical experiments have shown that this naive approach often performs similarly to CES in practice.
Of course, the naive benchmark may sometimes fail, and thus we would advise practitioners to apply the theoretically principled CES whenever its additional memory costs are not prohibitive. However, the empirical evidence suggests the naive benchmark may not be a completely unreasonable heuristic when CES is not applicable.

Software implementing the algorithms and data experiments are available online at \url{https://github.com/ZiyiLiang/Conformalized_early_stopping}.

\subsection*{Acknowledgements}
The authors thank the Center for Advanced Research Computing at the University of Southern California for providing computing resources to carry out numerical experiments.
The authors are also grateful to three anonymous reviewers for their insightful comments and suggestions.
M.~S. and Y.~Z.~are supported by NSF grant DMS 2210637.
M.~S.~is also supported by an Amazon Research Award. 


\printbibliography                                                                                                
\newpage

\appendix
\renewcommand{\thesection}{A\arabic{section}}
\renewcommand{\theequation}{A\arabic{equation}}
\renewcommand{\thetheorem}{A\arabic{theorem}}
\renewcommand{\thecorollary}{A\arabic{corollary}}
\renewcommand{\theproposition}{A\arabic{proposition}}
\renewcommand{\thelemma}{A\arabic{lemma}}
\renewcommand{\thetable}{A\arabic{table}}
\renewcommand{\thefigure}{A\arabic{figure}}
\renewcommand{\thealgorithm}{A\arabic{algorithm}}

\section{Naive Early Stopping Benchmarks} \label{app:naive-benchmarks}

\subsection{Detailed Implementation of the Naive Benchmarks}  \label{app:naive-benchmarks-details}

We detail here the implementation of the naive benchmark discussed in Section~\ref{sec:naive_benchmark}. This approach can serve as an informative benchmark and it becomes useful in Appendix~\ref{app:reg-noempty} to extend our rigorous conformalized early stopping method for regression problems in such as way as to explicitly avoid returning empty prediction intervals.
For completeness, we present the implementation of the naive benchmark separately for outlier detection, multi-class classification, and regression, respectively in Algorithms~\ref{alg:naive-one}, \ref{alg:naive-multi} and~\ref{alg:naive-reg}.
Note that Algorithm~\ref{alg:naive-multi} also allows for the possibility of computing prediction sets seeking (approximate) marginal coverage instead of (approximate) label-conditional coverage for multi-class classification problems; see Appendix~\ref{app:class-marg} for further details on multi-class classification with marginal coverage.

\begin{algorithm}[H]
    \caption{Naive conformal outlier detection benchmark with greedy early stopping}
    \label{alg:naive-one}
    \begin{algorithmic}[1]
        \STATE \textbf{Input}: Exchangeable data points $Z_1, \ldots, Z_n$; test point $Z_{n+1}$.
        \STATE \textcolor{white}{\textbf{Input}:} Maximum number of training epochs $t^{\text{max}}$; storage period hyper-parameter $\tau$.
        \STATE \textcolor{white}{\textbf{Input}:} One-class classifier trainable via (stochastic) gradient descent.
        \STATE Randomly split the exchangeable data points into $\mathcal{D}_{\text{train}}$ and $\mathcal{D}_{\text{es-cal}}$.
        \STATE Train the one-class classifier for $t^{\text{max}}$ epochs and save the intermediate models $M_{t_1} , \dots, M_{t_T}$.
        \STATE Pick the most promising model $t^* \in [T]$ minimizing $\mathcal{L}_{\text{es-cal}}(M_t)$ in \eqref{eq:loss-ces}, based on $\mathcal{D}_{\text{es-cal}}$.
        \STATE Compute nonconformity scores $\hat{S}_i(Z_{n+1})$ for all $i \in \mathcal{D}_{\text{es-cal}} \cup \{n+1\}$ using model $t^*$.
        \STATE \textbf{Output}: Naive conformal p-value $\hat{u}_0^{\textup{naive}}(Z_{n+1})$ given by \eqref{eq:conformal_pval}.
    \end{algorithmic}
\end{algorithm}

\begin{algorithm}[H]
    \caption{Naive conformal multi-class classification benchmark with greedy early stopping}
    \label{alg:naive-multi}
    \begin{algorithmic}[1]
        \STATE \textbf{Input}: Exchangeable data points $(X_{1},Y_{1}), \ldots, (X_{n},Y_{n})$ with labels $Y_i \in [K]$.
        \STATE \textcolor{white}{\textbf{Input}:} Test point with features $X_{n+1}$. Desired coverage level $1-\alpha$.
        \STATE \textcolor{white}{\textbf{Input}:} Maximum number of training epochs $t^{\text{max}}$; storage period hyper-parameter $\tau$.
        \STATE \textcolor{white}{\textbf{Input}:} $K$-class classifier trainable via (stochastic) gradient descent.
        \STATE Randomly split the exchangeable data points into $\mathcal{D}_{\text{train}}$ and $\mathcal{D}_{\text{es-cal}}$.
        \STATE Train the $K$-class classifier for $t^{\text{max}}$ epochs and save the intermediate models $M_{t_1} , \dots, M_{t_T}$.
        \STATE Pick the most promising model $t^* \in [T]$ minimizing $\mathcal{L}_{\text{es-cal}}(M_t)$ in \eqref{eq:loss-ces-class}, based on $\mathcal{D}_{\text{es-cal}}$.
        \FOR{$ y \in [K]$}
        \IF{Label-conditional coverage is desired}
        \STATE Define $\mathcal{D}^y_{\text{es-cal}} = \{i \in \mathcal{D}_{\text{es-cal}} : Y_i = y \}$.
        \STATE Compute scores $\hat{S}_i^y(X_{n+1})$ for all $i \in \mathcal{D}^y_{\text{es-cal}} \cup \{n+1\}$ using model $t^*$; see Appendix~\ref{app:class-scores}.
        \STATE Compute the naive conformal p-value $\hat{u}^{\textup{naive}}_y(X_{n+1})$ according to \eqref{eq:conformal_pval-class}.
        \ELSE
        \STATE Compute scores $\hat{S}_i^y(X_{n+1})$ for all $i \in \mathcal{D}_{\text{es-cal}} \cup \{n+1\}$ using model $t^*$; see Appendix~\ref{app:class-scores}.
        \STATE Compute the naive conformal p-value $\hat{u}^{\textup{naive}}_y(X_{n+1})$ according to
        \begin{align*}
          \hat{u}^{\textup{naive}}_y(X_{n+1}) = \frac{1 + |i \in \mathcal{D}_{\text{es-cal}}: \hat{S}_{i}^y(X_{n+1}) \leq \hat{S}_{n+1}^y(X_{n+1})|}{1+|\mathcal{D}_{\text{es-cal}}|}.
        \end{align*}
        \ENDIF
        \ENDFOR

        \STATE \textbf{Output}: Naive prediction set $\hat{C}^{\textup{naive}}_{\alpha}(X_{n+1})$ given by \eqref{eq:pred-set-class}.
    \end{algorithmic}
\end{algorithm}

\begin{algorithm}[H]
    \caption{Naive conformal regression benchmark with greedy early stopping}
    \label{alg:naive-reg}
    \begin{algorithmic}[1]
        \STATE \textbf{Input}: Exchangeable data points $(X_{1},Y_{1}), \ldots, (X_{n},Y_{n})$ with outcomes $Y_i \in \mathbb{R}$.
        \STATE \textcolor{white}{\textbf{Input}:} Test point with features $X_{n+1}$. Desired coverage level $1-\alpha$.
        \STATE \textcolor{white}{\textbf{Input}:} Maximum number of training epochs $t^{\text{max}}$; storage period hyper-parameter $\tau$.
        \STATE \textcolor{white}{\textbf{Input}:} Regression model trainable via (stochastic) gradient descent.
        \STATE Randomly split the exchangeable data points into $\mathcal{D}_{\text{train}}$ and $\mathcal{D}_{\text{es-cal}}$.
        \STATE Train the regression model for $t^{\text{max}}$ epochs and save the intermediate models $M_{t_1} , \dots, M_{t_T}$.
        \STATE Pick the most promising model $t^* \in [T]$ minimizing $\mathcal{L}_{\text{es-cal}}(M_t)$ in \eqref{eq:loss-ces-reg}.
        \STATE Evaluate nonconformity scores $\hat{S}_i(X_{n+1}) = | Y_i - \hat{\mu}_{t^*}(X_{i})|$ for all $i \in \mathcal{D}_{\text{es-cal}}$.
        \STATE Compute $\hat{Q}_{1-\alpha}(X_{n+1}) = \lceil (1-\alpha)(1+|\mathcal{D}_{\text{es-cal}}|) \rceil\text{-th smallest value in }
        \hat{S}_i(X_{n+1})$ for $i \in \mathcal{D}_{\text{es-cal}}$.
        \STATE \textbf{Output}: Prediction interval $\hat{C}^{\text{naive}}_{\alpha}(X_{n+1}) = \hat{\mu}_{t^*}(X_{n+1}) \pm \hat{Q}_{1-\alpha}(X_{n+1})$.
    \end{algorithmic}
\end{algorithm}

\subsection{Theoretical Analysis of the Naive Benchmark} \label{app:naive-analysis}

Although the naive benchmarks described above often perform similarly to CES in practice, they do not enjoy the same desirable theoretical guarantees.
Nonetheless, we can study their behaviour in sufficient detail as to prove that their inferences are too far from being valid.
Unfortunately, as demonstrated in Section~\ref{sec:numerical_results}, these theoretical results are still not tight enough to be very useful in practice.
For simplicity, we will begin by focusing on outlier detection.

\noindent \textbf{Review of existing results based on the DKW inequality.}\\
\citet{efficiency_first_cp} have recently studied the finite-sample coverage rate of a conformal prediction interval formed by naively calibrating a model selected among $T$ possible candidates based on its performance on the calibration data set itself, which we denote by $\mathcal{D}_{\text{es-cal}}$.
Although \citet{efficiency_first_cp} focus on conformal prediction intervals, here we find it easier to explain their ideas in the context of conformal p-values for outlier detection.

Let $\hat{S}_i(Z_{n+1};t)$, for all $i \in \mathcal{D}_{\text{es-cal}}$ and $t \in [T]$, denote the nonconformity scores corresponding to model $t$, and denote the  $\lfloor \alpha(1+|\mathcal{D}_{\text{es-cal}}|) \rfloor\text{-th smallest value in }\hat{S}_i(X_{n+1};t)$ as $\hat{Q}_{\alpha}(Z_{n+1};t)$.
Let $t^*$  indicate the selected model.
As we are interested in constructing a conformal p-value $\hat{u}_0^{\textup{naive}}(Z_{n+1})$, the goal is to bound from above the tail probability
\begin{align}
    \mathbb{P}\left( \hat{u}_0^{\textup{naive}}(Z_{n+1}) > \alpha \right) = \mathbb{E} \left[  \mathbb{P} \left( \hat{S}_i(X_{n+1};t^*)> \hat{Q}_{\alpha}(Z_{n+1};t^*) \mid \mathcal{D}_{\text{es-cal}}\right) \right].
\end{align}
Intuitively, if $n_{\text{es-cal}} = |\mathcal{D}_{\text{es-cal}}|$ is sufficiently large, the conditional probability inside the expected value on the right-hand-side above can be well-approximated by the following empirical quantity:
\begin{align*}
    \frac{1}{n}\sum_{i \in \mathcal{D}_{\text{es-cal}}} \mathbbm{1} \left\{ \hat{S}_i(X_{n+1};t^*) > \hat{Q}_{\alpha}(Z_{n+1};t^*) \right \} = \frac{\lceil (1+n_{\text{es-cal}})(1-\alpha) \rceil}{n_{\text{es-cal}}} \geq \left( 1+\frac{1}{n_{\text{es-cal}}}\right)(1-\alpha).
\end{align*}
The quality of this approximation in finite samples can be bound by the DKW inequality, which holds for any $\varepsilon \geq 0$:
\begin{align}
    \mathbb{P}\left( \sup_{s \in \mathbb{R}} \left| \frac{1}{n_{\text{es-cal}}}\sum_{i \in \mathcal{D}_{\text{es-cal}}} \mathbbm{1} \left\{\hat{S}_i(X_{n+1};t^*) > s \right \} -  \mathbb{P} \left( \hat{S}_i(X_{n+1};t^*)> s \mid \mathcal{D}_{\text{es-cal}}\right)\right| > \varepsilon \right) \leq 2 e^{-2 n_{\text{es-cal}} \varepsilon^2}.
\end{align}
Starting from this, Theorem 1 in \citet{efficiency_first_cp} shows that
\begin{align}
     \mathbb{P}(\hat{u}_0^{\textup{naive}}(Z_{n+1}) > \alpha) \geq \left(1+\frac{1}{n_{\text{es-cal}}} \right)(1-\alpha)-\frac{\sqrt{\log(2T)/2}+c(T)}{\sqrt{n_{\text{es-cal}}}},
\end{align}
where $c(T)$ is a constant that can be computed explicitly and is generally smaller than $1/3$.
Intuitively, the $[\sqrt{\log(2T)/2}+c(T)]/ \sqrt{n_{\text{es-cal}}}$ term above can be interpreted as the worst-case approximation error among all possible models $t \in [T]$.

\begin{figure}
    \centering
    \includegraphics[scale=0.63]{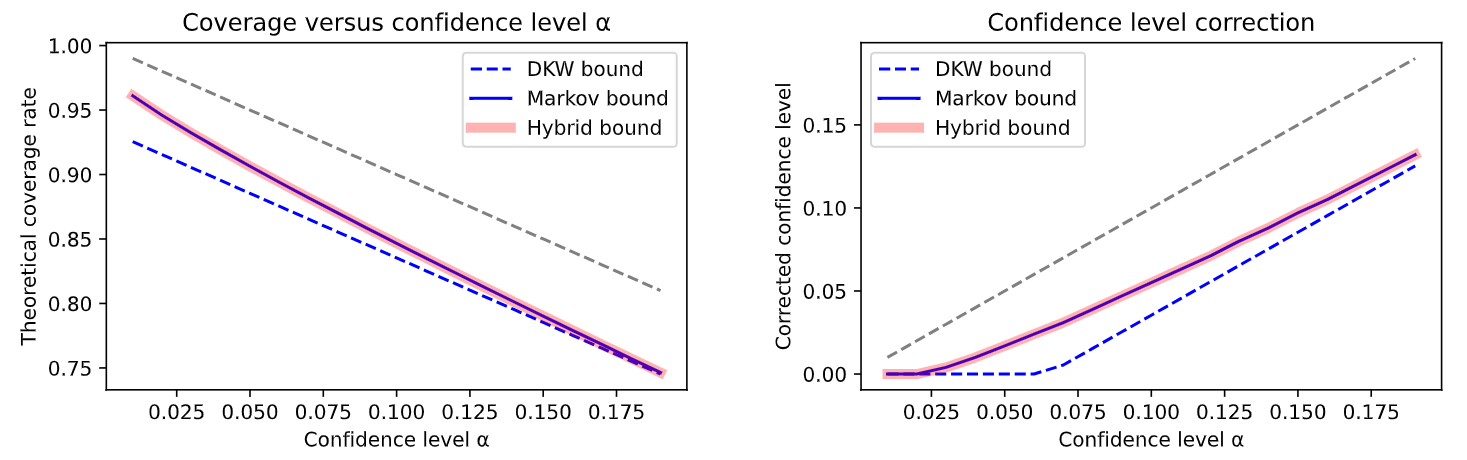}
    \caption{Numerical comparison of different theoretical lower bounds for the marginal coverage of conformal prediction sets computed with a naive early stopping benchmark (e.g., Algorithm~\ref{alg:naive-multi}), as a function of the nominal significance level $\alpha$. Left: lower bounds for the marginal coverage as a function of $\alpha$, when $T = 1000$ and $n_{\text{es-cal}}= 1000$.
Right: theoretically corrected significance level necessary needed to achieve the marginal coverage guarantees expected at the nominal $\alpha$ level, as a function of $\alpha$ when $T = 1000$ and $n_{\text{es-cal}}= 1000$. The dashed grey lines indicate the ideal values corresponding to standard conformal inferences based on calibration data that are independent of those used for early stopping. Higher values correspond to tighter bounds.
}
    \label{fig:bound_alpha}
\end{figure}

One limitation with this result is that is gives a worst-case correction that does not depend on the chosen level $\alpha$, and one would intuitively expect this bound to be tighter for $\alpha = 1/2$ and overly conservative for the small $\alpha$ values (e.g., $\alpha = 0.1$) that are typically interesting in practice. (This intuition will be confirmed empirically in Figure~\ref{fig:bound_alpha}.)
This observation motivates the following alternative analysis, which can often give tighter results.

\noindent \textbf{Alternative probabilistic bound based on Markov's inequality.}\\
Define $W_t=\P{\hat{u}_0^{\textup{naive}}(Z_{n+1};t) > \alpha \mid \mathcal{D}_{\text{es-cal}}}$. Lemma~3 in \citet{vovk2012conditional} tells us that $W_t$ follows a Beta distribution, assuming exchangeability among $\mathcal{D}_{\text{es-cal}}$ and the test point. That is,
\begin{align*}
    W_t \sim \text{Beta}(n_{\text{es-cal}}+1-l, l), \hspace{10pt} l=\lfloor \alpha(n_{\text{es-cal}}+1) \rfloor.
\end{align*}
In the following, we will denote the corresponding inverse Beta cumulative distribution function as $I^{-1}(x;n_{\text{es-cal}}+1-l,l)$.
This result can be used to derive an alternative upper bound for $\mathbb{P}(\hat{u}_0^{\textup{naive}}(Z_{n+1}) > \alpha)$ using the Markov's inequality.

\begin{proposition}\label{prop:naive-od}
    Assume $Z_{1}, \ldots, Z_{n}, Z_{n+1}$ are exchangeable random samples, and let $\hat{u}^{\textup{naive}}_0(Z_{n+1})$ be the output of Algorithm~\ref{alg:naive-one}, for any given $\alpha \in (0,1)$. 
Then, for any fixed $\alpha \in (0,1)$ and any $b>1$, letting $l=\lfloor \alpha(n_{\text{es-cal}}+1)  \rfloor$,
$$\P{\hat{u}^{\textup{naive}}_0(Z_{n+1}) > \alpha} \geq I^{-1}\left( \frac{1}{bT};n_{\text{es-cal}}+1-l,l \right) \cdot (1-1/b).$$
\end{proposition}

Note that this bound depends on $\alpha$ in a more complex way compared to that of \citet{efficiency_first_cp}. 
However, its asymptotic behaviour in the large-$T$ limit remains similar, as shown below.

\begin{lemma}\label{lemma:asymp}
Denote $I^{-1}(x;n_{\text{es-cal}}+1-l,l)$ as the inverse Beta cumulative distribution function. For any fixed $b > 1$ and $\alpha \in (0,1)$ , letting $l=\lfloor \alpha(n_{\text{es-cal}}+1) \rfloor$, for sufficiently large $T$ and $n_{\text{es-cal}}$, we have:
\begin{align*}
    I^{-1}\left( \frac{1}{bT};n_{\text{es-cal}}+1-l,l \right) = (1-\alpha) - \sqrt{\frac{\alpha(1-\alpha)}{n_{\text{es-cal}}+1}}\cdot \sqrt{2\log(bT)} + O\left( \frac{1}{\sqrt{n_{\text{es-cal}}\log(T)}} \right).
\end{align*}

\end{lemma}

In simpler terms, Lemma~\ref{lemma:asymp} implies that the coverage lower bound in Proposition \ref{prop:naive-od} is approximately equal to
$$ \left[ (1-\alpha) - \sqrt{\frac{\alpha(1-\alpha)}{n_{\text{es-cal}}+1}}\cdot \sqrt{2\log(bT)} \right] \cdot \left(1-\frac{1}{b}\right), $$ 
which displays an asymptotic behaviour similar to that of the bound from \citet{efficiency_first_cp}.
Further, the  Markov bound is easy to compute numerically and often turns out to be tighter as long as $b$ is moderately large (e.g., $b=100$), as we shall see below.
Naturally, the same idea can also be applied to bound the coverage of naive conformal prediction sets or intervals output by Algorithm~\ref{alg:naive-multi} or Algorithm~\ref{alg:naive-reg}, respectively.

\begin{corollary} \label{prop:naive-class}
    Assume $(X_1,Y_1), \ldots, (X_n,Y_n), (X_{n+1},Y_{n+1})$ are exchangeable random sample, and let $\hat{C}_{\alpha}^{\textup{naive}}(X_{n+1})$ be the output of Algorithm~\ref{alg:naive-multi}, for any given $\alpha \in (0,1)$. Then, for any $b > 1$, letting $l=\lfloor \alpha(n_{\text{es-cal}}+1) \rfloor$, $$\P{Y_{n+1} \in \hat{C}^{\textup{naive}}_{\alpha}(X_{n+1})} \geq I^{-1}\left(\frac{1}{bT}; n_{\text{es-cal}}+1-l,l\right) \cdot (1-1/b).$$
\end{corollary}

\begin{corollary}\label{prop:naive-reg}
Assume $(X_{1},Y_{1}), \ldots, (X_{n},Y_{n}), (X_{n+1},Y_{n+1})$ are exchangeable random samples, and let $\hat{C}^{\textup{naive}}_{\alpha}(X_{n+1})$ be the output of Algorithm~\ref{alg:naive-reg}, for any $\alpha \in (0,1)$. Then, for any $b > 1$, letting $l=\lfloor \alpha(n_{\text{es-cal}}+1) \rfloor$,
$$\P{Y_{n+1} \in \hat{C}^{\textup{naive}}_{\alpha}(X_{n+1})} \geq I^{-1}\left(\frac{1}{bT}; n_{\text{es-cal}}+1-l,l\right) \cdot (1-1/b).$$
\end{corollary}

\noindent \textbf{Hybrid probabilistic bound.}
Since neither the DKW nor the Markov bound described above always dominate the other for all possible combinations of $T$, $n_{\text{es-cal}}$, and $\alpha$, it makes sense to combine them to obtain a uniformly tighter {\em hybrid} bound.
For any fixed $b>1$ and any $T$, $n_{\text{es-cal}}$, and $\alpha$, let $M(T,n_{\text{es-cal}},\alpha) = I^{-1} \left(1/{bT}; n_{\text{es-cal}}+1-l,l \right) \cdot (1-1/b)$ denote the Markov bound and $D(T,n_{\text{es-cal}},\alpha) =\left(1+{1}/(n_{\text{es-cal}})\right)(1-\alpha)-(\sqrt{\log(2T)/2}+c(T))/{\sqrt{n_{\text{es-cal}}}}$ denote the DKW bound, define $H(T,n_{\text{es-cal}},\alpha)$ as
\begin{align*}
    H(T,n_{\text{es-cal}},\alpha)
  = \max\left\{ M(T,n_{\text{es-cal}},\alpha), D(T,n_{\text{es-cal}},\alpha) \right\}.
\end{align*}
It then follows immediately from \citet{efficiency_first_cp} and Proposition~\ref{prop:naive-od} that, under the same conditions of Proposition~\ref{prop:naive-od}, for any  fixed $b>1$,
$$\P{\hat{u}^{\textup{naive}}_0(Z_{n+1}) > \alpha} \geq H(T,n_{\text{es-cal}},\alpha).$$
Of course, the same argument can also be utilized to tighten the results of Corollaries~\ref{prop:naive-class}--\ref{prop:naive-reg}.

\begin{figure}
    \centering
    \includegraphics[scale=0.63]{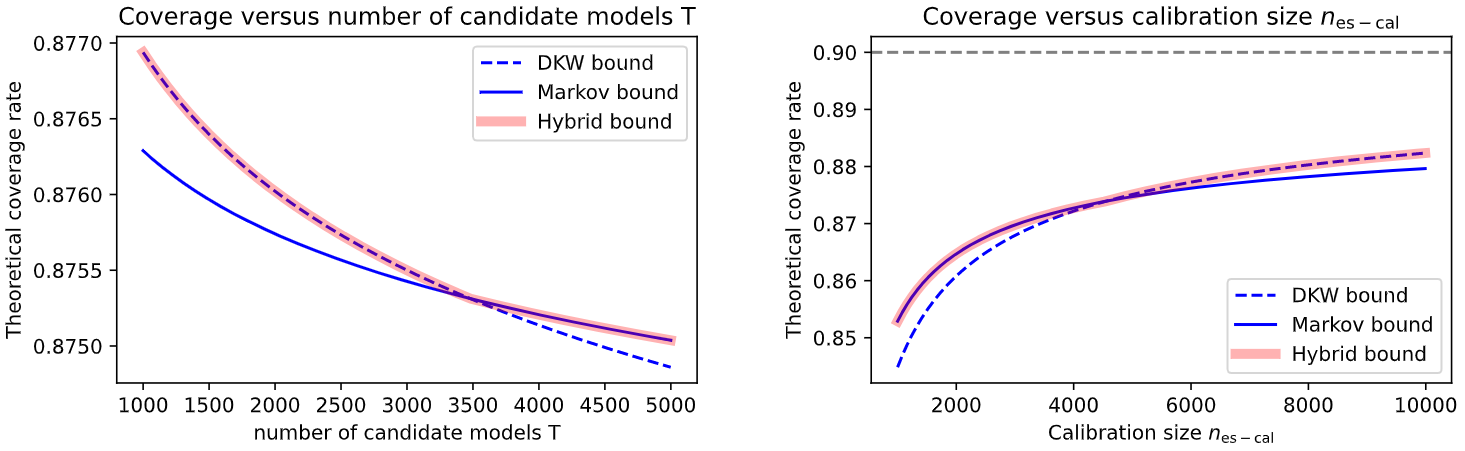}
    \caption{Numerical comparison of different theoretical lower bounds for the marginal coverage of conformal prediction sets computed with a naive early stopping benchmark (e.g., Algorithm~\ref{alg:naive-multi}). Left: lower bounds for the marginal coverage as a function of the number of candidate models $T$, when $\alpha=0.1$ and $n_{\text{es-cal}}=8000$.
Right: lower bounds for the marginal coverage as a function of the number of hold-out data points, $n_{\text{es-cal}}$, when $\alpha=0.1$ and $T=100$. Higher values correspond to tighter bounds.
}
    \label{fig:bound_tn}
\end{figure}

\noindent \textbf{Numerical comparison of different probabilistic bounds.}
Figure~\ref{fig:bound_tn} compares the three probabilistic bounds described above ({\em DKW}, {\em Markov}, and {\em hybrid}) as a function of the number of candidate models $T$ and of the number of hold-out data points $n_{\text{es-cal}}$, in the case of $\alpha=0.1$. For simplicity, the Markov and hybrid bounds are evaluated by setting $b=100$, which may not be the optimal choice but appears to work reasonably well. These results show that Markov bound tends to be tighter than the DKW bound for large values of $T$ and for small values of $n_{\text{es-cal}}$, while the hybrid bound generally achieves the best of both worlds.
Lastly, Figure~\ref{fig:bound_alpha} demonstrates that the Markov bound tends to be tighter when $\alpha$ is small. The Markov and hybrid bounds here are also evaluated using $b=100$.

\section{Classification with Marginal Coverage} \label{app:class-marg}

The conformalized early stopping method presented in Section~\ref{sec:classification} can be easily modified to produce prediction sets with marginal rather than label-conditional coverage, as outlined in Algorithm~\ref{alg:class_full_seq_marg}.
The difference between Algorithm~\ref{alg:class_full_seq}  and Algorithm~\ref{alg:class_full_seq_marg} is that the latter utilizes all calibration data in $\mathcal{D}_{\text{es-cal}}$ to compute each conformal p-value $\hat{u}_y(X_{n+1})$, not only the samples with true label $y$.
An advantage of this approach is that conformal p-values based on a larger calibration samples are less aleatoric~\cite{bates2021testing} and require less conservative finite-sample corrections (i.e., the ``+1'' term the numerator of the p-value formula becomes more negligible as the calibration set size increases).
In turn, this tends to lead to smaller prediction sets with potentially more stable coverage conditional on the calibration data~\cite{sesia2020comparison,bates2021testing}
Of course, the downside of these prediction sets is that they can only be guaranteed to provide marginal coverage, although they can sometimes also perform well empirically in terms of label-conditional coverage~\cite{romano2020classification}.

\begin{theorem} \label{thm:class_full_marg}
Assume $(X_{1},Y_{1}), \ldots, (X_{n},Y_{n}), (X_{n+1},Y_{n+1})$ are exchangeable random samples, and let $\hat{C}_{\alpha}(X_{n+1})$ be the output of Algorithm~\ref{alg:class_full_seq_marg}, for any $\alpha \in (0,1)$. 
Then, $\mathbb{P}[Y_{n+1} \in \hat{C}_{\alpha}(X_{n+1})] \geq 1-\alpha$.
\end{theorem}

\begin{algorithm}[H]
    \caption{Conformalized early stopping for multi-class classification with marginal coverage}
    \label{alg:class_full_seq_marg}
    \begin{algorithmic}[1]
        \STATE \textbf{Input}: Exchangeable data points $(X_{1},Y_{1}), \ldots, (X_{n},Y_{n})$ with labels $Y_i \in [K]$.
        \STATE \textcolor{white}{\textbf{Input}:} Test point with features $X_{n+1}$. Desired coverage level $1-\alpha$.
        \STATE \textcolor{white}{\textbf{Input}:} Maximum number of training epochs $t^{\text{max}}$; storage period hyper-parameter $\tau$.
        \STATE \textcolor{white}{\textbf{Input}:} $K$-class classifier trainable via (stochastic) gradient descent.
        \STATE Randomly split the exchangeable data points into $\mathcal{D}_{\text{train}}$ and $\mathcal{D}_{\text{es-cal}}$.
        \STATE Train the $K$-class classifier for $t^{\text{max}}$ epochs and save the intermediate models $M_{t_1} , \dots, M_{t_T}$.
        \FOR{$ y \in [K]$}
        \STATE Imagine $Y_{n+1}=y$.
        \STATE Pick the model $\hat{M}_{\text{ces}}(X_{n+1},y)$ according to~\eqref{eq:ces-model-class}, using the data in $\mathcal{D}_{\text{es-cal}} \cup \{n+1\}$.
        \STATE Compute scores $\hat{S}_i(X_{n+1},y)$ for all $i \in \mathcal{D}_{\text{es-cal}} \cup \{n+1\}$ using $\hat{M}_{\text{ces}}(X_{n+1},y)$; see Appendix~\ref{app:class-scores}.
        \STATE Compute the conformal p-value $\hat{u}^{\text{marg}}_y(X_{n+1})$ according to
        \begin{align}\label{eq:conformal_pval-class_marg}
          \hat{u}^{\text{marg}}_y(X_{n+1}) = \frac{1 + |i \in \mathcal{D}_{\text{es-cal}}: \hat{S}_{i}^y(X_{n+1}) \leq \hat{S}_{n+1}^y(X_{n+1})|}{1+|\mathcal{D}_{\text{es-cal}}|}.
        \end{align}
        \ENDFOR
        \STATE \textbf{Output}: Prediction set $\hat{C}_{\alpha}(X_{n+1})$ given by \eqref{eq:pred-set-class}, with $\hat{u}^{\text{marg}}_y(X_{n+1})$ instead of $\hat{u}_y(X_{n+1})$.
    \end{algorithmic}
\end{algorithm}

\section{Review of Nonconformity Scores for Classification} \label{app:class-scores}

This section reviews the relevant background on the adaptive nonconformity scores for classification developed by \citet{romano2020classification}.
For any $x \in \mathcal{X}$ and $y \in [K]$, let $\hat{\pi}_{y}(x)$ denote any (possibly very inaccurate) estimate of the true $\mathbb{P}[ Y = y \mid X =x]$ corresponding to the unknown data-generating distribution. Concretely, a typical choice of $\hat{\pi}$ may be given by the output of the final softmax layer of a neural network classifier, for example.
 For any $x \in \mathcal{X}$ and $\tau \in [0,1]$, define the \emph{generalized conditional quantile} function $L$, with input $x, \hat{\pi}, \tau$, as:
\begin{align} \label{eq:oracle-threshold}
  L(x; \hat{\pi}, \tau) & = \min \{ k \in [K] \ : \ \hat{\pi}_{(1)}(x) + \hat{\pi}_{(2)}(x) + \ldots + \hat{\pi}_{(k)}(x) \geq \tau \},
  \end{align}
where $\hat{\pi}_{(1)}(x) \leq \hat{\pi}_{(2)}(x) \leq \ldots \hat{\pi}_{(K)}(x)$ are the order statistics of $\hat{\pi}_{1}(x) \leq \hat{\pi}_{2}(x) \leq \ldots \hat{\pi}_{K}(x)$.
Intuitively, $L(x; \hat{\pi}, \tau)$ gives the size of the smallest possible subset of labels whose cumulative probability mass according to $\hat{\pi}$ is at least $\tau$.
Define also a function $\mathcal{S}$ with input $x$, $u \in (0,1)$, $\hat{\pi}$, and $\tau$ that computes the set of most likely labels up to (but possibly excluding) the one identified by $L(x; \hat{\pi}, \tau)$:
\begin{align} \label{eq:define-S}
    \mathcal{S}(x, u ; \hat{\pi}, \tau) & =
    \begin{cases}
    \text{ `$y$' indices of the $L(x ; \hat{\pi},\tau)-1$ largest $\hat{\pi}_{y}(x)$},
    & \text{ if } u \leq V(x ; \hat{\pi},\tau) , \\
    \text{ `$y$' indices of the $L(x ; \hat{\pi},\tau)$ largest $\hat{\pi}_{y}(x)$},
    & \text{ otherwise},
    \end{cases}
\end{align}
where
\begin{align*}
    V(x; \hat{\pi}, \tau) & =  \frac{1}{\hat{\pi}_{(L(x ; \hat{\pi}, \tau))}(x)} \left[\sum_{k=1}^{L(x ; \hat{\pi}, \tau)} \hat{\pi}_{(k)}(x) - \tau \right].
\end{align*}
Then, define the \textit{generalized inverse quantile} nonconformity score function $s$, with input $x,y,u;\hat{\pi}$, as:
\begin{align} \label{eq:define-scores}
    s(x,y,u;\hat{\pi}) & = \min \left\{ \tau \in [0,1] : y \in \mathcal{S}(x, u ; \hat{\pi}, \tau) \right\}.
\end{align}
Intuitively, $s(x,y,u;\hat{\pi})$ is the smallest value of $\tau$ for which the set $\mathcal{S}(x, u ; \hat{\pi}, \tau)$ contains the label $y$.
Finally, the nonconformity score for a data point $(X_i,Y_i)$ is given by:
\begin{align}
  \hat{S}_i
  & = s(X_i,Y_i,U_i;\hat{\pi}),
\end{align}
where $U_i$ is a uniform random variable independent of anything else. Note that this can also be equivalently written more explicitly as:
\begin{align}
  \hat{S}_i
  & = \hat{\pi}_{(1)}(X_i) + \hat{\pi}_{(2)}(X_i) + \ldots + \hat{\pi}_{(r(Y_i,\hat{\pi}(X_i)))}(X_i) - U_i\cdot \hat{\pi}_{(r(Y_i,\hat{\pi}(X_i)))}(X_i),
\end{align}
where $r(Y_i,\hat{\pi}(X_i))$ is the rank of $Y_i$ among the possible labels $y \in [K]$ based on $\hat{\pi}_y(X_i)$, so that $r(y,\hat{\pi}(X_i))=1$ if $\hat{\pi}_{y}(X_i) = \hat{\pi}_{(1)}(X_i)$.
The idea motivating this construction is that the nonconformity score $\hat{S}_i$ defined above is guaranteed to be uniformly distributed on $[0,1]$ conditional on $X$ if the model $\hat{\pi}$ estimates the true unknown $\mathbb{P}[ Y = y \mid X =x]$ accurately for all $x \in \mathcal{X}$.
This is a desirable property in conformal inference because it leads to statistically efficient prediction sets that can often achieve relatively high feature-conditional coverage in practice, even if the true data-generating distribution is such that some observations are much noisier than others; see \citet{romano2020classification} for further details.

Finally, we conclude this appendix by noting that the nonconformity scores in  Section~\ref{sec:classification} are written as $\hat{S}_i(X_{n+1},y)$, instead of the more compact notation $\hat{S}_i$ adopted here, simply to emphasize that they are computed based on class probabilities $\hat{\pi}$ estimated by a data-driven model $\hat{M}$ that depends on the test features $X_{n+1}$ as well as on the placeholder label $y$ for $Y_{n+1}$.

\section{Efficient Computation of the Lower Envelope} \label{app:lower-envelope}

This section explains how to implement a computationally efficient divide-and-conquer algorithm for finding the lower envelope of a family of $T$ parabolas or a family of shifted pinball loss functions at cost $\mathcal{O}(T \log T)$ \cite{devillers1995incremental,nielsen1998output}.
This solution, outlined in Algorithm~\ref{alg:envelope} and Algorithm~\ref{alg:envelope-cqr}, is useful to implement the proposed CES method for regression problems, as detailed in Algorithm~\ref{alg:reg} and Algorithm~\ref{alg:reg-cqr}.

\begin{algorithm}[H]
    \caption{Divide-and-conquer algorithm for finding the lower envelope of many parabolas}
    \label{alg:envelope}
    \begin{algorithmic} [1]
        \STATE \textbf{Input}: A set of parabolas $L = \{l_1, l_2, \dots, l_T \}$ of forms $l_i = a_i x^2 + b_i x + c_i$ for $i=1,\dots, T$.
        \STATE Randomly split $L$ into two subsets. Repeat splitting until each subset only contains one parabola or is empty.
        \STATE For each subset with only one parabola, set the parabola itself as the lower envelope and set the initial breakpoint list to $[-\infty, +\infty]$.
        \FOR{each interval constructed by adjacent breakpoints}
            \STATE Within the interval, identify the two parabolas contributing to the previous lower envelopes, denoted as $P_1$, $P_2$.
            \STATE Evaluate $P_1$ and $P_2$ at the current interval endpoints.
            \STATE Calculate the intersection point $p$ of $P_1$ and $P_2$. There exists at most one such $p$ because $a_i = 1, \forall i$, by~\eqref{eq:loss-ces-reg}.
            \IF{$p$ not exists or $p$ exists but lies outside the current interval}
            \STATE Set the new lower envelope as the parabola with smaller values computed at the interval endpoints.
            \ELSE \STATE Add $p$ as a breakpoint.
            \STATE Within the current interval, set the new lower envelope below and above $p$ based on evaluations of the parabolas at the interval endpoints.
            \ENDIF
            \STATE Update and sort the breakpoint list and update the new lower envelope.
        \ENDFOR
        \STATE Recursively merge two lower envelopes to form a new lower envelope by repeating Lines 4--15.
        \STATE \textbf{Output}: A sorted dictionary of breakpoints and parabola indices characterizing the lower envelope of $L$.
\end{algorithmic}
\end{algorithm}

\begin{algorithm}[H]
    \caption{Divide-and-conquer algorithm for finding the lower envelope of many pinball loss functions}
    \label{alg:envelope-cqr}
    \begin{algorithmic} [1]
        \STATE \textbf{Input}: A set of shifted pinball loss functions $L = \{l_1, l_2, \dots, l_T \}$ of forms $l_i = c_i + \rho_\beta(y, \hat{y})$ for $i=1,\dots, T$.
        \STATE Randomly split $L$ into two subsets. Repeat splitting until each subset only contains one pinball loss function or is empty.
        \STATE For each subset with only one pinball loss function, set the function itself as the lower envelope and set the initial breakpoint list to $[-\infty, +\infty]$.
        \FOR{each interval constructed by adjacent breakpoints}
            \STATE Within the interval, identify the two pinball loss functions contributing to the previous lower envelopes; i.e., $P_1$, $P_2$.
            \STATE Evaluate $P_1$ and $P_2$ at the current interval endpoints.
            \STATE Calculate the intersection point $p$ of $P_1$ and $P_2$. There exists at most one such $p$ because $\beta$ is the same $\forall i$, by~\eqref{eq:loss-ces-reg-cqr}.
            \IF{$p$ not exists or $p$ exists but lies outside the current interval}
            \STATE Set the new lower envelope as the pinball loss function with smaller values computed at the interval endpoints.
            \ELSE \STATE Add $p$ as a breakpoint.
            \STATE Within the current interval, set the new lower envelope below and above $p$ based on evaluations of the pinball loss functions at the interval endpoints.
            \ENDIF
            \STATE Update and sort the breakpoint list and update the new lower envelope.
        \ENDFOR
        \STATE Recursively merge two lower envelopes to form a new lower envelope by repeating Lines 4--15.
        \STATE \textbf{Output}: A sorted dictionary of breakpoints and pinball loss function indices characterizing the lower envelope of $L$.
\end{algorithmic}
\end{algorithm}

\section{Avoiding Empty Predictions in CES for Regression} \label{app:reg-noempty}

This section presents Algorithm~\ref{alg:reg-noempty}, which extends Algorithm~\ref{alg:reg} from Section~\ref{sec:regression} in such a way as to explicitly avoid returning empty prediction intervals.

\begin{algorithm}[H]
    \caption{Conformalized early stopping for regression, avoiding empty predictions}
    \label{alg:reg-noempty}
    \begin{algorithmic}[1]
        \STATE \textbf{Input}: Exchangeable data points $(X_{1},Y_{1}), \ldots, (X_{n},Y_{n})$ with outcomes $Y_i \in \mathbb{R}$.
        \STATE \textcolor{white}{\textbf{Input}:} Test point with features $X_{n+1}$. Desired coverage level $1-\alpha$.
        \STATE \textcolor{white}{\textbf{Input}:} Maximum number of training epochs $t^{\text{max}}$; storage period hyper-parameter $\tau$.
        \STATE \textcolor{white}{\textbf{Input}:} Regression model trainable via (stochastic) gradient descent.
        \STATE Randomly split the exchangeable data points into $\mathcal{D}_{\text{train}}$ and $\mathcal{D}_{\text{es-cal}}$.
        \STATE Train the regression model for $t^{\text{max}}$ epochs and save the intermediate models $M_{t_1} , \dots, M_{t_T}$.
        \STATE Evaluate $\hat{C}_{\alpha}(X_{n+1})$ using Algorithm~\ref{alg:reg}.
        \IF{$\hat{C}_{\alpha}(X_{n+1}) = \emptyset$}
        \STATE Evaluate $\hat{C}^{\text{naive}}_{\alpha}(X_{n+1})$ using Algorithm~\ref{alg:naive-reg}. Set $\hat{C}_{\alpha}(X_{n+1}) = \hat{C}^{\text{naive}}_{\alpha}(X_{n+1})$.
        \ENDIF
        \STATE \textbf{Output}: A non-empty prediction interval $\hat{C}_{\alpha}(X_{n+1})$.
    \end{algorithmic}
\end{algorithm}

\begin{corollary}\label{thm:reg-noempty}
Assume $(X_{1},Y_{1}), \ldots, (X_{n},Y_{n}), (X_{n+1},Y_{n+1})$ are exchangeable random samples, and let $\hat{C}_{\alpha}(X_{n+1})$ be the output of Algorithm~\ref{alg:reg-noempty}, for any $\alpha \in (0,1)$. 
Then, $\mathbb{P}[Y_{n+1} \in \hat{C}_{\alpha}(X_{n+1})] \geq 1-\alpha$.
\end{corollary}



\section{Review of Conformalized Quantile Regression}
\label{app:cqr_review}

\begin{figure}[!htb]
    \centering
    \includegraphics[width=0.4\linewidth]{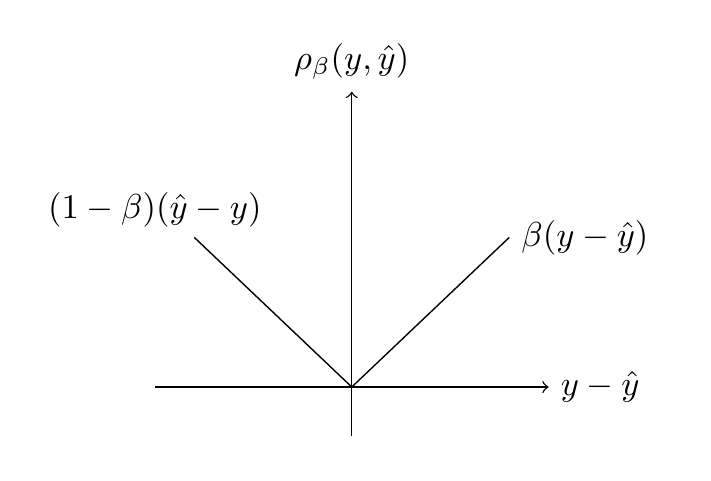}
    \caption{Visualization of the pinball loss function defined in~\eqref{eq:reg-pinball_loss}.}
    \label{fig:pinball_loss}%
\end{figure}

This section reviews the relevant background on conditional quantile regression \cite{koenker1978quantreg} and conformalized quantile regression (CQR) \cite{romano2019conformalized}. 
In contrast to the classical regression models that estimate the conditional mean of the test response $Y_{n+1}$ given the test feature $X_{n+1} = x$, quantile regression estimates the conditional quantile $q_\beta$ of $Y_{n+1}$ given $X_{n+1} = x$, which is defined as
\begin{align} \label{eq:reg-cond_quantile}
    q_\beta(x) = \inf\{ y\in \mathbb{R}: \mathbb{P}(Y_{n+1} \leq y |X_{n+1} = x) \geq \beta \}.
\end{align}
This can be formulated as solving the optimization problem: 
\begin{align} \label{eq:reg-quantile_reg}
    \hat{q}_\beta(x) = f(x,\hat{\theta}), \quad \hat{\theta} = \argmin_{\theta}\frac{1}{n}\sum_{i=1}^n \rho_\beta(Y_i, f(X_i, \theta)),
\end{align}
where $f(x,\theta)$ represents the quantile regression function \cite{koenker1978quantreg} and $\rho_\beta$ is the convex ``pinball loss" function \cite{Steinwart2011pinball}, illustrated in Figure~\ref{fig:pinball_loss} and mathematically defined as
\begin{align} \label{eq:reg-pinball_loss}
    \rho_\beta(y, \hat{y}) = 
  \begin{cases}\beta(y - \hat{y}), & \text{if } y - \hat{y} >0, \\
    (1-\beta)(\hat{y} - y), & \text{otherwise}.
    \end{cases}
\end{align}

To construct an efficient prediction interval $\hat{C}(X_{n+1})$ whose length is adaptive to the local variability of $X_{n+1}$, CQR operates as follows.
As in split conformal prediction, firstly the available data are randomly split into a proper training set, indexed by $\mathcal{I}_1$, and a calibration set, indexed by $\mathcal{I}_2$. Given any quantile regression algorithm $\mathcal{A}$, two conditional quantile functions, $\hat{q}_{\alpha_{\text{lo}}}$ and $\hat{q}_{\alpha_{\text{hi}}}$, are fitted on $\mathcal{I}_1$, where $\alpha_{\text{lo}} = \alpha/2$ and $\alpha_{\text{hi}} = 1-\alpha/2$: 
\begin{align} \label{eq:reg-cond_quantile_fit}
    \{ \hat{q}_{\alpha_{\text{lo}}}, \hat{q}_{\alpha_{\text{hi}}} \} \leftarrow \mathcal{A}(\{ (X_i, Y_i): i \in \mathcal{I}_1 \}).
\end{align}
Then, conformity scores are computed on the calibration data set $\mathcal{I}_2$ as:
\begin{align} \label{eq:reg-cond_quantile_calib}
    E_i = \max \{ \hat{q}_{\alpha_{\text{lo}}}(X_i) - Y_i, Y_i - \hat{q}_{\alpha_{\text{hi}}}(X_i) \} \quad \text{ for } i \in \mathcal{I}_2.
\end{align}
The conformity score in~\eqref{eq:reg-cond_quantile_calib} can account both for possible under-coverage and over-coverage of the quantile regression model~\cite{romano2019conformalized}. If $Y_i$ is outside the interval $[\hat{q}_{\alpha_{\text{lo}}}(X_i), \hat{q}_{\alpha_{\text{hi}}}(X_i)]$, then $E_i$ is the (positive) distance of $Y_i$ from the closest endpoint of the interval.
Otherwise, if $Y_i$ is inside the interval $[\hat{q}_{\alpha_{\text{lo}}}(X_i), \hat{q}_{\alpha_{\text{hi}}}(X_i)]$, then $E_i$ is the negative of the distance of $Y_i$ from the closest endpoint of the interval.
Therefore, if the quantile regression model is well-calibrated, approximately 90\% of the calibration data points should have $E_i \leq 0$~\cite{romano2019conformalized}.
Finally, CQR constructs the prediction interval for the test response value $Y_{n+1}$ through
\begin{align} \label{eq:reg-cond_quantile_construct_pi}
    \hat{C}(X_{n+1}) = [ \hat{q}_{\alpha_{\text{lo}}}(X_{n+1}) - \hat{Q}_{1-\alpha}(E, \mathcal{I}_2), \hat{q}_{\alpha_{\text{hi}}}(X_{n+1}) + \hat{Q}_{1-\alpha}(E, \mathcal{I}_2)    ],
\end{align}
where $\hat{Q}_{1-\alpha}(E, \mathcal{I}_2)$ is the $(1-\alpha)(1+1/|\mathcal{I}_2|)$-th empirical quantile of $\{E_i: i\in \mathcal{I}_2\}$.

\section{Avoiding Empty Predictions for Regression with CQR} \label{app:reg-noempty-quantile}

In this section, we introduce Algorithm~\ref{alg:reg-noempty-cqr} as an extension of Algorithm~\ref{alg:reg-cqr} in the main text to address the rare possibility of generating empty prediction sets. Algorithm~\ref{alg:reg-noempty-cqr} ensures that the intervals it produces are always non-empty, while still encompassing the intervals obtained from Algorithm~\ref{alg:reg-cqr}. Consequently, Algorithm~\ref{alg:reg-noempty-cqr} maintains the guaranteed coverage provided by Theorem~\ref{thm:reg-cqr}, as indicated in Corollary~\ref{thm:reg-noempty-cqr}.

\begin{algorithm}[H]
    \caption{Conformalized early stopping for quantile regression, avoiding empty predictions}
    \label{alg:reg-noempty-cqr}
    \begin{algorithmic}[1]
        \STATE \textbf{Input}: Exchangeable data points $(X_{1},Y_{1}), \ldots, (X_{n},Y_{n})$ with outcomes $Y_i \in \mathbb{R}$.
        \STATE \textcolor{white}{\textbf{Input}:} Test point with features $X_{n+1}$. Desired coverage level $1-\alpha$.
        \STATE \textcolor{white}{\textbf{Input}:} Maximum number of training epochs $t^{\text{max}}$; storage period hyper-parameter $\tau$.
        \STATE \textcolor{white}{\textbf{Input}:} Trainable quantile regression model with target quantiles [$\beta_{\text{low}}, \beta_{\text{high}}$].
        \STATE Randomly split the exchangeable data points into $\mathcal{D}_{\text{train}}$ and $\mathcal{D}_{\text{es-cal}}$.
        \STATE Train for $t^{\text{max}}$ epochs and save the intermediate models $M_{\beta_{\text{low}}, t_1} , \dots, M_{\beta_{\text{low}}, t_T}$, $M_{\beta_{\text{high}}, t_1} , \dots, M_{\beta_{\text{high}}, t_T}$.
        \STATE Evaluate $\hat{C}_{\alpha}(X_{n+1})$ using Algorithm~\ref{alg:reg-cqr}.
        \IF{$\hat{C}_{\alpha}(X_{n+1}) = \emptyset$}
        \STATE Evaluate $\hat{C}^{\text{naive}}_{\alpha}(X_{n+1})$ using Algorithm~\ref{alg:naive-reg-cqr}. Set $\hat{C}_{\alpha}(X_{n+1}) = \hat{C}^{\text{naive}}_{\alpha}(X_{n+1})$.
        \ENDIF
        \STATE \textbf{Output}: A non-empty prediction interval $\hat{C}_{\alpha}(X_{n+1})$.
    \end{algorithmic}
\end{algorithm}

\begin{corollary}\label{thm:reg-noempty-cqr}
Assume $(X_{1},Y_{1}), \ldots, (X_{n},Y_{n}), (X_{n+1},Y_{n+1})$ are exchangeable random samples, and let $\hat{C}_{\alpha}(X_{n+1})$ be the output of Algorithm~\ref{alg:reg-noempty-cqr}, for any $\alpha \in (0,1)$. 
Then, $\mathbb{P}[Y_{n+1} \in \hat{C}_{\alpha}(X_{n+1})] \geq 1-\alpha$.
\end{corollary}

\subsection{Implementation of the Naive Benchmark}  \label{app:naive-benchmarks-details-cqr}

\begin{algorithm}[H]
    \caption{Naive conformal quantile regression benchmark with greedy early stopping}
    \label{alg:naive-reg-cqr}
    \begin{algorithmic}[1]
        \STATE \textbf{Input}: Exchangeable data points $(X_{1},Y_{1}), \ldots, (X_{n},Y_{n})$ with outcomes $Y_i \in \mathbb{R}$.
        \STATE \textcolor{white}{\textbf{Input}:} Test point with features $X_{n+1}$. Desired coverage level $1-\alpha$.
        \STATE \textcolor{white}{\textbf{Input}:} Maximum number of training epochs $t^{\text{max}}$; storage period hyper-parameter $\tau$.
        \STATE \textcolor{white}{\textbf{Input}:} Trainable quantile regression model with target quantiles [$\beta_{\text{low}}, \beta_{\text{high}}$]. 
        \STATE Randomly split the exchangeable data points into $\mathcal{D}_{\text{train}}$ and $\mathcal{D}_{\text{es-cal}}$.
        \STATE Train for $t^{\text{max}}$ epochs and save the intermediate models $M_{\beta_{\text{low}},t_1} , \dots, M_{\beta_{\text{low}}t_T}$,  $M_{\beta_{\text{high}},t_1} , \dots, M_{\beta_{\text{high}}t_T}$.
        \STATE Pick the most promising models $t^*_{\text{low}}, t^*_{\text{high}}\in [T]$ minimizing $\mathcal{L}_{\text{es-cal}}(M_t)$ in \eqref{eq:loss-ces-reg-cqr}.
        \STATE Evaluate the scores $\hat{E}_i(X_{n+1}) = \max\{\hat{q}_{t^*_{\text{low}}}(X_i) - Y_i, Y_i -\hat{q}_{t^*_{\text{high}}}(X_i)\}$ for all $i \in \mathcal{D}_{\text{es-cal}}$.
        \STATE Compute $\hat{Q}_{1-\alpha}(X_{n+1}) = \lceil (1-\alpha)(1+|\mathcal{D}_{\text{es-cal}}|) \rceil\text{-th smallest value in }
        \hat{E}_i(X_{n+1})$ for $i \in \mathcal{D}_{\text{es-cal}}$.
        \STATE \textbf{Output}: $\hat{C}^{\text{naive}}_{\alpha}(X_{n+1}) = [\hat{q}_{t^*_{\text{low}}}(X_{n+1}) - \hat{Q}_{1-\alpha}(X_{n+1}), \hat{q}_{t^*_{\text{high}}}(X_{n+1}) + \hat{Q}_{1-\alpha}(X_{n+1})]$.
    \end{algorithmic}
\end{algorithm}

\clearpage

\section{Additional Results from Numerical Experiments} \label{app:numerical-results}

\subsection{Outlier Detection}

\begin{table}[!htb]
\centering
    \caption{Performance of outlier detection based on classification models trained with different methods, on the {\em CIFAR10} data set~\cite{cifar10}. Other details are as in Figure~\ref{fig:exp_oc}. The numbers in parenthesis indicate standard errors. The numbers in bold highlight TPR values within 1 standard error of the best TPR across all methods, for each sample size.}
  \label{tab:exp_oc}
  
\begin{tabular}[t]{rlll}
\toprule
Sample size & Method & TPR & FPR\\
\midrule
\addlinespace[0.3em]
\multicolumn{4}{l}{\textbf{500}}\\
\hspace{1em}500 & CES & \textbf{0.296 (0.008)} & 0.098 (0.003)\\
\hspace{1em}500 & Naive & \textbf{0.295 (0.008)} & 0.097 (0.003)\\
\hspace{1em}500 & Naive + theory & 0.114 (0.006) & 0.016 (0.001)\\
\hspace{1em}500 & Data splitting & 0.234 (0.008) & 0.091 (0.003)\\
\hspace{1em}500 & Full training & 0.217 (0.011) & 0.072 (0.004)\\
\addlinespace[0.3em]
\multicolumn{4}{l}{\textbf{1000}}\\
\hspace{1em}1000 & CES & \textbf{0.401 (0.007)} & 0.100 (0.004)\\
\hspace{1em}1000 & Naive & \textbf{0.401 (0.007)} & 0.100 (0.004)\\
\hspace{1em}1000 & Naive + theory & 0.237 (0.006) & 0.030 (0.002)\\
\hspace{1em}1000 & Data splitting & 0.337 (0.009) & 0.094 (0.003)\\
\hspace{1em}1000 & Full training & 0.189 (0.013) & 0.055 (0.004)\\
\addlinespace[0.3em]
\multicolumn{4}{l}{\textbf{2000}}\\
\hspace{1em}2000 & CES & \textbf{0.450 (0.005)} & 0.100 (0.003)\\
\hspace{1em}2000 & Naive & \textbf{0.450 (0.005)} & 0.100 (0.003)\\
\hspace{1em}2000 & Naive + theory & 0.337 (0.006) & 0.048 (0.002)\\
\hspace{1em}2000 & Data splitting & 0.404 (0.007) & 0.095 (0.003)\\
\hspace{1em}2000 & Full training & 0.050 (0.009) & 0.017 (0.003)\\
\bottomrule
\end{tabular}

\end{table}

\FloatBarrier

\subsection{Multi-class Classification}

\begin{table}[!htb]
\centering
    \caption{Performance of multi-class classification based on classification models trained with different methods, on the {\em CIFAR10} data set~\cite{cifar10}. Other details are as in Figure~\ref{fig:exp_mc}. The numbers in parenthesis indicate standard errors. The numbers in bold highlight cardinality values within 1 standard error of the best cardinality across all methods, for each sample size.}
  \label{tab:exp_mc}
  
\begin{tabular}[t]{rlll}
\toprule
Sample size & Method & Cardinality & Marignal coverage\\
\midrule
\addlinespace[0.3em]
\multicolumn{4}{l}{\textbf{500}}\\
\hspace{1em}500 & CES & \textbf{6.754 (0.074)} & 0.908 (0.003)\\
\hspace{1em}500 & Naive & \textbf{6.735 (0.072)} & 0.906 (0.003)\\
\hspace{1em}500 & Naive + theory & 9.193 (0.052) & 0.988 (0.001)\\
\hspace{1em}500 & Data splitting & 7.022 (0.077) & 0.900 (0.004)\\
\hspace{1em}500 & Full training & \textbf{6.759 (0.091)} & 0.909 (0.004)\\
\addlinespace[0.3em]
\multicolumn{4}{l}{\textbf{1000}}\\
\hspace{1em}1000 & CES & \textbf{5.902 (0.060)} & 0.902 (0.003)\\
\hspace{1em}1000 & Naive & \textbf{5.908 (0.059)} & 0.901 (0.003)\\
\hspace{1em}1000 & Naive + theory & 7.767 (0.064) & 0.972 (0.002)\\
\hspace{1em}1000 & Data splitting & 6.294 (0.063) & 0.900 (0.004)\\
\hspace{1em}1000 & Full training & 6.270 (0.092) & 0.897 (0.004)\\
\addlinespace[0.3em]
\multicolumn{4}{l}{\textbf{2000}}\\
\hspace{1em}2000 & CES & \textbf{5.352 (0.045)} & 0.903 (0.003)\\
\hspace{1em}2000 & Naive & \textbf{5.347 (0.045)} & 0.902 (0.003)\\
\hspace{1em}2000 & Naive + theory & 6.609 (0.049) & 0.955 (0.002)\\
\hspace{1em}2000 & Data splitting & 5.674 (0.040) & 0.904 (0.003)\\
\hspace{1em}2000 & Full training & 7.776 (0.194) & 0.934 (0.006)\\
\bottomrule
\end{tabular}

\end{table}

\FloatBarrier
\subsection{Regression}

\begin{figure}[!htb]
    \centering
    \includegraphics[width=0.8\linewidth]{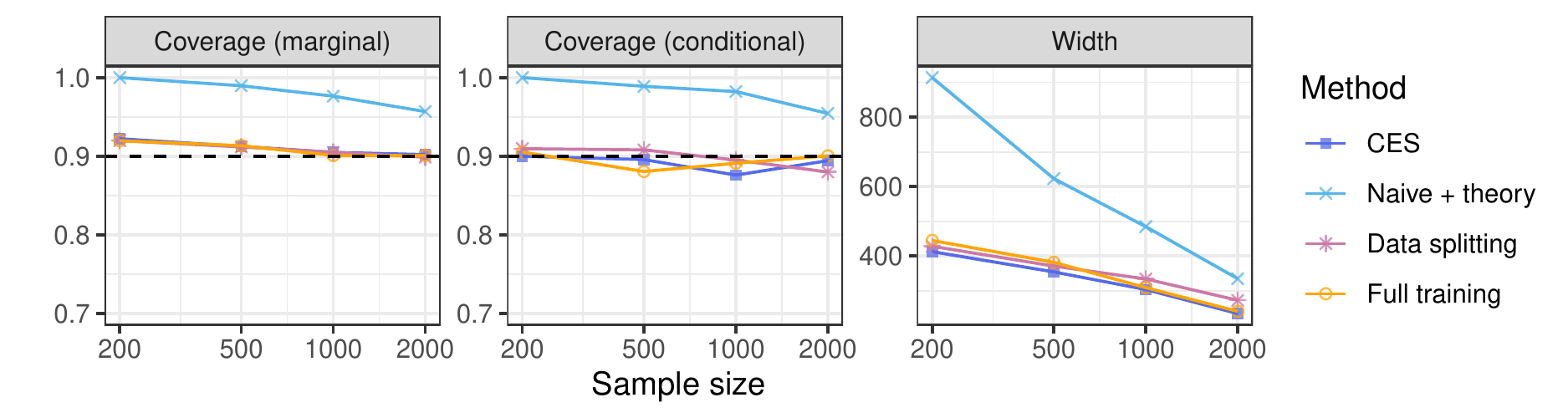}
    \caption{Performance of conformal prediction intervals based on regression models trained with different methods, on the {\em bike} data set~\cite{data-bike}. The results are shown as a function of the total sample size. The nominal marginal coverage level is 90\%. See Table~\ref{tab:exp_regression_bike} for additional details and standard errors.}
    \label{fig:exp_regression_bike}
\end{figure}

\begin{figure}[!htb]
    \centering
    \includegraphics[width=0.8\linewidth]{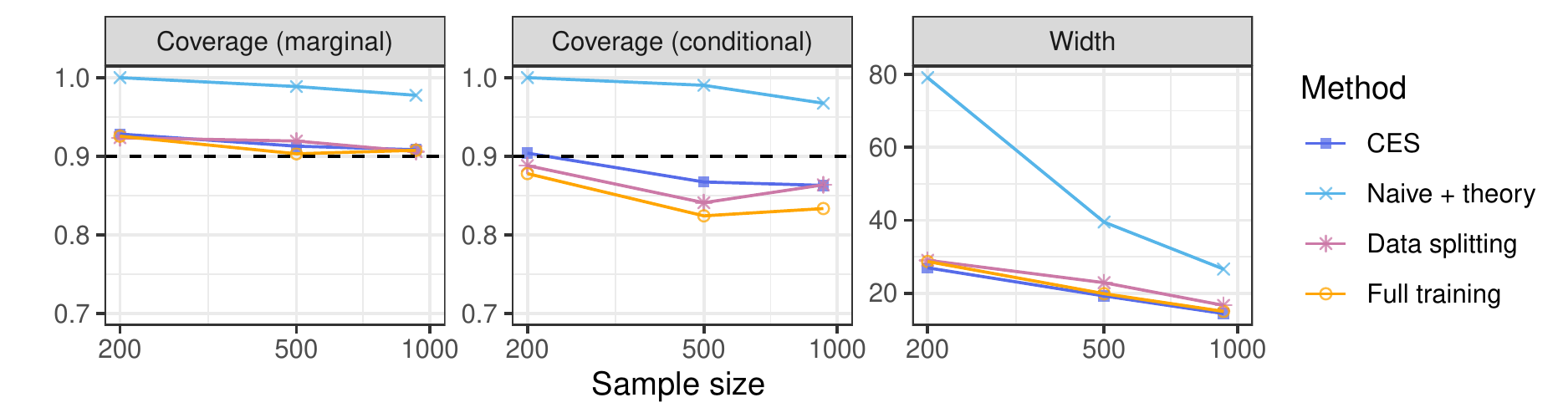}
    \caption{Performance of conformal prediction intervals based on regression models trained with different methods, on the {\em concrete} data set~\cite{data-concrete}. The results are shown as a function of the total sample size. The nominal marginal coverage level is 90\%. See Table~\ref{tab:exp_regression_concrete} for additional details and standard errors.}
    \label{fig:exp_regression_concrete}
\end{figure}

\begin{table}[!htb]
\centering
    \caption{Performance of conformal prediction intervals based on regression models trained with different methods, on the {\em bio} data set~\cite{data-bio}. Other details are as in Figure~\ref{fig:exp_regression_bio}. The numbers in parenthesis indicate standard errors. The numbers in bold highlight width values within 1 standard error of the best width across all methods, for each sample size. The numbers in red highlight coverage values below 0.85.}
  \label{tab:exp_regression_bio}
  \begin{tabular}[t]{rlllll}
\toprule
\multicolumn{4}{c}{ } & \multicolumn{2}{c}{Coverage} \\
\cmidrule(l{3pt}r{3pt}){5-6}
Sample size & Data & Method & Width & Marginal & Conditional\\
\midrule
\addlinespace[0.3em]
\multicolumn{6}{l}{\textbf{200}}\\
\hspace{1em}200 & bio & CES & \textbf{18.740 (0.123)} & 0.924 (0.005) & 0.898 (0.014)\\
\hspace{1em}200 & bio & Naive & \textbf{18.544 (0.133)} & 0.917 (0.005) & 0.899 (0.015)\\
\hspace{1em}200 & bio & Naive + theory & 20.942 (0.006) & 1.000 (0.000) & 1.000 (0.000)\\
\hspace{1em}200 & bio & Data splitting & 19.068 (0.113) & 0.925 (0.005) & 0.902 (0.015)\\
\hspace{1em}200 & bio & Full training & \textbf{18.673 (0.125)} & 0.919 (0.004) & 0.890 (0.018)\\
\addlinespace[0.3em]
\multicolumn{6}{l}{\textbf{500}}\\
\hspace{1em}500 & bio & CES & \textbf{17.435 (0.125)} & 0.914 (0.004) & 0.909 (0.013)\\
\hspace{1em}500 & bio & Naive & \textbf{17.363 (0.134)} & 0.910 (0.004) & 0.890 (0.017)\\
\hspace{1em}500 & bio & Naive + theory & 20.391 (0.061) & 0.993 (0.001) & 0.996 (0.001)\\
\hspace{1em}500 & bio & Data splitting & 18.076 (0.123) & 0.911 (0.004) & 0.888 (0.015)\\
\hspace{1em}500 & bio & Full training & 18.245 (0.129) & 0.918 (0.003) & 0.890 (0.019)\\
\addlinespace[0.3em]
\multicolumn{6}{l}{\textbf{1000}}\\
\hspace{1em}1000 & bio & CES & \textbf{16.251 (0.081)} & 0.901 (0.003) & 0.885 (0.015)\\
\hspace{1em}1000 & bio & Naive & \textbf{16.219 (0.084)} & 0.900 (0.003) & 0.890 (0.012)\\
\hspace{1em}1000 & bio & Naive + theory & 19.167 (0.073) & 0.977 (0.002) & 0.976 (0.006)\\
\hspace{1em}1000 & bio & Data splitting & 16.728 (0.089) & 0.903 (0.003) & 0.887 (0.016)\\
\hspace{1em}1000 & bio & Full training & 17.962 (0.141) & 0.902 (0.004) & \textcolor{red}{0.814 (0.022)}\\
\addlinespace[0.3em]
\multicolumn{6}{l}{\textbf{2000}}\\
\hspace{1em}2000 & bio & CES & \textbf{15.812 (0.042)} & 0.899 (0.003) & 0.893 (0.015)\\
\hspace{1em}2000 & bio & Naive & \textbf{15.805 (0.042)} & 0.899 (0.003) & 0.897 (0.014)\\
\hspace{1em}2000 & bio & Naive + theory & 17.691 (0.047) & 0.957 (0.002) & 0.963 (0.006)\\
\hspace{1em}2000 & bio & Data splitting & 16.043 (0.059) & 0.900 (0.003) & 0.903 (0.015)\\
\hspace{1em}2000 & bio & Full training & 17.014 (0.113) & 0.902 (0.003) & \textcolor{red}{0.829 (0.019)}\\
\bottomrule
\end{tabular}

\end{table}

\begin{table}[!htb]
\centering
    \caption{Performance of conformal prediction intervals based on regression models trained with different methods, on the {\em bike} data set~\cite{data-bike}. Other details are as in Figure~\ref{fig:exp_regression_bike}. The numbers in parenthesis indicate standard errors. The numbers in bold highlight width values within 1 standard error of the best width across all methods, for each sample size. The numbers in red highlight coverage values below 0.85.}
  \label{tab:exp_regression_bike}
  \begin{tabular}[t]{rlllll}
\toprule
\multicolumn{4}{c}{ } & \multicolumn{2}{c}{Coverage} \\
\cmidrule(l{3pt}r{3pt}){5-6}
Sample size & Data & Method & Width & Marginal & Conditional\\
\midrule
\addlinespace[0.3em]
\multicolumn{6}{l}{\textbf{200}}\\
\hspace{1em}200 & bike & CES & 412.534 (6.439) & 0.922 (0.004) & 0.900 (0.018)\\
\hspace{1em}200 & bike & Naive & \textbf{392.474 (6.019)} & 0.908 (0.005) & 0.878 (0.016)\\
\hspace{1em}200 & bike & Naive + theory & 913.440 (3.737) & 1.000 (0.000) & 1.000 (0.000)\\
\hspace{1em}200 & bike & Data splitting & 427.964 (7.282) & 0.920 (0.004) & 0.910 (0.016)\\
\hspace{1em}200 & bike & Full training & 444.656 (6.760) & 0.920 (0.005) & 0.905 (0.013)\\
\addlinespace[0.3em]
\multicolumn{6}{l}{\textbf{500}}\\
\hspace{1em}500 & bike & CES & 354.180 (4.183) & 0.913 (0.004) & 0.896 (0.017)\\
\hspace{1em}500 & bike & Naive & \textbf{343.641 (4.220)} & 0.902 (0.004) & 0.889 (0.018)\\
\hspace{1em}500 & bike & Naive + theory & 622.837 (8.381) & 0.990 (0.001) & 0.989 (0.004)\\
\hspace{1em}500 & bike & Data splitting & 371.079 (3.777) & 0.912 (0.004) & 0.908 (0.014)\\
\hspace{1em}500 & bike & Full training & 381.951 (5.175) & 0.913 (0.004) & 0.881 (0.018)\\
\addlinespace[0.3em]
\multicolumn{6}{l}{\textbf{1000}}\\
\hspace{1em}1000 & bike & CES & \textbf{303.516 (3.047)} & 0.905 (0.004) & 0.876 (0.017)\\
\hspace{1em}1000 & bike & Naive & \textbf{300.091 (3.127)} & 0.903 (0.004) & 0.894 (0.015)\\
\hspace{1em}1000 & bike & Naive + theory & 484.565 (4.958) & 0.977 (0.002) & 0.982 (0.003)\\
\hspace{1em}1000 & bike & Data splitting & 333.939 (2.891) & 0.905 (0.003) & 0.895 (0.018)\\
\hspace{1em}1000 & bike & Full training & 308.981 (3.760) & 0.901 (0.004) & 0.891 (0.016)\\
\addlinespace[0.3em]
\multicolumn{6}{l}{\textbf{2000}}\\
\hspace{1em}2000 & bike & CES & \textbf{234.322 (1.935)} & 0.902 (0.003) & 0.894 (0.018)\\
\hspace{1em}2000 & bike & Naive & \textbf{231.571 (1.956)} & 0.897 (0.003) & 0.893 (0.018)\\
\hspace{1em}2000 & bike & Naive + theory & 334.724 (2.988) & 0.957 (0.002) & 0.954 (0.012)\\
\hspace{1em}2000 & bike & Data splitting & 272.589 (2.532) & 0.899 (0.003) & 0.880 (0.019)\\
\hspace{1em}2000 & bike & Full training & 240.714 (2.389) & 0.901 (0.003) & 0.901 (0.015)\\
\bottomrule
\end{tabular}

\end{table}

\begin{table}[!htb]
\centering
    \caption{Performance of conformal prediction intervals based on regression models trained with different methods, on the {\em concrete} data set~\cite{data-concrete}. Other details are as in Figure~\ref{fig:exp_regression_concrete}. The numbers in parenthesis indicate standard errors. The numbers in bold highlight width values within 1 standard error of the best width across all methods, for each sample size. The numbers in red highlight coverage values below 0.85.}
  \label{tab:exp_regression_concrete}
  \begin{tabular}[t]{rlllll}
\toprule
\multicolumn{4}{c}{ } & \multicolumn{2}{c}{Coverage} \\
\cmidrule(l{3pt}r{3pt}){5-6}
Sample size & Data & Method & Width & Marginal & Conditional\\
\midrule
\addlinespace[0.3em]
\multicolumn{6}{l}{\textbf{200}}\\
\hspace{1em}200 & concrete & CES & 26.948 (0.515) & 0.928 (0.005) & 0.904 (0.016)\\
\hspace{1em}200 & concrete & Naive & \textbf{24.793 (0.520)} & 0.903 (0.006) & 0.856 (0.021)\\
\hspace{1em}200 & concrete & Naive + theory & 79.089 (0.130) & 1.000 (0.000) & 1.000 (0.000)\\
\hspace{1em}200 & concrete & Data splitting & 29.021 (0.564) & 0.924 (0.004) & 0.888 (0.018)\\
\hspace{1em}200 & concrete & Full training & 28.676 (0.568) & 0.926 (0.005) & 0.878 (0.021)\\
\addlinespace[0.3em]
\multicolumn{6}{l}{\textbf{500}}\\
\hspace{1em}500 & concrete & CES & 19.232 (0.263) & 0.913 (0.004) & 0.867 (0.018)\\
\hspace{1em}500 & concrete & Naive & \textbf{18.340 (0.276)} & 0.896 (0.004) & \textcolor{red}{0.829 (0.022)}\\
\hspace{1em}500 & concrete & Naive + theory & 39.492 (0.861) & 0.989 (0.002) & 0.990 (0.003)\\
\hspace{1em}500 & concrete & Data splitting & 22.876 (0.323) & 0.919 (0.003) & \textcolor{red}{0.841 (0.023)}\\
\hspace{1em}500 & concrete & Full training & 19.857 (0.300) & 0.903 (0.004) & \textcolor{red}{0.824 (0.024)}\\
\addlinespace[0.3em]
\multicolumn{6}{l}{\textbf{930}}\\
\hspace{1em}930 & concrete & CES & 14.399 (0.134) & 0.908 (0.002) & 0.863 (0.014)\\
\hspace{1em}930 & concrete & Naive & \textbf{13.738 (0.127)} & 0.899 (0.002) & 0.863 (0.011)\\
\hspace{1em}930 & concrete & Naive + theory & 26.596 (0.334) & 0.978 (0.001) & 0.967 (0.004)\\
\hspace{1em}930 & concrete & Data splitting & 16.659 (0.122) & 0.906 (0.002) & 0.864 (0.014)\\
\hspace{1em}930 & concrete & Full training & 14.998 (0.143) & 0.908 (0.002) & \textcolor{red}{0.834 (0.016)}\\
\bottomrule
\end{tabular}

\end{table}

\FloatBarrier
\subsection{Quantile Regression}

\begin{figure}[!htb]
    \centering
    \includegraphics[width=0.8\linewidth]{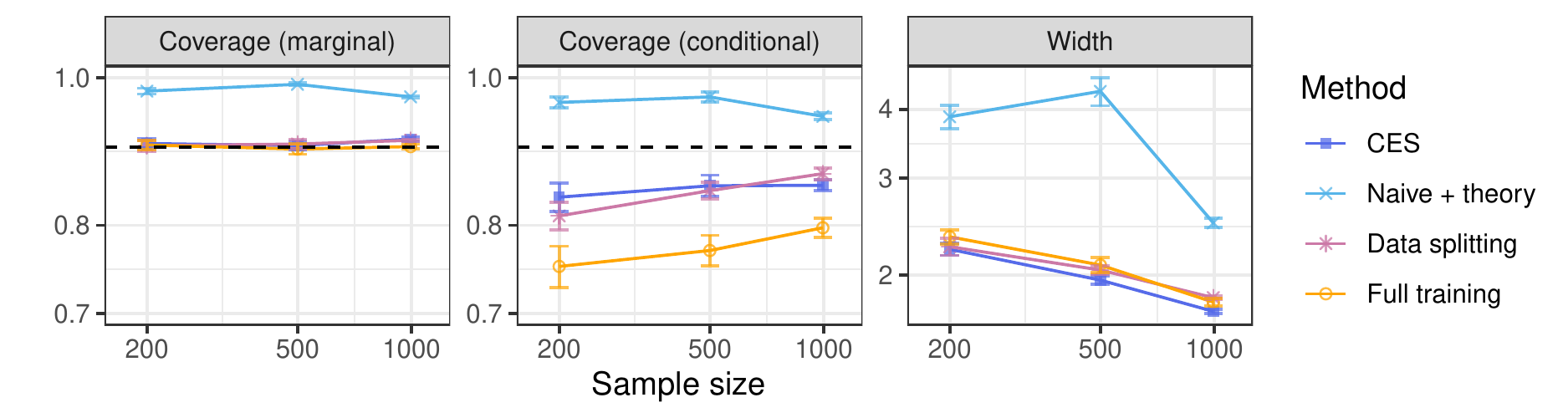}
    \caption{Performance of conformal prediction intervals based on quantile regression models trained with different methods, on the {\em community} data set~\cite{community}. The results are shown as a function of the total sample size with error bars corresponding to standard error. The nominal marginal coverage level is 90\%. See Table~\ref{tab:exp_qt_regression_community_tab} for additional details and standard errors.}
    \label{fig:exp_qt_regression_community}
\end{figure}

\begin{figure}[!htb]
    \centering
    \includegraphics[width=0.8\linewidth]{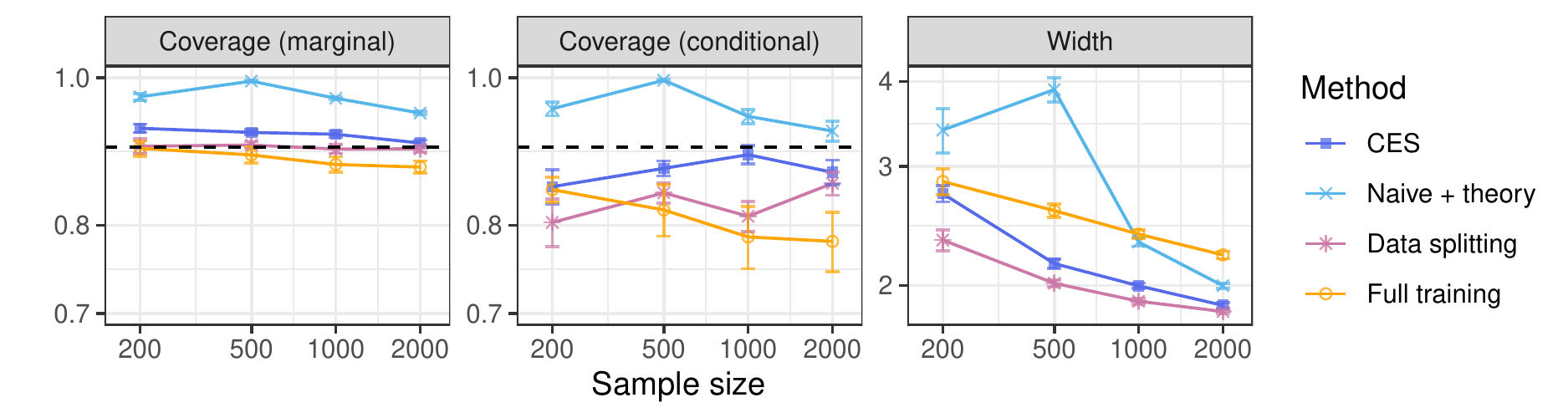}
    \caption{Performance of conformal prediction intervals based on quantile regression models trained with different methods, on the {\em bio} data set~\cite{data-bio}. The results are shown as a function of the total sample size with error bars corresponding to standard error. The nominal marginal coverage level is 90\%. See Table~\ref{tab:exp_qt_regression_bio_tab} for additional details and standard errors.}
    \label{fig:exp_qt_regression_bio}
\end{figure}

\begin{figure}[!htb]
    \centering
    \includegraphics[width=0.8\linewidth]{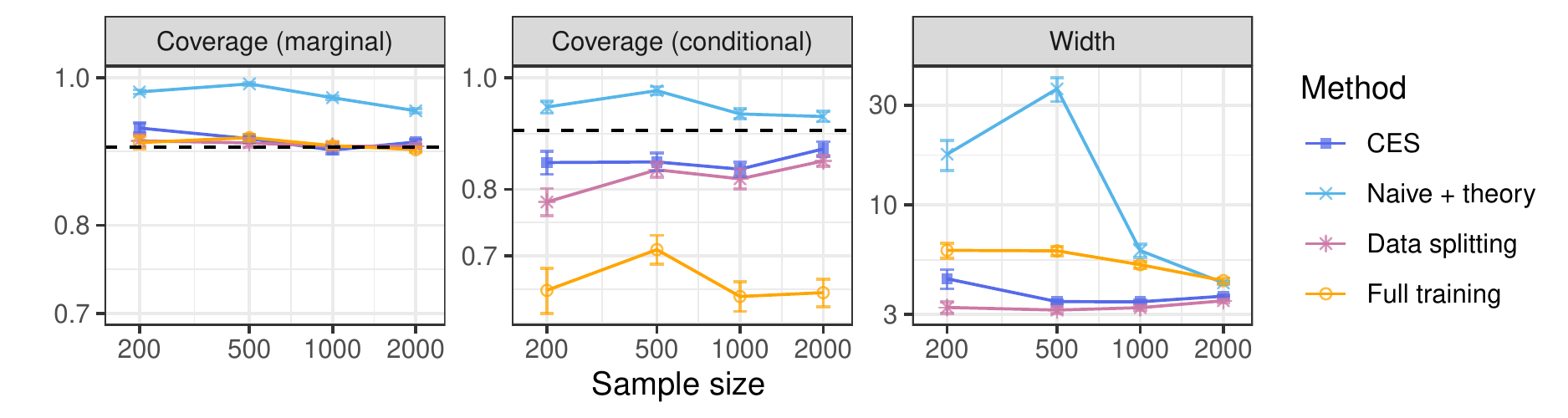}
    \caption{Performance of conformal prediction intervals based on quantile regression models trained with different methods, on the {\em meps\_21} data set~\cite{meps_21}.The results are shown as a function of the total sample size with error bars corresponding to standard error. The nominal marginal coverage level is 90\%. See Table~\ref{tab:exp_qt_regression_meps_21_tab} for additional details and standard errors.}
    \label{fig:exp_qt_regression_meps_21}
\end{figure}

\begin{figure}[!htb]
    \centering
    \includegraphics[width=0.8\linewidth]{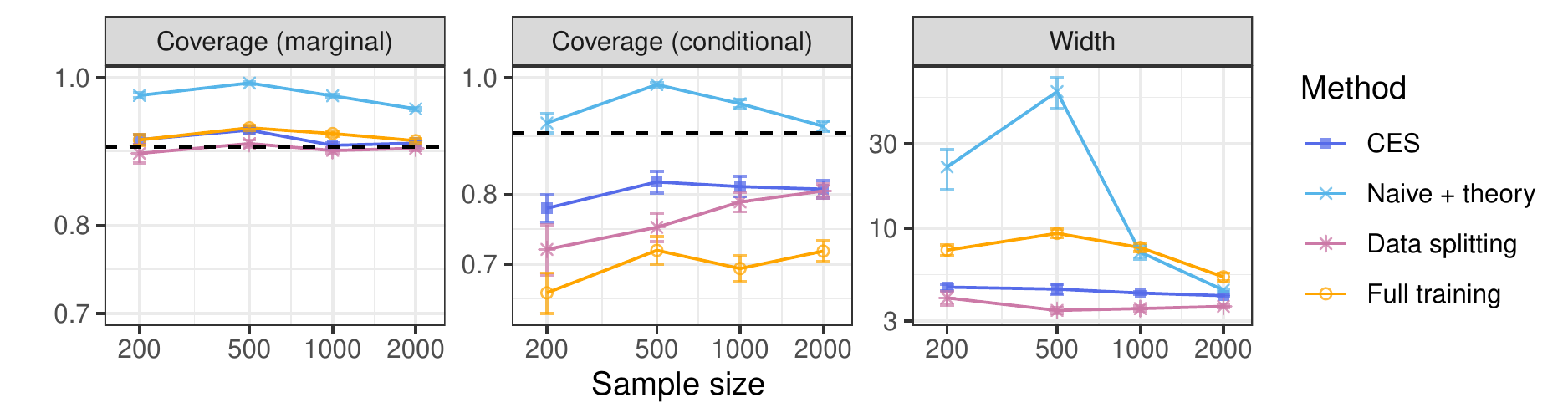}
    \caption{Performance of conformal prediction intervals based on quantile regression models trained with different methods, on the {\em blog\_data} data set~\cite{blog_data}. The results are shown as a function of the total sample size with error bars corresponding to standard error. The nominal marginal coverage level is 90\%. See Table~\ref{tab:exp_qt_regression_blog_data_tab} for additional details and standard errors.}
    \label{fig:exp_qt_regression_blog_data}
\end{figure}

\begin{figure}[!htb]
    \centering
    \includegraphics[width=0.8\linewidth]{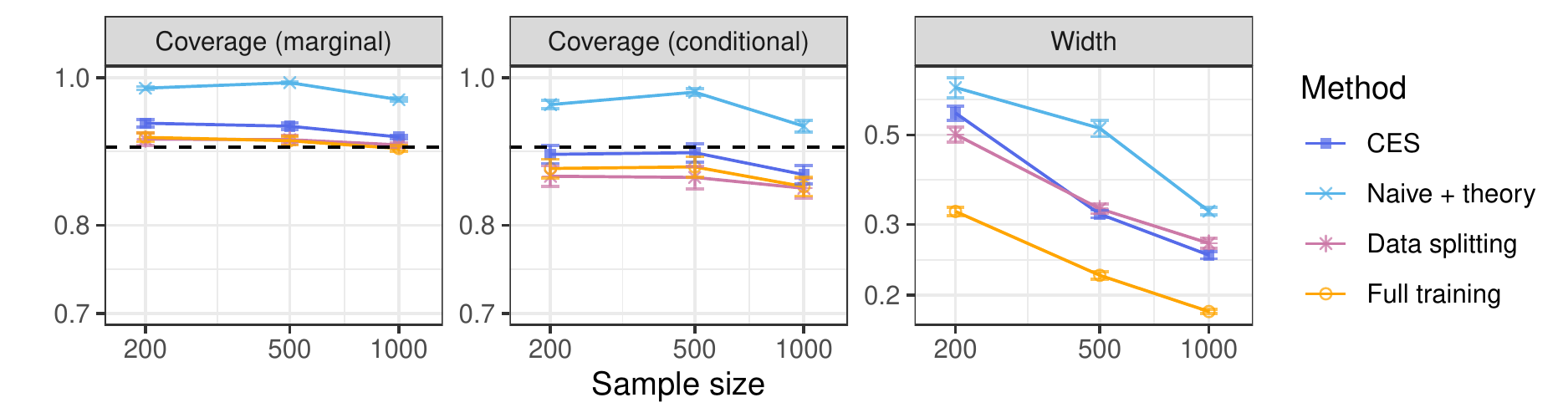}
    \caption{Performance of conformal prediction intervals based on quantile regression models trained with different methods, on the {\em STAR} data set~\cite{star}. The results are shown as a function of the total sample size with error bars corresponding to standard error. The nominal marginal coverage level is 90\%. See Table~\ref{tab:exp_qt_regression_star_tab} for additional details and standard errors.}
    \label{fig:exp_qt_regression_star}
\end{figure}

\begin{figure}[!htb]
    \centering
    \includegraphics[width=0.8\linewidth]{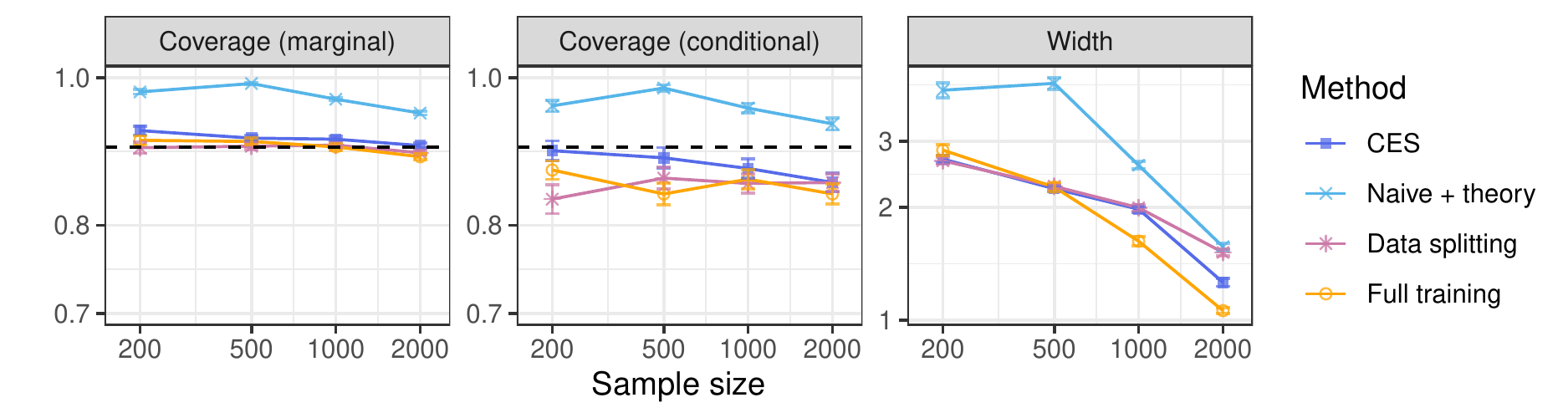}
    \caption{Performance of conformal prediction intervals based on quantile regression models trained with different methods, on the {\em bike} data set~\cite{data-bike}. The results are shown as a function of the total sample size with error bars corresponding to standard error. The nominal marginal coverage level is 90\%. See Table~\ref{tab:exp_qt_regression_bike_tab} for additional details and standard errors.}
    \label{fig:exp_qt_regression_bike}
\end{figure}

\begin{table}[!htb]
\centering
    \caption{Performance of conformal prediction intervals based on quantile regression models trained with different methods, on the {\em homes} data set~\cite{homes}. Other details are as in Figure~\ref{fig:exp_qt_regression_homes}. The numbers in parenthesis indicate standard errors. The numbers in bold highlight width values within 1 standard error of the best width across all methods, for each sample size. The numbers in red highlight coverage values below 0.85.}
    \label{tab:exp_qt_regression_homes_tab}
  
\begin{tabular}[t]{rlllll}
\toprule
\multicolumn{4}{c}{ } & \multicolumn{2}{c}{Coverage} \\
\cmidrule(l{3pt}r{3pt}){5-6}
Sample size & Data & Method & Width & Marginal & Conditional\\
\midrule
\addlinespace[0.3em]
\multicolumn{6}{l}{\textbf{200}}\\
\hspace{1em}200 & homes & CES & 1.396 (0.047) & 0.924 (0.007) & \textcolor{red}{0.804 (0.016)}\\
\hspace{1em}200 & homes & Naive & \textbf{1.171 (0.048)} & 0.904 (0.009) & \textcolor{red}{0.761 (0.029)}\\
\hspace{1em}200 & homes & Naive + theory & 2.607 (0.222) & 0.983 (0.004) & 0.931 (0.015)\\
\hspace{1em}200 & homes & Data splitting & 1.329 (0.049) & 0.907 (0.006) & \textcolor{red}{0.730 (0.029)}\\
\hspace{1em}200 & homes & Full training & \textbf{1.195 (0.053)} & 0.919 (0.008) & \textcolor{red}{0.674 (0.030)}\\
\addlinespace[0.3em]
\multicolumn{6}{l}{\textbf{500}}\\
\hspace{1em}500 & homes & CES & 0.997 (0.022) & 0.909 (0.007) & \textcolor{red}{0.842 (0.016)}\\
\hspace{1em}500 & homes & Naive & \textbf{0.940 (0.026)} & 0.899 (0.007) & \textcolor{red}{0.798 (0.018)}\\
\hspace{1em}500 & homes & Naive + theory & 2.698 (0.308) & 0.989 (0.003) & 0.950 (0.017)\\
\hspace{1em}500 & homes & Data splitting & 0.985 (0.019) & 0.896 (0.006) & \textcolor{red}{0.810 (0.020)}\\
\hspace{1em}500 & homes & Full training & \textbf{0.940 (0.043)} & 0.913 (0.007) & \textcolor{red}{0.710 (0.019)}\\
\addlinespace[0.3em]
\multicolumn{6}{l}{\textbf{1000}}\\
\hspace{1em}1000 & homes & CES & 0.858 (0.015) & 0.910 (0.005) & \textcolor{red}{0.827 (0.017)}\\
\hspace{1em}1000 & homes & Naive & 0.824 (0.017) & 0.894 (0.006) & \textcolor{red}{0.784 (0.015)}\\
\hspace{1em}1000 & homes & Naive + theory & 1.229 (0.039) & 0.969 (0.004) & 0.917 (0.015)\\
\hspace{1em}1000 & homes & Data splitting & 0.926 (0.020) & 0.905 (0.004) & \textcolor{red}{0.786 (0.021)}\\
\hspace{1em}1000 & homes & Full training & \textbf{0.755 (0.019)} & 0.902 (0.005) & \textcolor{red}{0.698 (0.019)}\\
\addlinespace[0.3em]
\multicolumn{6}{l}{\textbf{2000}}\\
\hspace{1em}2000 & homes & CES & 0.726 (0.016) & 0.902 (0.005) & \textcolor{red}{0.824 (0.013)}\\
\hspace{1em}2000 & homes & Naive & 0.667 (0.013) & 0.891 (0.003) & \textcolor{red}{0.799 (0.016)}\\
\hspace{1em}2000 & homes & Naive + theory & 0.865 (0.013) & 0.949 (0.003) & 0.880 (0.013)\\
\hspace{1em}2000 & homes & Data splitting & 0.779 (0.011) & 0.900 (0.003) & \textcolor{red}{0.804 (0.013)}\\
\hspace{1em}2000 & homes & Full training & \textbf{0.653 (0.013)} & 0.889 (0.005) & \textcolor{red}{0.686 (0.019)}\\
\bottomrule
\end{tabular}

\end{table}

\begin{table}[!htb]
\centering
    \caption{Performance of conformal prediction intervals based on quantile regression models trained with different methods, on the {\em community} data set~\cite{community}. Other details are as in Figure~\ref{fig:exp_qt_regression_community}. The numbers in parenthesis indicate standard errors. The numbers in bold highlight width values within 1 standard error of the best width across all methods, for each sample size. The numbers in red highlight coverage values below 0.85.}
    \label{tab:exp_qt_regression_community_tab}
  
\begin{tabular}[t]{rlllll}
\toprule
\multicolumn{4}{c}{ } & \multicolumn{2}{c}{Coverage} \\
\cmidrule(l{3pt}r{3pt}){5-6}
Sample size & Data & Method & Width & Marginal & Conditional\\
\midrule
\addlinespace[0.3em]
\multicolumn{6}{l}{\textbf{200}}\\
\hspace{1em}200 & community & CES & \textbf{2.226 (0.057)} & 0.905 (0.006) & \textcolor{red}{0.835 (0.018)}\\
\hspace{1em}200 & community & Naive & \textbf{2.100 (0.065)} & 0.891 (0.009) & \textcolor{red}{0.814 (0.018)}\\
\hspace{1em}200 & community & Naive + theory & 3.876 (0.193) & 0.980 (0.004) & 0.963 (0.008)\\
\hspace{1em}200 & community & Data splitting & \textbf{2.250 (0.081)} & 0.902 (0.007) & \textcolor{red}{0.811 (0.017)}\\
\hspace{1em}200 & community & Full training & 2.345 (0.070) & 0.903 (0.007) & \textcolor{red}{0.752 (0.023)}\\
\addlinespace[0.3em]
\multicolumn{6}{l}{\textbf{500}}\\
\hspace{1em}500 & community & CES & \textbf{1.957 (0.033)} & 0.902 (0.006) & \textcolor{red}{0.849 (0.014)}\\
\hspace{1em}500 & community & Naive & \textbf{1.866 (0.043)} & 0.889 (0.007) & \textcolor{red}{0.830 (0.012)}\\
\hspace{1em}500 & community & Naive + theory & 4.312 (0.254) & 0.990 (0.002) & 0.971 (0.008)\\
\hspace{1em}500 & community & Data splitting & 2.041 (0.037) & 0.904 (0.006) & \textcolor{red}{0.843 (0.011)}\\
\hspace{1em}500 & community & Full training & 2.084 (0.067) & 0.898 (0.007) & \textcolor{red}{0.770 (0.018)}\\
\addlinespace[0.3em]
\multicolumn{6}{l}{\textbf{994}}\\
\hspace{1em}994 & community & CES & \textbf{1.717 (0.016)} & 0.911 (0.003) & \textcolor{red}{0.850 (0.007)}\\
\hspace{1em}994 & community & Naive & \textbf{1.601 (0.019)} & 0.902 (0.003) & 0.853 (0.008)\\
\hspace{1em}994 & community & Naive + theory & 2.486 (0.049) & 0.971 (0.002) & 0.943 (0.005)\\
\hspace{1em}994 & community & Data splitting & 1.818 (0.013) & 0.910 (0.002) & 0.865 (0.008)\\
\hspace{1em}994 & community & Full training & 1.784 (0.029) & 0.901 (0.003) & \textcolor{red}{0.797 (0.011)}\\
\bottomrule
\end{tabular}

\end{table}

\begin{table}[!htb]
\centering
    \caption{Performance of conformal prediction intervals based on quantile regression models trained with different methods, on the {\em bio} data set~\cite{data-bio}. Other details are as in Figure~\ref{fig:exp_qt_regression_bio}. The numbers in parenthesis indicate standard errors. The numbers in bold highlight width values within 1 standard error of the best width across all methods, for each sample size. The numbers in red highlight coverage values below 0.85.}
    \label{tab:exp_qt_regression_bio_tab}
  
\begin{tabular}[t]{rlllll}
\toprule
\multicolumn{4}{c}{ } & \multicolumn{2}{c}{Coverage} \\
\cmidrule(l{3pt}r{3pt}){5-6}
Sample size & Data & Method & Width & Marginal & Conditional\\
\midrule
\addlinespace[0.3em]
\multicolumn{6}{l}{\textbf{200}}\\
\hspace{1em}200 & bio & CES & 2.732 (0.074) & 0.926 (0.006) & \textcolor{red}{0.848 (0.022)}\\
\hspace{1em}200 & bio & Naive & \textbf{2.018 (0.064)} & 0.876 (0.011) & \textcolor{red}{0.805 (0.025)}\\
\hspace{1em}200 & bio & Naive + theory & 3.392 (0.254) & 0.971 (0.005) & 0.954 (0.010)\\
\hspace{1em}200 & bio & Data splitting & \textbf{2.334 (0.083)} & 0.902 (0.010) & \textcolor{red}{0.803 (0.029)}\\
\hspace{1em}200 & bio & Full training & 2.848 (0.124) & 0.899 (0.010) & \textcolor{red}{0.844 (0.016)}\\
\addlinespace[0.3em]
\multicolumn{6}{l}{\textbf{500}}\\
\hspace{1em}500 & bio & CES & 2.154 (0.034) & 0.921 (0.004) & 0.872 (0.010)\\
\hspace{1em}500 & bio & Naive & \textbf{1.900 (0.032)} & 0.892 (0.007) & \textcolor{red}{0.839 (0.014)}\\
\hspace{1em}500 & bio & Naive + theory & 3.893 (0.161) & 0.995 (0.001) & 0.996 (0.002)\\
\hspace{1em}500 & bio & Data splitting & \textbf{2.015 (0.024)} & 0.903 (0.004) & \textcolor{red}{0.840 (0.013)}\\
\hspace{1em}500 & bio & Full training & 2.579 (0.057) & 0.890 (0.011) & \textcolor{red}{0.819 (0.032)}\\
\addlinespace[0.3em]
\multicolumn{6}{l}{\textbf{1000}}\\
\hspace{1em}1000 & bio & CES & 1.997 (0.019) & 0.918 (0.004) & 0.890 (0.013)\\
\hspace{1em}1000 & bio & Naive & \textbf{1.871 (0.018)} & 0.894 (0.005) & \textcolor{red}{0.838 (0.018)}\\
\hspace{1em}1000 & bio & Naive + theory & 2.318 (0.033) & 0.969 (0.002) & 0.943 (0.010)\\
\hspace{1em}1000 & bio & Data splitting & \textbf{1.895 (0.018)} & 0.898 (0.006) & \textcolor{red}{0.811 (0.018)}\\
\hspace{1em}1000 & bio & Full training & 2.381 (0.035) & 0.877 (0.010) & \textcolor{red}{0.786 (0.037)}\\
\addlinespace[0.3em]
\multicolumn{6}{l}{\textbf{2000}}\\
\hspace{1em}2000 & bio & CES & 1.869 (0.018) & 0.906 (0.004) & 0.867 (0.016)\\
\hspace{1em}2000 & bio & Naive & \textbf{1.763 (0.017)} & 0.890 (0.004) & \textcolor{red}{0.834 (0.016)}\\
\hspace{1em}2000 & bio & Naive + theory & 1.998 (0.017) & 0.948 (0.003) & 0.922 (0.014)\\
\hspace{1em}2000 & bio & Data splitting & \textbf{1.830 (0.013)} & 0.898 (0.004) & 0.852 (0.015)\\
\hspace{1em}2000 & bio & Full training & 2.219 (0.029) & 0.874 (0.008) & \textcolor{red}{0.781 (0.035)}\\
\bottomrule
\end{tabular}

\end{table}

\begin{table}[!htb]
\centering
    \caption{Performance of conformal prediction intervals based on quantile regression models trained with different methods, on the {\em MEPS\_21} data set~\cite{meps_21}. Other details are as in Figure~\ref{fig:exp_qt_regression_meps_21}. The numbers in parenthesis indicate standard errors. The numbers in bold highlight width values within 1 standard error of the best width across all methods, for each sample size. The numbers in red highlight coverage values below 0.85.}
    \label{tab:exp_qt_regression_meps_21_tab}
  
\begin{tabular}[t]{rlllll}
\toprule
\multicolumn{4}{c}{ } & \multicolumn{2}{c}{Coverage} \\
\cmidrule(l{3pt}r{3pt}){5-6}
Sample size & Data & Method & Width & Marginal & Conditional\\
\midrule
\addlinespace[0.3em]
\multicolumn{6}{l}{\textbf{200}}\\
\hspace{1em}200 & meps\_21 & CES & 4.442 (0.468) & 0.927 (0.007) & \textcolor{red}{0.844 (0.020)}\\
\hspace{1em}200 & meps\_21 & Naive & \textbf{3.602 (0.452)} & 0.901 (0.010) & \textcolor{red}{0.786 (0.025)}\\
\hspace{1em}200 & meps\_21 & Naive + theory & 17.470 (2.874) & 0.979 (0.003) & 0.943 (0.012)\\
\hspace{1em}200 & meps\_21 & Data splitting & \textbf{3.239 (0.212)} & 0.909 (0.008) & \textcolor{red}{0.780 (0.021)}\\
\hspace{1em}200 & meps\_21 & Full training & 6.070 (0.509) & 0.906 (0.009) & \textcolor{red}{0.654 (0.030)}\\
\addlinespace[0.3em]
\multicolumn{6}{l}{\textbf{500}}\\
\hspace{1em}500 & meps\_21 & CES & 3.453 (0.074) & 0.912 (0.004) & \textcolor{red}{0.845 (0.015)}\\
\hspace{1em}500 & meps\_21 & Naive & \textbf{3.077 (0.068)} & 0.898 (0.005) & \textcolor{red}{0.802 (0.012)}\\
\hspace{1em}500 & meps\_21 & Naive + theory & 35.882 (4.705) & 0.991 (0.002) & 0.974 (0.008)\\
\hspace{1em}500 & meps\_21 & Data splitting & \textbf{3.147 (0.109)} & 0.906 (0.006) & \textcolor{red}{0.832 (0.012)}\\
\hspace{1em}500 & meps\_21 & Full training & 6.038 (0.305) & 0.913 (0.005) & \textcolor{red}{0.709 (0.021)}\\
\addlinespace[0.3em]
\multicolumn{6}{l}{\textbf{1000}}\\
\hspace{1em}1000 & meps\_21 & CES & 3.447 (0.068) & 0.896 (0.005) & \textcolor{red}{0.833 (0.012)}\\
\hspace{1em}1000 & meps\_21 & Naive & \textbf{3.102 (0.073)} & 0.888 (0.006) & \textcolor{red}{0.825 (0.014)}\\
\hspace{1em}1000 & meps\_21 & Naive + theory & 6.054 (0.443) & 0.970 (0.002) & 0.930 (0.009)\\
\hspace{1em}1000 & meps\_21 & Data splitting & \textbf{3.231 (0.093)} & 0.901 (0.005) & \textcolor{red}{0.817 (0.016)}\\
\hspace{1em}1000 & meps\_21 & Full training & 5.183 (0.188) & 0.902 (0.004) & \textcolor{red}{0.646 (0.019)}\\
\addlinespace[0.3em]
\multicolumn{6}{l}{\textbf{2000}}\\
\hspace{1em}2000 & meps\_21 & CES & 3.666 (0.054) & 0.907 (0.004) & 0.867 (0.012)\\
\hspace{1em}2000 & meps\_21 & Naive & \textbf{3.454 (0.047)} & 0.902 (0.004) & 0.875 (0.013)\\
\hspace{1em}2000 & meps\_21 & Naive + theory & 4.242 (0.069) & 0.951 (0.003) & 0.925 (0.010)\\
\hspace{1em}2000 & meps\_21 & Data splitting & \textbf{3.485 (0.060)} & 0.902 (0.003) & \textcolor{red}{0.847 (0.009)}\\
\hspace{1em}2000 & meps\_21 & Full training & 4.349 (0.128) & 0.897 (0.003) & \textcolor{red}{0.651 (0.018)}\\
\bottomrule
\end{tabular}


\end{table}

\begin{table}[!htb]
\centering
    \caption{Performance of conformal prediction intervals based on quantile regression models trained with different methods, on the {\em blog\_data} data set~\cite{blog_data}. Other details are as in Figure~\ref{fig:exp_qt_regression_blog_data}. The numbers in parenthesis indicate standard errors. The numbers in bold highlight width values within 1 standard error of the best width across all methods, for each sample size. The numbers in red highlight coverage values below 0.85.}
    \label{tab:exp_qt_regression_blog_data_tab}
  
\begin{tabular}[t]{rlllll}
\toprule
\multicolumn{4}{c}{ } & \multicolumn{2}{c}{Coverage} \\
\cmidrule(l{3pt}r{3pt}){5-6}
Sample size & Data & Method & Width & Marginal & Conditional\\
\midrule
\addlinespace[0.3em]
\multicolumn{6}{l}{\textbf{200}}\\
\hspace{1em}200 & blog\_data & CES & 4.658 (0.194) & 0.911 (0.007) & \textcolor{red}{0.779 (0.021)}\\
\hspace{1em}200 & blog\_data & Naive & \textbf{3.427 (0.210)} & 0.897 (0.008) & \textcolor{red}{0.766 (0.021)}\\
\hspace{1em}200 & blog\_data & Naive + theory & 22.166 (5.645) & 0.973 (0.004) & 0.917 (0.017)\\
\hspace{1em}200 & blog\_data & Data splitting & \textbf{4.049 (0.359)} & 0.892 (0.013) & \textcolor{red}{0.720 (0.034)}\\
\hspace{1em}200 & blog\_data & Full training & 7.524 (0.544) & 0.910 (0.007) & \textcolor{red}{0.662 (0.026)}\\
\addlinespace[0.3em]
\multicolumn{6}{l}{\textbf{500}}\\
\hspace{1em}500 & blog\_data & CES & 4.538 (0.291) & 0.924 (0.005) & \textcolor{red}{0.819 (0.017)}\\
\hspace{1em}500 & blog\_data & Naive & 3.986 (0.333) & 0.917 (0.005) & \textcolor{red}{0.796 (0.014)}\\
\hspace{1em}500 & blog\_data & Naive + theory & 58.965 (11.745) & 0.992 (0.002) & 0.987 (0.004)\\
\hspace{1em}500 & blog\_data & Data splitting & \textbf{3.443 (0.139)} & 0.905 (0.005) & \textcolor{red}{0.751 (0.020)}\\
\hspace{1em}500 & blog\_data & Full training & 9.364 (0.535) & 0.927 (0.003) & \textcolor{red}{0.719 (0.019)}\\
\addlinespace[0.3em]
\multicolumn{6}{l}{\textbf{1000}}\\
\hspace{1em}1000 & blog\_data & CES & 4.313 (0.088) & 0.903 (0.004) & \textcolor{red}{0.812 (0.016)}\\
\hspace{1em}1000 & blog\_data & Naive & \textbf{3.559 (0.073)} & 0.896 (0.004) & \textcolor{red}{0.816 (0.016)}\\
\hspace{1em}1000 & blog\_data & Naive + theory & 7.336 (0.620) & 0.973 (0.002) & 0.951 (0.007)\\
\hspace{1em}1000 & blog\_data & Data splitting & \textbf{3.525 (0.119)} & 0.896 (0.003) & \textcolor{red}{0.788 (0.015)}\\
\hspace{1em}1000 & blog\_data & Full training & 7.784 (0.427) & 0.919 (0.003) & \textcolor{red}{0.694 (0.018)}\\
\addlinespace[0.3em]
\multicolumn{6}{l}{\textbf{2000}}\\
\hspace{1em}2000 & blog\_data & CES & 4.169 (0.066) & 0.906 (0.003) & \textcolor{red}{0.808 (0.013)}\\
\hspace{1em}2000 & blog\_data & Naive & \textbf{3.671 (0.047)} & 0.899 (0.003) & \textcolor{red}{0.783 (0.018)}\\
\hspace{1em}2000 & blog\_data & Naive + theory & 4.486 (0.074) & 0.954 (0.002) & 0.911 (0.009)\\
\hspace{1em}2000 & blog\_data & Data splitting & \textbf{3.625 (0.069)} & 0.899 (0.003) & \textcolor{red}{0.805 (0.011)}\\
\hspace{1em}2000 & blog\_data & Full training & 5.323 (0.274) & 0.909 (0.003) & \textcolor{red}{0.718 (0.014)}\\
\bottomrule
\end{tabular}

\end{table}

\begin{table}[!htb]
\centering
    \caption{Performance of conformal prediction intervals based on quantile regression models trained with different methods, on the {\em STAR} data set~\cite{star}. Other details are as in Figure~\ref{fig:exp_qt_regression_star}. The numbers in parenthesis indicate standard errors. The numbers in bold highlight width values within 1 standard error of the best width across all methods, for each sample size. The numbers in red highlight coverage values below 0.85.}
    \label{tab:exp_qt_regression_star_tab}
  
\begin{tabular}[t]{rlllll}
\toprule
\multicolumn{4}{c}{ } & \multicolumn{2}{c}{Coverage} \\
\cmidrule(l{3pt}r{3pt}){5-6}
Sample size & Data & Method & Width & Marginal & Conditional\\
\midrule
\addlinespace[0.3em]
\multicolumn{6}{l}{\textbf{200}}\\
\hspace{1em}200 & star & CES & 0.567 (0.023) & 0.933 (0.006) & 0.891 (0.012)\\
\hspace{1em}200 & star & Naive & 0.421 (0.031) & 0.904 (0.008) & \textcolor{red}{0.847 (0.012)}\\
\hspace{1em}200 & star & Naive + theory & 0.656 (0.038) & 0.984 (0.002) & 0.960 (0.006)\\
\hspace{1em}200 & star & Data splitting & 0.502 (0.022) & 0.911 (0.009) & 0.862 (0.013)\\
\hspace{1em}200 & star & Full training & \textbf{0.323 (0.008)} & 0.914 (0.006) & 0.872 (0.012)\\
\addlinespace[0.3em]
\multicolumn{6}{l}{\textbf{500}}\\
\hspace{1em}500 & star & CES & 0.318 (0.007) & 0.929 (0.004) & 0.893 (0.012)\\
\hspace{1em}500 & star & Naive & 0.288 (0.008) & 0.902 (0.007) & \textcolor{red}{0.848 (0.016)}\\
\hspace{1em}500 & star & Naive + theory & 0.520 (0.025) & 0.992 (0.001) & 0.978 (0.005)\\
\hspace{1em}500 & star & Data splitting & 0.328 (0.009) & 0.910 (0.006) & 0.860 (0.015)\\
\hspace{1em}500 & star & Full training & \textbf{0.224 (0.005)} & 0.909 (0.006) & 0.874 (0.013)\\
\addlinespace[0.3em]
\multicolumn{6}{l}{\textbf{1000}}\\
\hspace{1em}1000 & star & CES & 0.252 (0.005) & 0.914 (0.003) & 0.864 (0.012)\\
\hspace{1em}1000 & star & Naive & 0.236 (0.007) & 0.897 (0.003) & 0.862 (0.011)\\
\hspace{1em}1000 & star & Naive + theory & 0.323 (0.007) & 0.967 (0.003) & 0.929 (0.009)\\
\hspace{1em}1000 & star & Data splitting & 0.269 (0.007) & 0.903 (0.003) & \textcolor{red}{0.846 (0.012)}\\
\hspace{1em}1000 & star & Full training & \textbf{0.182 (0.002)} & 0.898 (0.004) & \textcolor{red}{0.848 (0.012)}\\
\addlinespace[0.3em]
\multicolumn{6}{l}{\textbf{1161}}\\
\hspace{1em}1161 & star & CES & 0.237 (0.005) & 0.920 (0.004) & 0.882 (0.012)\\
\hspace{1em}1161 & star & Naive & 0.220 (0.005) & 0.900 (0.005) & 0.860 (0.012)\\
\hspace{1em}1161 & star & Naive + theory & 0.308 (0.007) & 0.969 (0.002) & 0.944 (0.007)\\
\hspace{1em}1161 & star & Data splitting & 0.254 (0.007) & 0.903 (0.004) & 0.864 (0.014)\\
\hspace{1em}1161 & star & Full training & \textbf{0.174 (0.003)} & 0.896 (0.004) & 0.863 (0.013)\\
\bottomrule
\end{tabular}

\end{table}

\begin{table}[!htb]
\centering
    \caption{Performance of conformal prediction intervals based on quantile regression models trained with different methods, on the {\em bike} data set~\cite{data-bike}. Other details are as in Figure~\ref{fig:exp_qt_regression_bike}. The numbers in parenthesis indicate standard errors. The numbers in bold highlight width values within 1 standard error of the best width across all methods, for each sample size. The numbers in red highlight coverage values below 0.85.}
    \label{tab:exp_qt_regression_bike_tab}
  \begin{tabular}[t]{rlllll}
\toprule
\multicolumn{4}{c}{ } & \multicolumn{2}{c}{Coverage} \\
\cmidrule(l{3pt}r{3pt}){5-6}
Sample size & Data & Method & Width & Marginal & Conditional\\
\midrule
\addlinespace[0.3em]
\multicolumn{6}{l}{\textbf{200}}\\
\hspace{1em}200 & bike & CES & \textbf{2.690 (0.065)} & 0.923 (0.006) & 0.895 (0.014)\\
\hspace{1em}200 & bike & Naive & \textbf{2.535 (0.053)} & 0.905 (0.008) & 0.868 (0.015)\\
\hspace{1em}200 & bike & Naive + theory & 4.106 (0.192) & 0.979 (0.003) & 0.958 (0.008)\\
\hspace{1em}200 & bike & Data splitting & \textbf{2.663 (0.056)} & 0.899 (0.008) & \textcolor{red}{0.832 (0.018)}\\
\hspace{1em}200 & bike & Full training & 2.843 (0.095) & 0.910 (0.006) & 0.870 (0.012)\\
\addlinespace[0.3em]
\multicolumn{6}{l}{\textbf{500}}\\
\hspace{1em}500 & bike & CES & \textbf{2.244 (0.039)} & 0.913 (0.004) & 0.886 (0.014)\\
\hspace{1em}500 & bike & Naive & \textbf{2.137 (0.047)} & 0.901 (0.006) & 0.880 (0.012)\\
\hspace{1em}500 & bike & Naive + theory & 4.284 (0.152) & 0.991 (0.002) & 0.984 (0.005)\\
\hspace{1em}500 & bike & Data splitting & \textbf{2.275 (0.032)} & 0.902 (0.004) & 0.859 (0.014)\\
\hspace{1em}500 & bike & Full training & \textbf{2.265 (0.062)} & 0.908 (0.004) & \textcolor{red}{0.839 (0.014)}\\
\addlinespace[0.3em]
\multicolumn{6}{l}{\textbf{1000}}\\
\hspace{1em}1000 & bike & CES & 1.978 (0.036) & 0.911 (0.004) & 0.872 (0.013)\\
\hspace{1em}1000 & bike & Naive & 1.779 (0.057) & 0.893 (0.004) & \textcolor{red}{0.821 (0.016)}\\
\hspace{1em}1000 & bike & Naive + theory & 2.590 (0.061) & 0.968 (0.002) & 0.955 (0.007)\\
\hspace{1em}1000 & bike & Data splitting & 1.996 (0.040) & 0.903 (0.004) & 0.852 (0.013)\\
\hspace{1em}1000 & bike & Full training & \textbf{1.625 (0.045)} & 0.901 (0.005) & 0.858 (0.013)\\
\addlinespace[0.3em]
\multicolumn{6}{l}{\textbf{2000}}\\
\hspace{1em}2000 & bike & CES & 1.262 (0.030) & 0.902 (0.003) & 0.854 (0.012)\\
\hspace{1em}2000 & bike & Naive & 1.113 (0.029) & 0.884 (0.004) & \textcolor{red}{0.847 (0.012)}\\
\hspace{1em}2000 & bike & Naive + theory & 1.570 (0.033) & 0.948 (0.003) & 0.933 (0.009)\\
\hspace{1em}2000 & bike & Data splitting & 1.517 (0.033) & 0.893 (0.003) & 0.853 (0.011)\\
\hspace{1em}2000 & bike & Full training & \textbf{1.061 (0.019)} & 0.887 (0.004) & \textcolor{red}{0.839 (0.013)}\\
\bottomrule
\end{tabular}

\end{table}

\clearpage

\section{Mathematical Proofs} \label{appendix:proofs}

\begin{proof}[Proof of Theorem \ref{thm:od_full}]
It suffices to show that the nonconformity scores $\hat{S}_i$ for $i \in \{n+1\}\cup \mathcal{D}_{\text{es-cal}}$ are exchangeable. In fact, if the nonconformity scores are almost-surely unique, this implies the rank of $\hat{S}_{n+1}$ is uniformly distributed over $\{\hat{S}_{i}\}_{i \in \{n+1\}\cup \mathcal{D}_{\text{es-cal}}}$, and in that case the conformal p-value is uniformly distributed over $\{ 1/(1+|\mathcal{D}_{\text{es-cal}}|), 2/(1+|\mathcal{D}_{\text{es-cal}}|), \dots, 1\}$. If the nonconformity scores are not almost-surely unique and ties are not broken at random, then the distribution of the conformal p-value becomes stochastically larger than uniform, in which case the result still holds.
To prove the exchangeability of the nonconformity scores, let $\sigma$ be any permutation of $\{n+1\}\cup \mathcal{D}_{\text{es-cal}}$, and imagine applying Algorithm~\ref{alg:od_full_seq}, with the same random seed, to the shuffled data set indexed by $\sigma(\{n+1\}\cup \mathcal{D}_{\text{es-cal}})$, which has the same distribution as the original data set. To clarify the notation, we will refer to quantities computed under this data shuffling scenario with their usual symbol followed by an apostrophe; i.e., $M'_{t_1}$ instead of $M_{t_1}$.
As the gradient updates only involve the unperturbed observations in $\mathcal{D}_{\text{train}}$ and the maximum number of epochs $t^{\text{max}}$ is fixed, the sequence of saved models remains exactly the same under this scenario: $(M'_{t_1} , \dots, M'_{t_T}) = (M_{t_1} , \dots, M_{t_T})$.
Further, the loss function in~\eqref{eq:loss-ces} is also invariant to permutations of $\{n+1\}\cup \mathcal{D}_{\text{es-cal}}$, in the sense that $\mathcal{L}_{\text{es-cal}}^{+1'} = \mathcal{L}_{\text{es-cal}}^{+1}$, because $\mathcal{L}$ is additive.
Therefore, the model selected according to \eqref{eq:ces-model} is also invariant, $\hat{M}'_{\text{ces}} = \hat{M}_{\text{ces}}$, which implies the nonconformity scores are simply re-ordered: $\hat{S}'_{\sigma(i)} = \hat{S}_{i}$.
Therefore, we have:
\begin{align*}
  \sigma(\{\hat{S}_i\}_{i \in \{n+1\}\cup \mathcal{D}_{\text{cal}}})
  &= \{\hat{S}'_i\}_{i \in \{n+1\}\cup \mathcal{D}_{\text{cal}}} \\
  & \overset{d}{=} \{\hat{S}_i\}_{i \in \{n+1\}\cup \mathcal{D}_{\text{cal}}},
\end{align*}
where the last equality follows from the initial data exchangeability assumption.
\end{proof}

\begin{proof}[Proof of Theorem \ref{thm:class_full}]
Note that, conditional on $Y_{n+1}=y$, the miscoverage event $Y_{n+1} \not\in \hat{\mathcal{C}}_{\alpha}(X_{n+1})$ occurs if and only if $\hat{u}_y(X_{n+1}) \leq \alpha$, where $\hat{u}_y(X_{n+1})$ is defined as in~\eqref{eq:conformal_pval-class}.
Therefore, it suffices to show $\mathbb{P}\left[\hat{u}_y(X_{n+1}) \leq \alpha \mid Y_{n+1}=y\right] \leq \alpha$ for any $\alpha \in (0,1)$.
However, this is directly implied by Theorem \ref{thm:od_full}, because the $\hat{u}_y(X_{n+1})$ calculated by Algorithm~\ref{alg:class_full_seq} is equivalent to the conformal p-value $\hat{u}_0(Z_{n+1})$ given by Algorithm~\ref{alg:od_full_seq} applied to the subset of the data in $\mathcal{D}_{\text{es-cal}}$ with $Y_{i} = y$, with the understanding that $Z_i=(X_i,Y_i)$ for all $i \in \{n+1\} \cup \mathcal{D}_{\text{es-cal}}$.
\end{proof}

\begin{proof}[Proof of Theorem \ref{thm:class_full_marg}]
Note that $Y_{n+1} \not\in \hat{\mathcal{C}}^{\text{m}}_{\alpha}(X_{n+1})$ if and only if $\hat{u}^{\text{marg}}(X_{n+1};Y_{n+1}) \leq \alpha$, where $\hat{u}^{\text{marg}}(X_{n+1};Y_{n+1})$ is defined as in~\eqref{eq:conformal_pval-class_marg}. Hence it suffices to show that $\mathbb{P}\left[\hat{u}^{\text{marg}}(X_{n+1};Y_{n+1}) \leq \alpha \right] \leq \alpha$ for any $\alpha \in (0,1)$.
This can be established using the same approach as in the proof of Theorem~\ref{thm:od_full}, setting $Z_i=(X_i,Y_i)$ for all $i \in \{n+1\} \cup \mathcal{D}_{\text{es-cal}}$.
In fact, the maximum number of epochs $t^{\text{max}}$ is fixed, the sequence of saved models is invariant to permutations of $\{n+1\} \cup \mathcal{D}_{\text{es-cal}}$, and the model $\hat{M}_{\text{ces}}$ selected according to \eqref{eq:ces-model-class} is also invariant.
Thus, it follows that the nonconformity scores $\hat{S}_i$ are exchangeable with one another for all $i \in \{n+1\} \cup \mathcal{D}_{\text{es-cal}}$.
\end{proof}

\begin{proof}[Proof of Theorem \ref{thm:reg}]
Consider an imaginary oracle algorithm producing an interval $\hat{C}^{\text{oracle}}_{\alpha}(X_{n+1})$ defined as $\hat{C}^{\text{oracle}}_{\alpha}(X_{n+1}) = \mathcal{B}_{l^*(Y_{n+1})} \bigcap \hat{C}_{\alpha}(X_{n+1}, \mathcal{B}_{l^*(Y_{n+1})})$, where $l^*(Y_{n+1})$ is the exact index of the bin $\mathcal{B}_l$ to which the true $Y_{n+1}$ belongs. Clearly, this oracle is just a theoretical tool, not a practical method because the outcome value for the test point is unknown.
However, this oracle is useful because it is easier to analyze, and it suffices to establish that $\mathbb{P}[Y_{n+1} \in \hat{C}^{\text{oracle}}_{\alpha}(X_{n+1})] \geq 1-\alpha$, for any $\alpha \in (0,1)$, since $\hat{C}_{\alpha}(X_{n+1}) \supseteq \hat{C}^{\text{oracle}}_{\alpha}(X_{n+1})$ almost-surely.
The coverage property for the oracle can be established using an approach similar to that of the proof of Theorem~\ref{thm:od_full}, setting $Z_i=(X_i,Y_i)$ for all $i \in \{n+1\} \cup \mathcal{D}_{\text{es-cal}}$.
In fact, the maximum number of epochs $t^{\text{max}}$ is fixed, the sequence of saved models is invariant to permutations of $\{n+1\} \cup \mathcal{D}_{\text{es-cal}}$, and the  model $\hat{M}_{\text{ces}}$ selected by the oracle according to \eqref{eq:reg-step-func} is also invariant.
Thus, it follows that the oracle nonconformity scores $\hat{S}_i^* = \hat{S}_i(X_{n+1}, \mathcal{B}_{l^*(Y_{n+1})})$ are exchangeable with one another for all $i \in \{n+1\} \cup \mathcal{D}_{\text{es-cal}}$.
Further, by construction of the prediction intervals~\eqref{eq:reg-int-tmp}, we know that the miscoverage event $Y_{n+1} \not\in \hat{C}^{\text{oracle}}_{\alpha}(X_{n+1})$ occurs if and only if $\hat{S}^*_i > \hat{Q}^*_{1-\alpha}$, where $\hat{Q}^*_{1-\alpha}$ is the $\lceil (1-\alpha)(1+|\mathcal{D}_{\text{es-cal}}|) \rceil$-th smallest value among all nonconformity scores $\hat{S}_i(X_{n+1},\mathcal{B}_l)$.
However, it is a well-known exchangeability result that $\mathbb{P}[\hat{S}^*_i \leq \hat{Q}^*_{1-\alpha}] \geq 1-\alpha$; see for example Lemma 1 in \citet{romano2019conformalized}.
\end{proof}

\begin{proof}[Proof of Theorem \ref{thm:reg-cqr}]
Same as the proof of Theorem \ref{thm:reg}.
\end{proof}

\begin{proof}[Proof of Corollary~\ref{thm:reg-noempty}]
This corollary follows immediately from Theorem~\ref{thm:reg} because the prediction interval given by Algorithm~\ref{alg:reg-noempty} is always contained in that output by Algorithm~\ref{alg:reg}.
\end{proof}

\begin{proof}[Proof of Corollary~\ref{thm:reg-noempty-cqr}]
Same as the proof of Corollary~\ref{thm:reg-noempty}.
\end{proof}

\begin{proof}[Proof of Proposition~\ref{prop:naive-od}]
Note that $\hat{u}_0^{\textup{naive}}(Z_{n+1}) = \hat{u}_0^{\textup{naive}}(Z_{n+1};t^*)$, hence
    \begin{align*}
        \P{\hat{u}_0^{\textup{naive}}(Z_{n+1}) > \alpha} &= \mathbb{E} \bigl [\P{\hat{u}_0^{\textup{naive}}(Z_{n+1};t^*) > \alpha \mid \mathcal{D}_{\text{es-cal}}} \bigr ] \\
        &\geq \E{\min_{t\in [T]} W_t}        \\
        & \geq \sup_{a \in [0,1]} a \cdot \P{\min_{t\in [T]} W_t \geq a}\\
        &= \sup_{a \in [0,1]} a \left(1-\P{\min_{t\in [T]} W_t \leq a} \right)  \\
        &\geq \sup_{a \in [0,1]} a \left(1-T \cdot \P{W_t \leq a} \right),
   \end{align*}
where the last inequality follows from a union bound. To simplify the right-hand-side term above, let $a= I^{-1} \left(\frac{1}{bT};n_{\text{es-cal}}+1-l,l \right)$, where $b$ is any large constant. Hence we obtain
\begin{align*}
    \P{\hat{u}_0^{\textup{naive}}(Z_{n+1}) > \alpha} \geq I^{-1} \left(\frac{1}{bT};n_{\text{es-cal}}+1-l,l \right) \cdot (1-1/b).
\end{align*}
\end{proof}

\begin{proof}[Proof of Corollary~\ref{prop:naive-class}]
   Note that $Y_{n+1} \in \hat{C}^{\textup{naive}}_{\alpha}(X_{n+1})$ if and only if $\hat{u}^{\textup{naive}}_{Y_{n+1}}(X_{n+1};t^*) > \alpha$. Let $W_t$ denote the calibration conditional coverage $\P{\hat{u}^{\textup{naive}}_{Y_{n+1}}(X_{n+1};t) > \alpha \mid \mathcal{D}_{\text{es-cal}}}$. Then, we have
   \begin{align*}
       \P{Y_{n+1} \in \hat{C}^{\textup{naive}}_{\alpha}(X_{n+1})}
     = \mathbb{E}\left [\P{\hat{u}^{\textup{naive}}_{Y_{n+1}}(X_{n+1};t^*) > \alpha \mid \mathcal{D}_{\text{es-cal}}} \right ]
       = \E{W_{t^*}} \geq \E{\min_{t\in [T]} W_t}.
   \end{align*}
The rest of the proof follows the same argument as in the proof of Proposition \ref{prop:naive-od}.
\end{proof}

\begin{proof}[Proof of Corollary~\ref{prop:naive-reg}]
    Let $\hat{S}_i(X_{n+1}, t) = | Y_i - \hat{\mu}_{t}(X_i)|$ denote the residual score calculated with model $t \in [T]$, for all $i \in \mathcal{D}_{\text{es-cal}} $. Note that $Y_{n+1} \in \hat{C}^{\textup{naive}}_{\alpha}(X_{n+1})$ if and only if $\hat{S}_{X_{n+1}}(X_{n+1}, t^*) \leq \hat{Q}_{1-\alpha}$. Then, we just need to bound $W_t=\P{ \hat{S}_{X_{n+1}}(X_{n+1}, t^*) \leq \hat{Q}_{1-\alpha} \mid \mathcal{D}_{\text{es-cal}}}$, and the rest of the proof follows the same steps as the proof of Proposition~\ref{prop:naive-od}.
\end{proof}

\begin{proof}[Proof of Lemma~\ref{lemma:asymp}]
Recall that $l=\lfloor \alpha(n_{\text{es-cal}}+1) \rfloor $, and define the following helpful notations:
\begin{align*}
    \text{Beta}\left(n_{\text{es-cal}}+1-l, l  \right) \coloneqq \text{Beta}\left(n_{\text{es-cal}}\cdot c, n_{\text{es-cal}}\cdot d \right), \quad \text{where } c=\frac{n_{\text{es-cal}}+1-l}{n_{\text{es-cal}}}, \quad d=\frac{l}{n_{\text{es-cal}}}.
\end{align*}
 Denote $\text{Gamma}(k, \theta)$ as the gamma distribution with shape parameter $k$ and scale parameter $\theta$. It is a well known fact that the beta distribution can be expressed as a ratio of gamma distributions as:
 \begin{align*}
     \text{Beta}\left(n_{\text{es-cal}} \cdot c, n_{\text{es-cal}} \cdot d \right) = \frac{\text{Gamma}(n_{\text{es-cal}} \cdot c , 1)}{\text{Gamma}(n_{\text{es-cal}} \cdot c , 1) + \text{Gamma}(n_{\text{es-cal}} \cdot d , 1)}.
 \end{align*}
Further, $\text{Gamma}(n_{\text{es-cal}} \cdot c , 1)$ can be seen as the distribution of a sum of $n_{\text{es-cal}} \cdot c$ independent exponentially distributed random variables with mean equal to 1; therefore, by the central limit theorem, $\text{Gamma}(n_{\text{es-cal}} \cdot c , 1)$ has an asymptotic Gaussian distribution as $n_{\text{es-cal}} \rightarrow \infty$. Denote $\Phi(x,\mu,\sigma^2)$ as the cumulative distribution function of a Gaussian random variable with mean $\mu$ and variance $\sigma^2$. Applying the delta method, it follows that, in the limit of large $n_{\text{es-cal}}$,
\begin{align*}
     I \left(x;n_{\text{es-cal}}\cdot c, n_{\text{es-cal}}\cdot d \right) = \Phi \left(x; \frac{c}{c+d}, \frac{1}{n_{\text{es-cal}}} \cdot \frac{cd}{(c+d)^3} \right) + O\left( \frac{1}{n_{\text{es-cal}}} \right), \quad \text{for any } x \in [0,1].
\end{align*}
Since $I$ and $\Phi$ are continuous and strictly increasing over $[0,1]$, letting $\Phi^{-1}$ be the inverse Gaussian CDF, we have
\begin{align*}
    I^{-1} \left(\frac{1}{bT};n_{\text{es-cal}}\cdot c, n_{\text{es-cal}}\cdot d \right) 
    &= \Phi^{-1} \left(\frac{1}{bT}; \frac{c}{c+d}, \frac{1}{n_{\text{es-cal}}} \cdot \frac{cd}{(c+d)^3} \right) + O\left( \frac{1}{n_{\text{es-cal}}} \right) \\
    &=\Phi^{-1} \left(\frac{1}{bT};1-\alpha, \frac{\alpha(1-\alpha)}{n_{\text{es-cal}}+1}
    \right) + O\left( \frac{1}{n_{\text{es-cal}}} \right)\\
    &=(1-\alpha) + \sqrt{\frac{\alpha(1-\alpha)}{n_{\text{es-cal}}+1}} \cdot \Phi^{-1} \left(\frac{1}{bT}; 0, 1\right) + O\left( \frac{1}{n_{\text{es-cal}}} \right) \\
    &= (1-\alpha) - \sqrt{\frac{\alpha(1-\alpha)}{n_{\text{es-cal}}+1}}\cdot \sqrt{2\log(bT)} + O\left( \frac{1}{\sqrt{n_{\text{es-cal}}\log(T)}} \right),
\end{align*}
where the second equality is obtained by substituting $c$ and $d$ with their defined values  and the last inequality follows from Equation 26.2.23 in \citet{book_a&s} for sufficiently large $T$.
\end{proof}


\end{document}